\documentclass[runningheads]{llncs}

\usepackage{cite}

\usepackage{graphicx}
\usepackage{amsmath,amssymb} 
\usepackage[usenames, dvipsnames]{color}
\usepackage{placeins}

\usepackage{mathrsfs}
\usepackage{rotating}
\usepackage{caption}
\usepackage{bookmark}

\hypersetup{
    colorlinks=true,
    linkcolor=blue,
    filecolor=magenta,      
    urlcolor=ForestGreen,
}
\usepackage{booktabs,tabularx}
\usepackage[ruled, vlined, linesnumbered]{algorithm2e}
    \SetAlFnt{\small}
    \SetAlCapFnt{\small}
    \SetAlCapNameFnt{\small}
\usepackage{booktabs}
\usepackage{placeins}

\usepackage[width=122mm,left=12mm,paperwidth=146mm,height=193mm,top=12mm,paperheight=217mm]{geometry}
\usepackage{authblk}

\begin{document}
\pagestyle{headings}
\mainmatter

\author{Amirhossein Kardoost\inst{1} \and Kalun Ho\inst{1,2,3} \and
Peter Ochs\inst{4} \and Margret Keuper\inst{1}}

\institute{Data and Web Science Group, University of Mannheim, Germany\\\and Fraunhofer Center Machine Learning, Germany\\ \and Fraunhofer ITWM, Competence Center HPC, Kaiserslautern, Germany\\ \and
Mathematical Optimization Group, Saarland University, Germany}


\title{Self-supervised Sparse to Dense Motion Segmentation} 
\titlerunning{Self-supervised Sparse to Dense Motion Segmentation}


\authorrunning{A. Kardoost et al.}


\maketitle

\vspace{-8pt}
\begin{abstract}
Observable motion in videos can give rise to the definition of objects moving with respect to the scene. The task of segmenting such moving objects is referred to as motion segmentation
and is usually tackled either by aggregating motion information in long, sparse point trajectories, or by directly producing per frame dense segmentations relying on large amounts of training data.
In this paper, we propose a self supervised method to learn the densification of sparse motion segmentations from single video frames. While previous approaches towards motion segmentation build upon pre-training on large surrogate datasets and use dense motion information as an essential cue for the pixelwise segmentation, our model does not require pre-training and operates at test time on single frames. It can be trained in a sequence specific way to produce high quality dense segmentations from sparse and noisy input. 
We evaluate our method on the well-known motion segmentation datasets $FBMS59$ and $\mathrm{DAVIS}_{16}$.
\end{abstract}

\section{Introduction}
\label{sec:introduction}

\begin{figure}[t]
    \begin{center}
	    \small 
	    \begin{tabular}{@{}c@{}c@{}c@{}c@{}}
	        \includegraphics[width=0.24\textwidth]{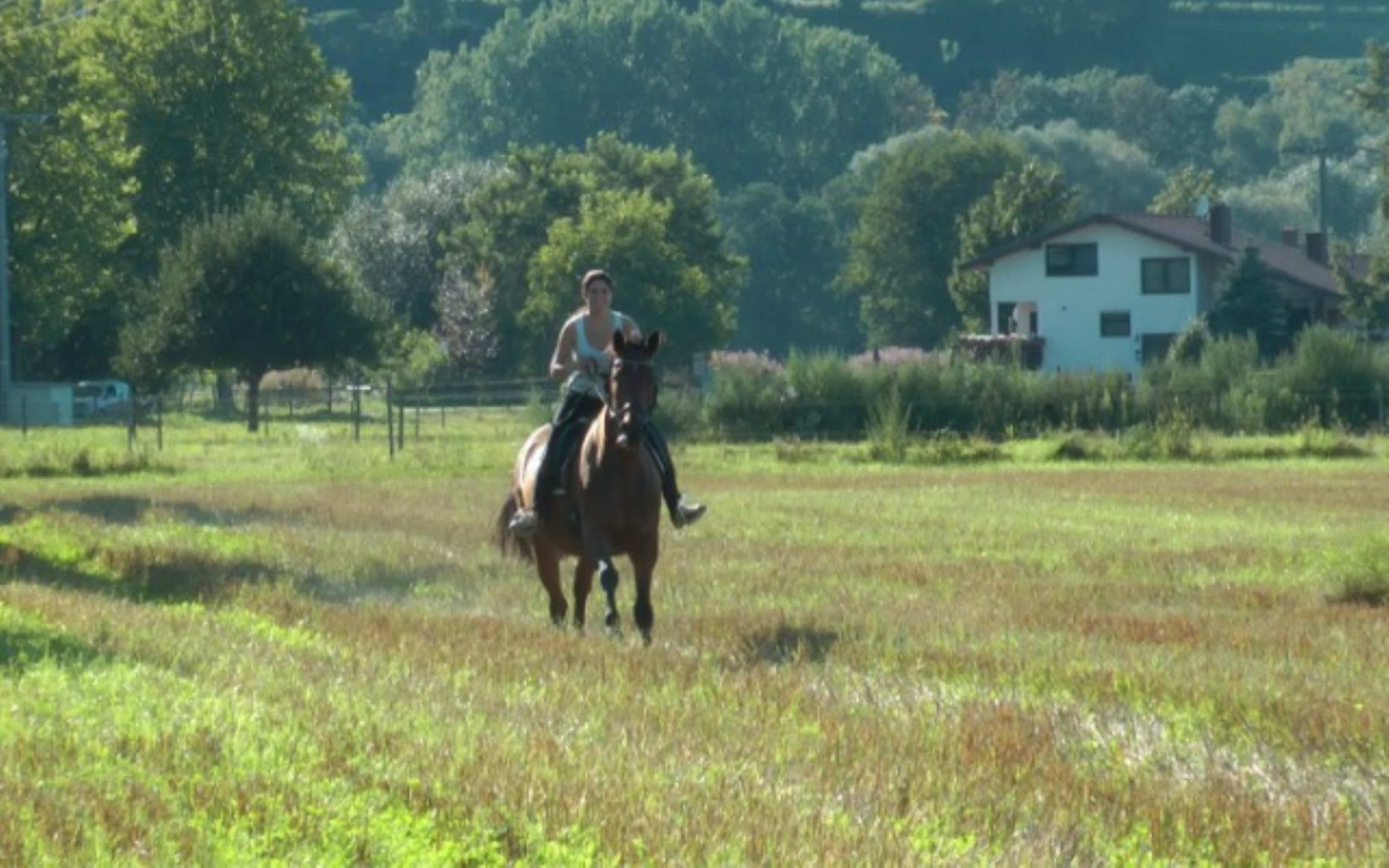}\,
	        \includegraphics[width=0.24\textwidth]{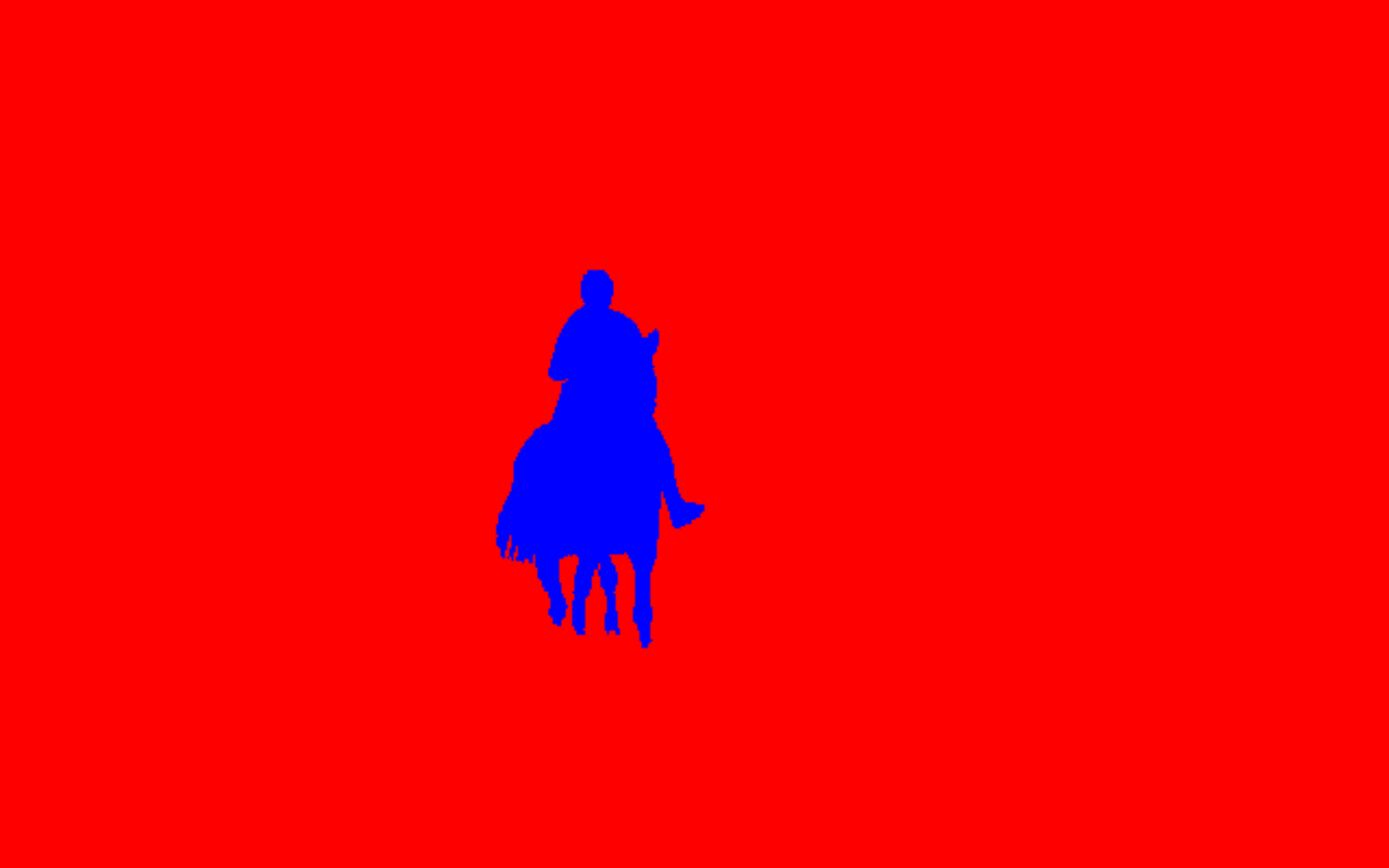}\,
            \includegraphics[width=0.24\textwidth]{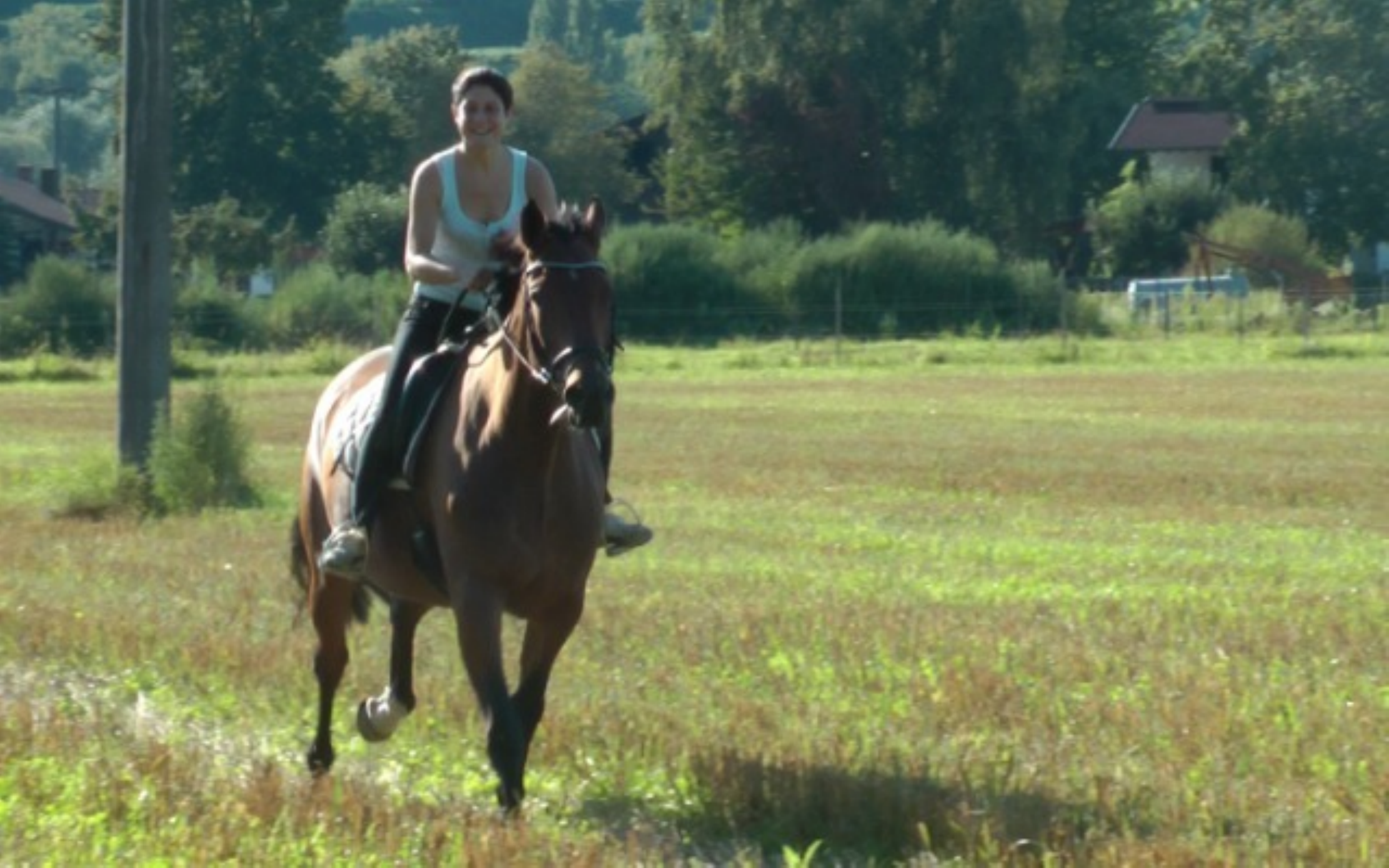}\,
            \includegraphics[width=0.24\textwidth]{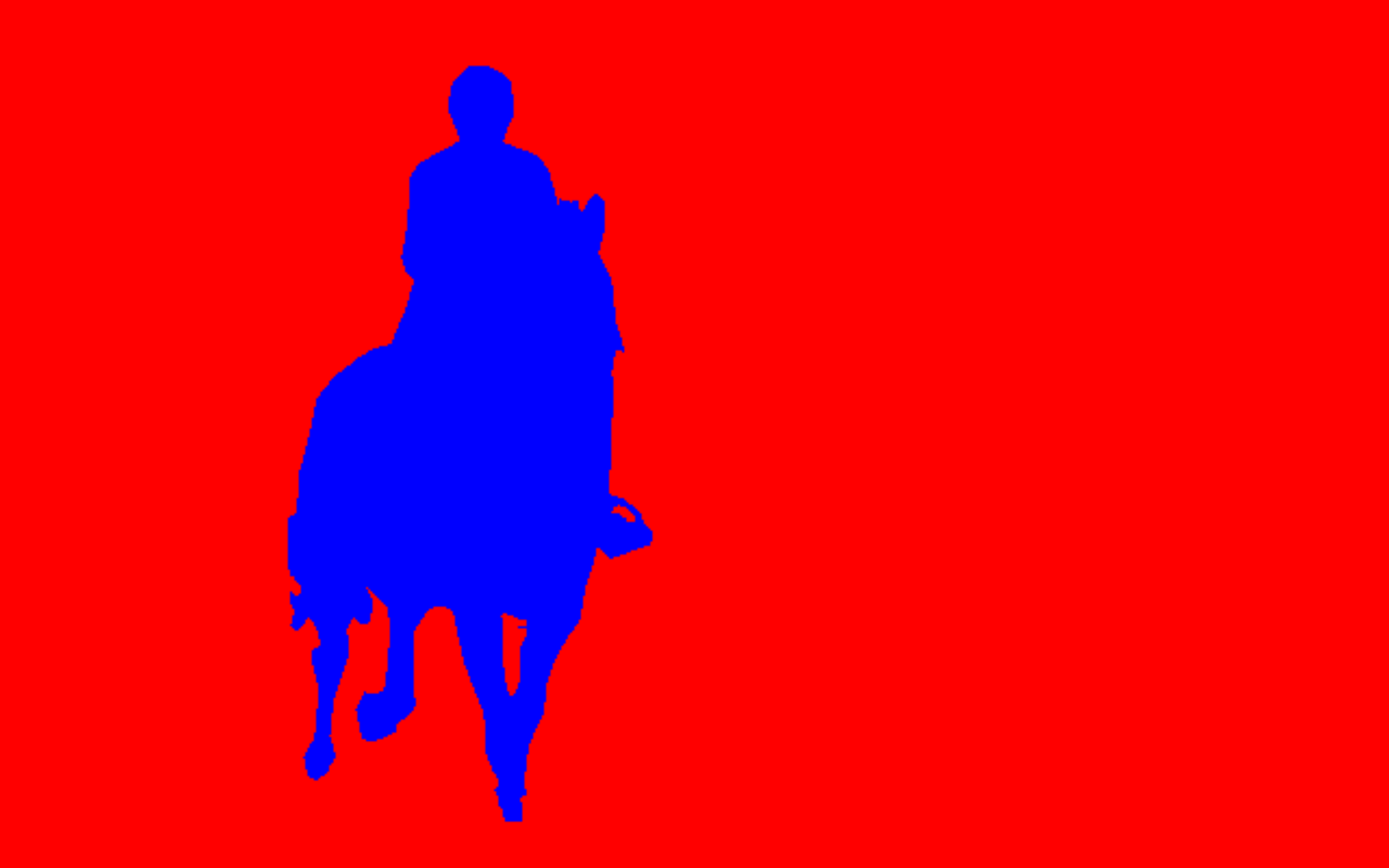}
        \end{tabular}
        \caption{Motion segmentation example is provided for frames 1 and 160 of the ``horses01'' sequence (with their ground-truth motion segmentation) in the FBMS59 \cite{Ochs14} dataset. Due to the reason that the person and the horse are moving together and have the same motion pattern, they are assigned the same motion label (blue color).}
        \label{motion_seg_example}
    \end{center}
\end{figure}

The importance of motion for visual learning has been emphasized in recent years. Following the Gestalt principle of common fate~\cite{gestaltp}, motion patterns within an object are often more homogeneous than its appearance, and therefore provide reliable cues for segmenting (moving) objects in a video. 
Motion segmentation is the task of segmenting motion patterns. This is in contrast to semantic segmentation, where one seeks to assign pixel-wise class labels in an image. Thus, for motion segmentation, we need motion information and at least two frames to be visible in order to distinguish between motion segments. Ideally, such motion patterns separate meaningful objects., e.g. an object moving w.r.t.~the scene 
(refer to Fig.\ref{motion_seg_example}). 
To illustrate the importance of motion segmentation, consider an autonomous driving scenario. A first step to classify potential danger caused by the presence of a possibly static object, e.g., a parking car, is the knowledge about its mobility. Unknowingly waiting for the observation that the door of the parking car suddenly opens may be too late to avoid an accident. The speed of the autonomously driving vehicle must be adjusted based on the mobility and danger of other objects. 
While models for the direct prediction of pixel-wise motion segmentations are highly accurate~\cite{TokmakovICCV,Tokmakov17}, they can only take very limited account of an objects motion history. 

\begin{figure}[t]
    \begin{center}
	    \small
	    \begin{tabular}{@{}c@{}c@{}c@{}}
	        \includegraphics[width=0.25\textwidth]{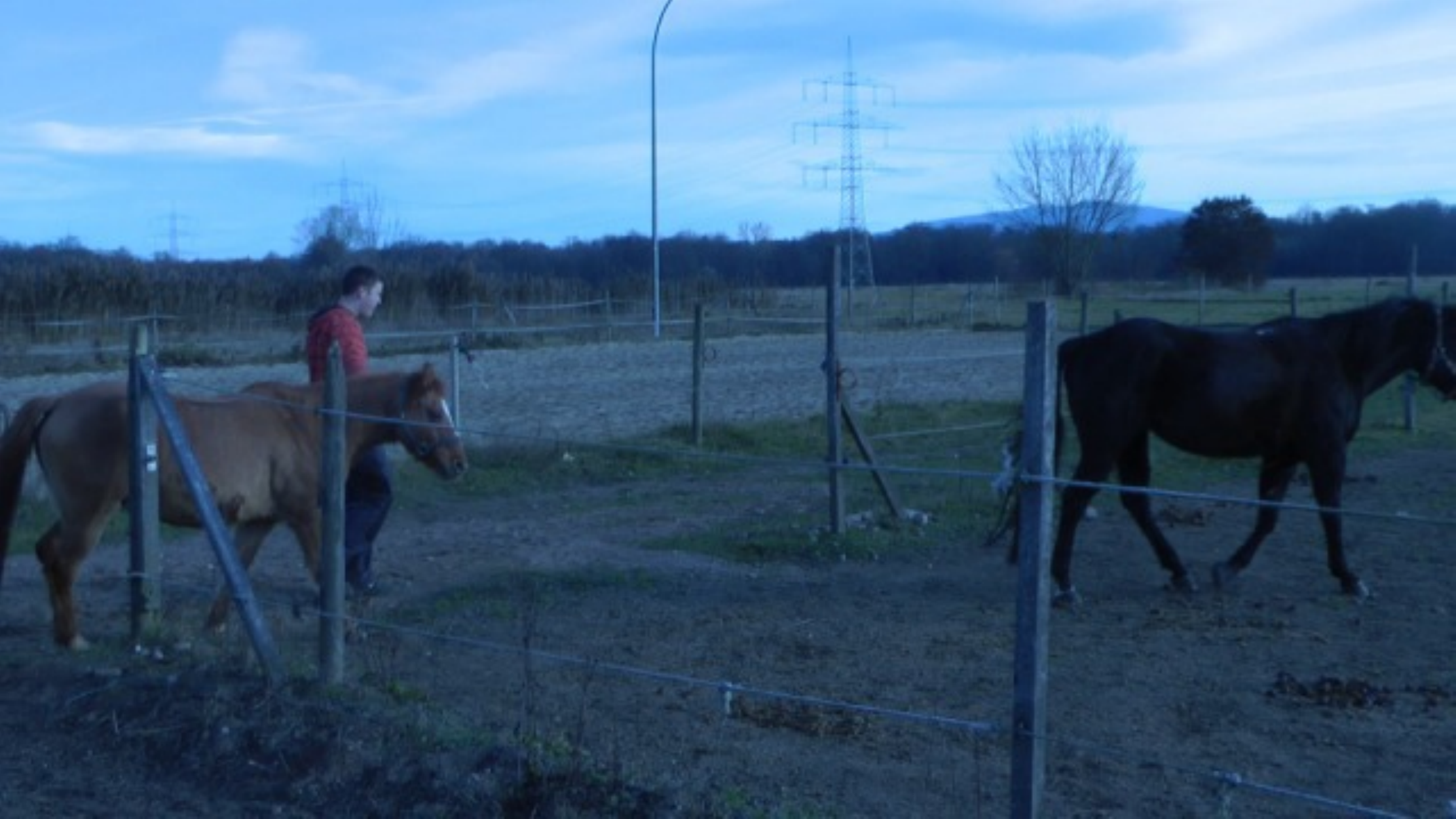}\,
	        \includegraphics[width=0.25\textwidth]{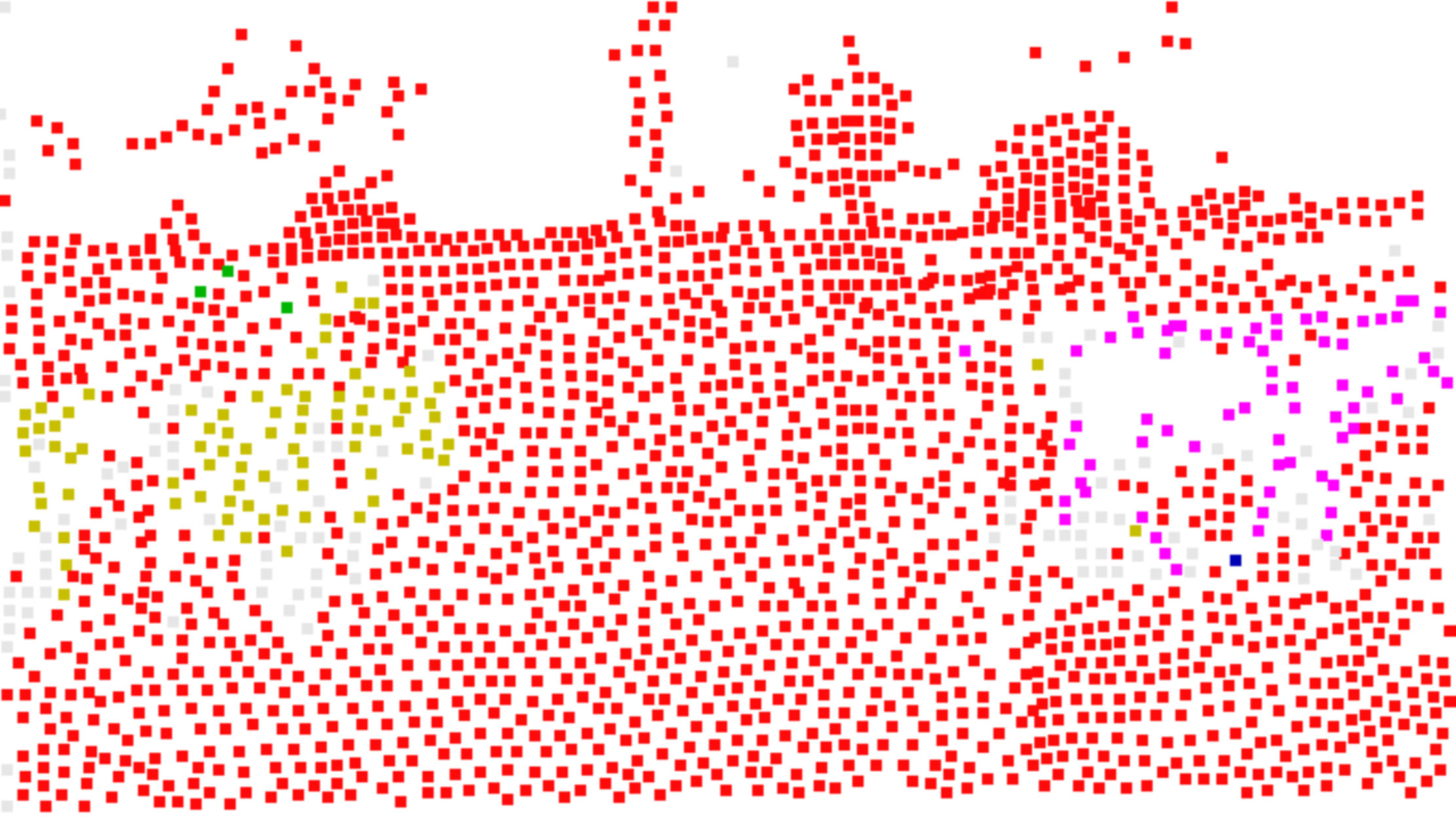}\,
            \includegraphics[width=0.25\textwidth]{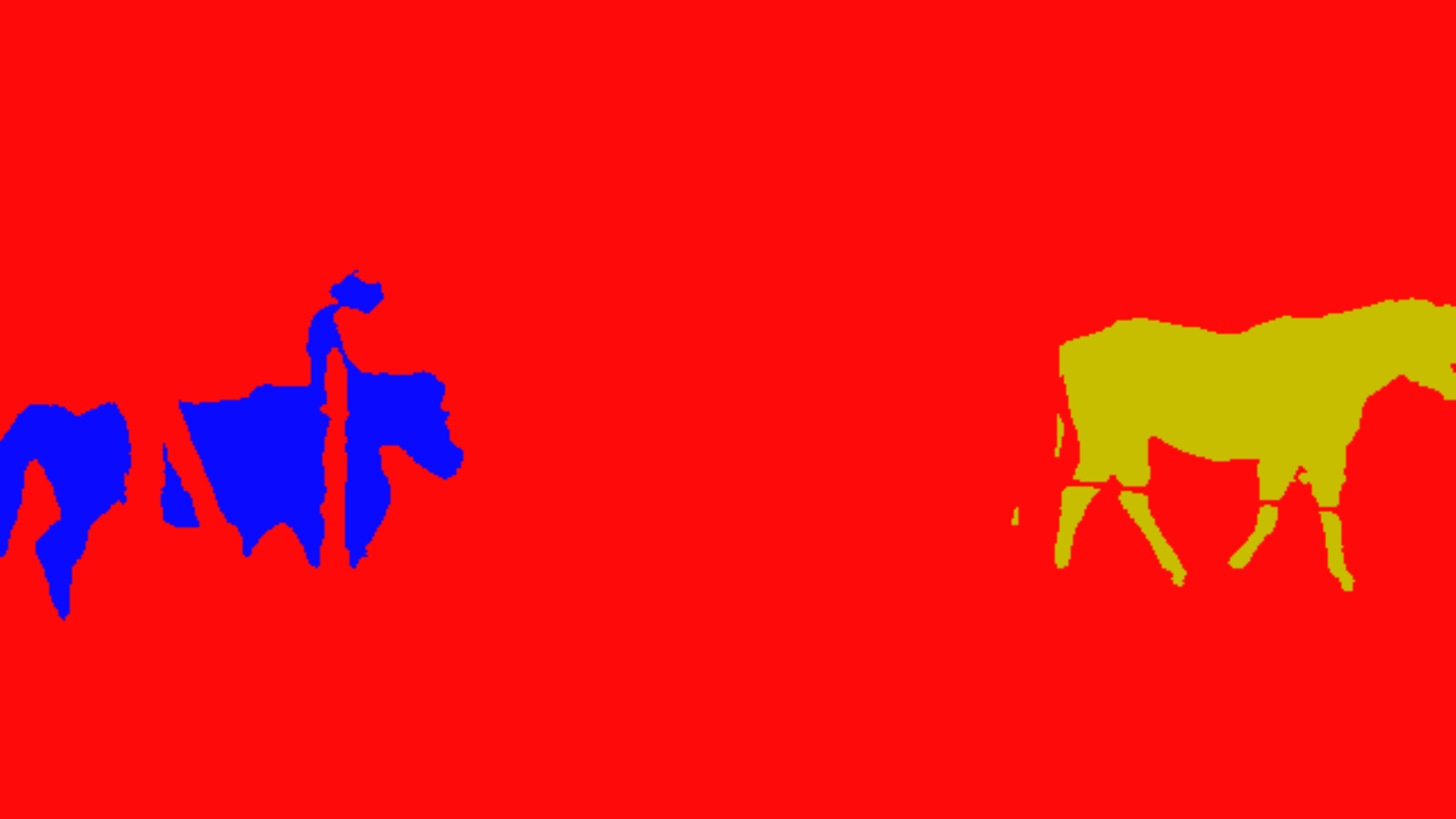}
        \end{tabular}
        \caption{Exemplary multi-label motion segmentation results showing \emph{(left)} the image and its  sparse \emph{(middle)} and dense \emph{(right)} segmentation. The sparse segmentation is produced by \cite{Keuper_15} and the dense segmentation is the result of the proposed model.}
        \label{page1_fig}
    \end{center}
\end{figure}

In this paper, we propose a model to produce high quality dense segmentations from sparse and noisy input (i.e. \emph{densification}). It can be trained in a sequence specific way using sparse motion segmentations as training data, i.e. the densification model can be trained in a self-supervised way. 
Our approach is based on sparse (semi-dense) motion trajectories that are extracted from videos via optical flow (Fig.~\ref{page1_fig}, center). Point trajectory based motion segmentation algorithms have proven to be robust and fast~\cite{Ochs14,Keuper_15,Keuper_17,Bro10c,FragkiadakiECCV12,shiICCV2013,vidal}. 
%
By a long term analysis of a whole video shot at once by the means of such trajectories, even objects that are static for most of the time and only move for few frames can be identified, i.e. the model would not ``forget'' that a car has been moving, even after it has been static for a while. The same argument allows articulated motion to be assigned to a common moving object. 

In our approach we use object annotations that are generated using established, sparse motion segmentation techniques~\cite{Keuper_15}. We also propose an alternative, a GRU-based multicut model, which allows to learn the similarity between the motion of two trajectories and potentially allows for a more flexible application. In both cases, the result is a sparse segmentation of video sequences, providing labels only for points lying on the original trajectories, e.g.~every 8 pixels (compare Fig.~\ref{page1_fig}). 
From such sparse segmentations, pixel-wise segmentations can be generated by variational methods~\cite{Ochs14}. In order to better leverage the consistency within the video sequences, we propose to train sequence specific densification models using only the sparse motion segmentations as labels. 

Specifically, we train a U-Net like model~\cite{unet} to predict dense segmentations from given images (Fig.~\ref{page1_fig}), while we only have sparse and potentially noisy labels given by the trajectories. The training task can thus be interpreted as a label densification. Yet, the resulting model does not need any sparse labels at test time but can generalize to unseen frames.

In contrast to end-to-end motion segmentation methods such as e.g. \cite{Tokmakov17}, we are not restricted to binary labels but can distinguish between different motion patterns belonging to different objects and instances per image and only require single images to be given at test time. Also, in contrast to such approaches, our model does not rely on the availability of large surrogate datasets such as ImageNet~\cite{imagenet_cvpr09} or FlyingChairs~\cite{FlyingChairs} for pre-training but can be trained directly on the sequences from for example the $\mathrm{FBMS59}$ \cite{Ochs14} and $\mathrm{DAVIS}_{16}$ \cite{davis_16} datasets.

To summarize, we make the following contributions:
\begin{itemize}
    \item We provide an improved affinity graph for motion segmentation in the minimum cost multicut framework using a GRU-model. Our model generates a sparse segmentation of motion patterns.
    \item We utilize the sparse and potentially noisy motion segmentation labels to train a U-Net model to produce class agnostic and dense motion segmentation. Sparse motion labels are not required during prediction.
    \item We provide competitive video object segmentation and motion segmentation results on the $\mathrm{FBMS59}$ \cite{Ochs14} and $\mathrm{DAVIS}_{16}$ \cite{davis_16} datasets.
\end{itemize}


\section{Related Work}

Our goal is to learn to segment moving objects based on their motion pattern. For efficiency and robustness, we focus on point trajectory based techniques as initial cues. 
Trained in a sequence specific way, our model can be used for label \emph{densification}. We therefore consider in the following related work in the areas \emph{motion segmentation} and \emph{sparse to dense labeling}.

\subsection{Motion Segmentation}

Common end-to-end trained CNN based approaches to motion segmentation are based on single frame segmentations from optical flow \cite{Tokmakov17,TokmakovICCV,tdms,learn_segmo,modnet}. Tokmakov et al.~\cite{Tokmakov17,TokmakovICCV} make use of large amounts of synthetic training data \cite{FlyingChairs} to learn the concept of object motion. Further, \cite{Tokmakov17} combine these cues with an ImageNet~\cite{imagenet_cvpr09} pre-trained appearance stream and achieve long-term temporal consistency by using a GRU optimized on top. They learn a binary video segmentation and distinguish between static and moving elements.
Siam et al.~\cite{modnet} use a single convolutional network to model motion and appearance cues jointly for autonomous driving. A frame-wise classification problem is formulated in \cite{learn_segmo} to detect motion saliency in videos.
In \cite{tdms} for each frame multiple figure-ground segmentations are produced based on motion boundaries. A moving objectness detector trained on image and motion fields is used to rank the segment candidates. Methods like \cite{moving,moes_var} approach the problem in a probabilistic way. In \cite{moving} the camera motion is subtracted from each frame to improve the training. A variational formulation based on optical flow is used in \cite{moes_var}. Addressing motion segmentation is different than object segmentation, as in motion segmentation, different motion patterns are segmented, which makes connected objects seem as one object if they move together with the same motion pattern. As an example, we refer to the example of a person riding a horse (Fig. \ref{motion_seg_example}). As long as they move together they will be considered as same object while in object segmentation we deal with two separate objects, this makes our method different than object segmentation approaches~\cite{vide_seg_1,video_seg2,video_seg_3}.

A different line of work relies on point trajectories for motion segmentation.
Here, long-term motion information is first used to establish sparse but coherent motion segments over time. Dense segmentations are usually generated in a post-processing step such as \cite{Ochs14}. However, in contrast to end-to-end CNN motion segmentation approaches, trajectory based methods have the desirable property to directly handle multiple motion pattern segmentations.
While in \cite{Keuper_15,Keuper_17} the partitioning of trajectories into motion segments uses the minimum cost multicut problem, other approaches employ sparse subspace clustering \cite{Elhamifar2009_6} or spectral clustering and normalized cuts \cite{HO_SC,moseg_norm_cut,Fragkiadaki11}. In this setting, a model selection is needed to determine the final number of segments. 
In contrast, the minimum cost multicut formulation allows for a direct optimization of the number of components via repulsive cost. 

Our GRU approach uses the same graph construction policy as \cite{Keuper_15} for motion segmentation while the edge costs are assigned using a Siamese (also known as twin) gated recurrent network. The Siamese networks~\cite{siamese_net} are metric learning approach used to provide a comparison between two different inputs. Similar trajectory embeddings have been used in \cite{Apratim_pred} to predict a pedestrians future walking direction. For motion segmentation, we stick to the formulation as a minimum cost multicut problem \cite{Keuper_15}. 

\subsection{Sparse to Dense Labeling}

While the trajectory based motion segmentation methods can propagate object information through frames, they produce sparse results. Therefore, specific methods, e.g. \cite{Ochs14,SO16} are needed to produce dense results.
A commonly used densification approach is the variational framework of Ochs et al.~\cite{Ochs14}. In this approach the underlying color and boundary information of the images are used for the diffusion of the sparsely segmented trajectory points, which sometimes leaks the pixel labels to unwanted regions, e.g. loosely textured areas of the image. 
Furthermore, \cite{trace_discon} address the densification problem using Gabriel graphs as per frame superpixel maps. Gabriel edges bridge the gaps between contours using geometric reasoning. However, super-pixel maps are prone to neglect fine structure of the underlying image and leads to low segmentation quality.

Our method benefits from trajectory based methods for producing a sparse multi-label segmentation. A sparsely trained U-Net \cite{unet} produces dense results for each frame purely from appearance cues, potentially specific for a scene or sequence.

\section{Proposed Self-Supervised Learning Framework}

\begin{figure}[t]
	\centering
	\small
	\includegraphics[width=0.99\textwidth]{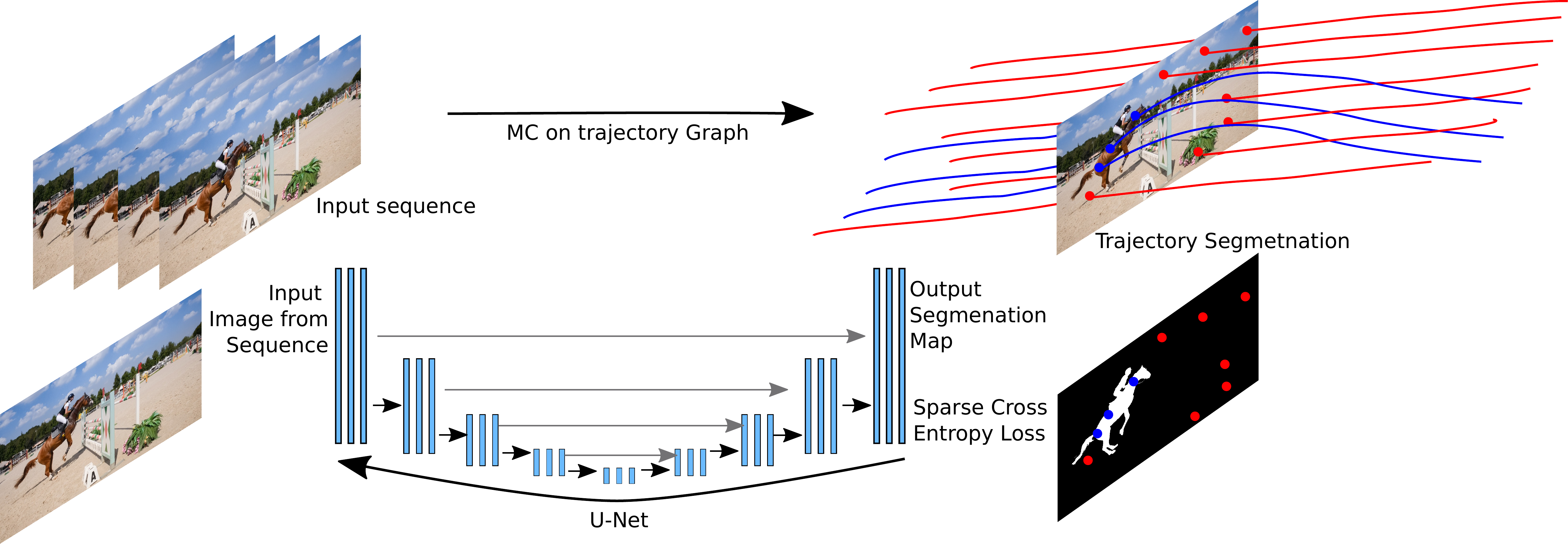}
	\caption{Sparsely segmented trajectories are produced by minimum cost multicuts (MC) either with our Siamese-GRU model or simple motion cues as in \cite{Keuper_15} (top). The sparsely labeled points are used to train the U-Net model (bottom). At test time, U-Net model can produce dense segmentations without requiring any sparse labels as input.}
	\label{overview}
\end{figure}

We propose a self-supervised learning framework for sparse-to-dense segmentation of the sparsely segmented point trajectories. In another words, a U-Net model is trained from sparse annotations to estimate dense segmentation maps (Section~\ref{sec:DL-model}). The sparse annotations can be provided either with some potentially unsupervised state-of-the-art trajectory segmentation methods~\cite{Keuper_15} or our proposed Siamese-GRU model.


\subsection{Annotation Generation} 
\label{sec:unsupervised-data-gen}

Point trajectories are generated from optical flow~\cite{Bro10c}. Each point trajectory corresponds to the set of sub-pixel positions connected through consecutive frames using the optical flow information. Such point trajectories are clustered by the minimum cost multicut approach~\cite{andres-2012-globally} (aka. correlation clustering) with respect to their underlying motion model estimated (i) from a translational motion model or (ii) from a proposed Siamese GRU network.

\subsubsection{Point Trajectories} are spatio-temporal curves represented by their frame-wise sub-pixel-accurate (x,y)-coordinates. They can be generated by tracking points using optical flow by the method of Brox et al.~\cite{Bro10c}. The resulting trajectory set aims for a minimal target density (e.g.~one trajectory in every 8 pixels). Trajectories are initialized in some video frame and end when the point cannot be tracked reliably anymore, e.g.~due to occlusion.
In order to achieve the desired density, possibly new trajectories are initialized throughout the video. Using trajectories brings the benefit of accessing the object motion in prior frames.

\subsubsection{Translational Motion Affinities} can be assigned based on motion distances of each trajectory pair with some temporal overlap~\cite{Keuper_15}. For trajectories $A$ and $B$, the frame-wise motion distance at time $t$ is computed by

\begin{equation}
d_t(A, B) = \frac{\rVert \partial_tA - \partial_tB \rVert}{\sigma_t}.
\end{equation}

It solely measures in-plane translational motion differences, normalized by the variation of the optical flow $\sigma_t$ (refer to \cite{Ochs14} for more information). The $\partial_tA$ and $\partial_tB$ represent the partial derivatives of $A$ and $B$ with respect to the time dimension.

The overall motion distance of a pair of trajectories $A$ and $B$ is computed by maximum pooling over their joint life-time, 

\begin{equation}
    d^{motion}(A, B) = \max_t d_t (A, B)
\label{max_motion}
\end{equation}

Color and spatial cues are added in~\cite{Keuper_15} for robustness. Instead of using translational motion affinities we propose a Siamese Gated Recurrent Units (GRUs) based model to provide affinities between the trajectory pairs.

\subsubsection{Siamese Gated Recurrent Units (GRUs)} 
\label{siam_GRU_model}
can be used to learn trajectory affinities by taking trajectory pairs as input. The proposed network consists of two legs with shared weights, where in our model each leg consists of a GRU model which takes a trajectory as input. Specifically, for two trajectories $A$ and $B$, the $\partial A$ and $\partial B$ (partial derivative of the trajectories with respect to the time dimension) on the joint life-time of the trajectories are given to each leg of the Siamese-GRU model. The partial derivatives represent their motion information, while no information about their location in image coordinates is provided. The motion cues are embedded by the GRU network, i.e.~the hidden units are gathered for each time step. Afterwards, the difference between two embedded motion vectors $embed_{\partial A}$ and $embed_{\partial B}$ is computed as

\begin{equation}
d_{(embed_{\partial A}, embed_{\partial B})} = \sum_{i=1}^h ({embed_{\partial A}}_i - {embed_{\partial B}}_i)^2,
\label{diff_equ}
\end{equation}

where $h$ denotes the number of hidden units for each time step. The result of equation \eqref{diff_equ} is a vector which is consequently given to two fully connected layers and the final similarity value is computed by a Sigmoid function.
Therefore, for each pair of trajectories given to the Siamese~\cite{siamese_net} GRU network, it provides a measure of their likelihood to belong to the same motion pattern.

The joint life-time of the two trajectories could in practice be different from pair to pair and the GRU network requires a fixed number of time steps as input ($N$). This problem is dealt as follows:

If the joint life-time of the two trajectories is less than $N$, each trajectory is padded with its respective final partial derivative value in the intersection part.
Otherwise, when the joint life-time has more than $N$ time steps, the time step $t$ with maximum motion distance, similar to equation \eqref{max_motion}, is determined for the entire lifetime, 
\begin{equation}
    t_{A, B} = \arg\max_t d_t (A, B).
\label{max_motion1}
\end{equation}
The trajectory values are extracted before and after $t$ so that the required number of time steps $N$ is reached. The reason for doing this is that the important part of the trajectories are not lost. Consider a case where an object does not move in the first $x$ frames and starts moving from frame $x+1$, the most important information will be available around frames $x$ and $x+1$ and it is better not to lose such information by cutting this part out. 

In our approach, the frame-wise Euclidean distance of trajectory embeddings (extracted from the hidden units) of the GRU model is fed to a fully connected layer for dimensionality reduction and passed to a Sigmoid function for classification into the classes 0 (same label - pair of trajectories belong to the same motion pattern) or 1 (different label - pair of trajectories belong to different motion pattern) using a mean squared error (MSE) loss.

To train the Siamese-GRU model two labels are considered for each pair of trajectories, label 0 where the trajectory pair correspond to the same motion pattern and label 1 otherwise (the trajectories belong to different motion patterns). 
To produce the ground-truth labeling to train the Siamese-GRU model, a subset of the edges in the produced graph $G=(V,E)$ by the method of Keuper~et~al.~\cite{Keuper_15} are sampled (information about the graph is provided in the next paragraph). For each edge, which corresponds to a pair of trajectories, we look into each trajectory and its label in the provided ground-truth segmentation. We only take those trajectories which belong to exactly one motion pattern in the provided ground-truth. Some trajectories change their labels while passing through frames which are considered as unreliable.
Furthermore, the same amount of edges with label 0 (same motion pattern) and label 1 (different motion pattern) are sampled to have a balanced training signal. At test time, costs for each edge $E$ in graph $G=(V,E)$ are generated by the trained Siamese-GRU network. 

\subsubsection{Trajectory Clustering}
\label{sec:cluster_trajs}
yields a grouping of trajectories according to their modeled or learned motion affinities. 
We formalize the motion segmentation problem as minimum cost multicut problem~\cite{Keuper_15}. It aims to optimally decompose a graph $G=(V,E)$, where trajectories are represented by nodes in $V$ and their affinities define costs on the edges $E$. In our approach the costs are assigned using the Siamese-GRU model.

While the multicut problem is APX-hard~\cite{Bansal2004}, heuristic solvers~\cite{KL,KB15a,Beier2016,node_agglom,BAPCW} are expected to provide practical solutions. We use the Kernighan Lin \cite{KL} implementation of \cite{keuper15a}. This solver is proved to be practically successful in motion segmentation \cite{Keuper_15,Keuper_17}, image decomposition and mesh segmentation~\cite{KB15a} scenarios. 

\subsection{Deep Learning Model for Sparse to Dense Segmentation} 
\label{sec:DL-model}
We use the sparse motion segmentation annotated video data as described in Section~\ref{sec:unsupervised-data-gen} for our deep learning based sparse-to-dense motion segmentation model. Specifically, the training data consist of input images (video frames) or edge maps and their sparse motion segmentations, which we use as annotations. 
Although, the loss function only applies at the sparse labels, the network learns to predict dense segmentations.

\subsubsection{Deep Learning Model} 

We use a U-Net \cite{unet} type architecture for dense segmentation, which is known for its high quality predictions in tasks such as semantic segmentation \cite{rtseg,desemseg,stdenet}. A U-Net is an encoder-decoder network with skip connections. During encoding, characteristic appearance properties of the input are extracted and are learnt to be associated with objectness. In the decoding phase, the extracted properties are traced back to locations causing the observed effect, while details from the downsampling phase are taken into account to ease the localisation (see Fig.~\ref{overview} for details). The output is a dense (pixel-wise) segmentation of objects, i.e., a function $u\colon \Omega \to \{1,\ldots,K\}$, where $\Omega$ is the image domain and $K$ is the number of classes which corresponds to number of trajectory labels. This means that, after clustering the trajectories each cluster takes a label and overall we have class-agnostic motion segmentation of sparse trajectories.
Such labels are only used during training. 

\subsubsection{Loss Function}  
The U-Net is trained via the Binary Cross Entropy (BCE) and Cross Entropy (CE) loss function for the single and multiple object case, respectively.
As labels are only present at a sparse subset of pixels, the loss function is restricted to those pixels. Intuitively, since the label locations where the loss is evaluated are unknown to the network, it is forced to predict a label at every location. (A more detailed discussion is provided below.)

\subsubsection{Dense Predictions with Sparse Loss Functions} 
At first glance, a sparse loss function may not force the network to produce a meaningful dense segmentation. Since trajectories are generated according to a simple deterministic criterion, namely extreme points of the local structure tensor~\cite{Bro10c}, the network could internally reproduce this generation criterion and focus on labeling such points only.
Actually, we observed exactly the problematic behaviour mentioned above, and, therefore, suggest variants for the learning process employing either RGB images or (deterministic) Sobel edge maps \cite{sobel} as input. One remedy is to alleviate the local structure by smoothing the input RGB-image, making it harder for the network to pick up on local dominant texture and to stimulate the usage of globally extracted features that can be associated with movable object properties.

\subsubsection{Conditional Random Filed (CRF) Segmentation Refinement}
To build the fine-grained segmentation maps from the blurred images, we employ Conditional Random Fields (CRF): We compare 
\begin{itemize}
    \item the fully connected pre-trained CRF layer (dense-CRF) \cite{dense_crf}, with parameters learnded from pixel-level segmentation~\cite{chen14semantic} and 
    \item a CRFasRNN \cite{crfasrnn_iccv2015} model which we train using the output of our U-Net model as unaries on our same sparse annotations. To generate a training signal even in case the U-Net output perfectly fits the sparse labels, we add Gaussian noise to the unaries.
\end{itemize}

\subsubsection{Discussion: Sparse Trajectories vs. Dense Segmentation} 
The handling of sparse labels could be avoided if dense unsupervised motion segmentations were given. Although, in principle, dense trajectories can be generated by the motion segmentation algorithm, the clustering algorithm does not scale linearly 
with the number of trajectories and the computational cost explodes.
Instead of densely tracking pixels throughout the video, a frame-wise computationally affordable densification step, for example based on variational or inpainting strategies~\cite{Ochs14}, could be used. However, some sparse labels might be erroneous, an issue that can be amplified by the densification. Although some erroneous labels can also be corrected by~\cite{Ochs14}, especially close to object boundaries, we observed that the negative effect prevails when it comes to learning from such unsupervised annotations. Moreover, variational methods often rely on local information to steer the propagation of label information in the neighborhood. In contrast, the U-Net architecture can incorporate global information and possibly objectness properties to construct its implicit regularization.

\section{Experiments}
We evaluate the performance of the proposed models on the two datasets \hbox{} $\mathrm{DAVIS}_{16}$~ \cite{davis_16} and FBMS59 \cite{Ochs14}, which contain challenging video sequences with high quality segmentation annotations of moving objects.

\subsection{Datasets and Implementation Details}
\subsubsection{Datasets}
The $\mathrm{DAVIS}_{16}$ dataset \cite{davis_16} is produced for high-precision binary object segmentation tracking of rigidly and non-rigidly moving objects. It contains 30 train and 20 validation sequences. The pixel-wise binary ground truth segmentation is provided per frame for each sequence. The evaluation metrics are Jaccard index (also known as Intersection over Union) and F-measure. Even though this dataset is produced for object segmentation, it is commonly used to evaluate motion segmentation because only one object is moving in each sequence which makes the motion pattern of the foreground object to be different from the background motion.

The FBMS59 \cite{Ochs14} dataset is specifically designed for motion segmentation and consists of 29 train and 30 test sequences. The sequences cover camera shaking, rigid/non-rigid motion as well as occlusion/dis-occlusion of single and multiple objects with ground-truth segmentations given for a subset of the frames. We evaluate precision, recall and F-measure. 

\subsubsection{Implementation Details}
Our Siamese-GRU model with 2 hidden units ($h=2$, equation~\ref{diff_equ}) and experimentally selected 25 time steps ($N=25$, for more information refer to section \ref{siam_GRU_model}) is trained for 3 epochs, a batch size of 256 and a learning rate of $0.001$ where the trajectories are produced by large displacement optical flow (LDOF) \cite{LDOF} at 8 pixel sampling on the training set of $\mathrm{DAVIS}_{16}$ \cite{davis_16}. 

We employ two different strategies of using the sparse motion segmentations of the resulting trajectories as labels, depending whether we face binary ($\mathrm{DAVIS}_{16}$) or multi-label (FBMS59) problems. In case of single label, the most frequent trajectory label \emph{overall frames} and second most frequent label \emph{per frame} are considered as background and foreground, respectively. For multi-label cases, the most frequent class-agnostic labels are selected, i.e. we take only those labels which are frequent enough compared to the other labels. 

Our U-Net model is trained in a sequence specific way. Such model can be used for example for label densification and is trained using color and edge-map data with a learning rate of $0.01$ and batch size of 1 for 15 epochs. The overall train and prediction process takes around (maximally) 30 minutes per sequence on a NVIDIA Tesla V100 GPU. 
The CRFasRNN \cite{crfasrnn_iccv2015} is trained with learning rate of $0.0001$, batch size 1 and 10 epochs with 5 and 10 inference iterations at train and test time, respectively.
\begin{table}[t]
\centering
\caption{The trajectories are segmented by 1. the method of Keuper et al. \cite{Keuper_15} and 2. our Siamese-GRU model. The densified results are generated based on 1. the method of Ochs et al. \cite{Ochs14} and 2. the proposed U-Net model. The results are provided for the validation set of $\mathrm{DAVIS}_{16}$.}\small 
\begin{tabular}{l|l|c}
\toprule
Traj. Seg. Method & Densification Method   & Jaccard Index 
\\ \midrule
Keuper et al. \cite{Keuper_15}& Ochs et al. \cite{Ochs14} & 55.3 \\
Keuper et al. \cite{Keuper_15}& U-Net model (ours)        & 58.5 \\
Siamese-GRU (ours)                   & Ochs et al. \cite{Ochs14} & 57.7  \\ 
Siamese-GRU (ours)                  & U-Net model (ours)               & \bf{66.2}  \\ 
\bottomrule
\end{tabular}
\label{tab:motionseg}
\end{table}
\subsection{Sparse Trajectory Motion-Model}
We first evaluate our GRU model for sparse motion segmentation on the validation set of $\mathrm{DAVIS}_{16}$ \cite{davis_16}. Therefore, we produce, in a first iteration, densified segmentations using the variational approach from \cite{Ochs14}. The results are given in Tab. \ref{tab:motionseg} (line 3) and show an improvement over the motion model from Keuper et al.~\cite{Keuper_15} by 2\% in Jaccard index. In the following experiments on $\mathrm{DAVIS}_{16}$, we thus use sparse labels from our Siamese GRU approach.

\subsubsection{Knowledge Transfer}
Next, we investigate the generalization ability of this motion model. We evaluate the $\mathrm{DAVIS}_{16}$-trained GRU-model on the train set of FBMS59 \cite{Ochs14}. Results are given in Tab. \ref{tab:davis_to_fbms}. While this model performs below the hand-tuned model of \cite{Keuper_15} on this dataset, results are comparable, especially considering that our GRU model does not use color information. In further experiments on FBMS59, we use sparse labels from \cite{Keuper_15}.

\begin{table}[t]
\centering
\caption{Sparse Motion Segmentation trained on $\mathrm{DAVIS}_{16}$ (all seq.) and evaluated on FBMS59 (train set). 
We compare to \cite{Keuper_15} and their variant only using motion cues.
}\small
\begin{tabular}{l| c c c | c}
\toprule
                                         & Precision       & Recall &  F-measure  & \#of extracted objs.   \\   \midrule
Keuper et al. \cite{Keuper_15} (motion cues) & 83.88          & 69.97 & 76.30 & 20\\
Keuper et al. \cite{Keuper_15} (full)              &83.69            & 73.17 & \bf{78.07} & \bf{27}  \\
Siamese-GRU (ours - transfer)                              &81.01            & 70.07 & 75.14 & 24  \\
\bottomrule
\end{tabular}
\label{tab:davis_to_fbms}
\end{table}

\subsection{Dense Segmentation of Moving Objects}
\label{dense_seg_move_obj}

Next, we evaluate our self-supervised dense segmentation framework with sequence specific training on the color images as well as edge maps on the validation set of $\mathrm{DAVIS}_{16}$ \cite{davis_16}.
Tab. \ref{tab:motionseg} shows that this model, trained on the GRU based labels, outperforms the model trained on the motion models from \cite{Keuper_15} as well as the densification of Ochs et al.~\cite{Ochs14} by a large margin. 
Tab. \ref{densif} (top) provides a comparison between different versions of our model, the densification model of Ochs et al. \cite{Ochs14}, and the per-frame evaluation of Tokmakov et al. \cite{TokmakovICCV} on $\mathrm{DAVIS}_{16}$. \cite{TokmakovICCV} use large amounts of data for pre-training. Their better performing model variants require optical flow and full sequence information to be given at test time. Our results based on RGB and edge maps are better than those solely using edge maps. We also compare the different CRF versions: 
\begin{itemize}
 \item pre-trained dense-CRF \cite{dense_crf},
 \item our trained CRFasRNN \cite{crfasrnn_iccv2015} model trained per sequence \emph{(CRF-per-seq)},
 \item CRFasRNN \cite{crfasrnn_iccv2015} trained on all frames in the train set of $\mathrm{DAVIS}_{16}$ \emph{(CRF-general)} with sparse labels
\end{itemize}
 
\begin{figure}[ht!]
    \begin{center}
	    \small
	    \begin{tabular}{@{}c@{}c@{}c@{}c@{}c@{}}
	        \includegraphics[width=0.19\textwidth]{1_p_1_00002.pdf}\,
	        \includegraphics[width=0.19\textwidth]{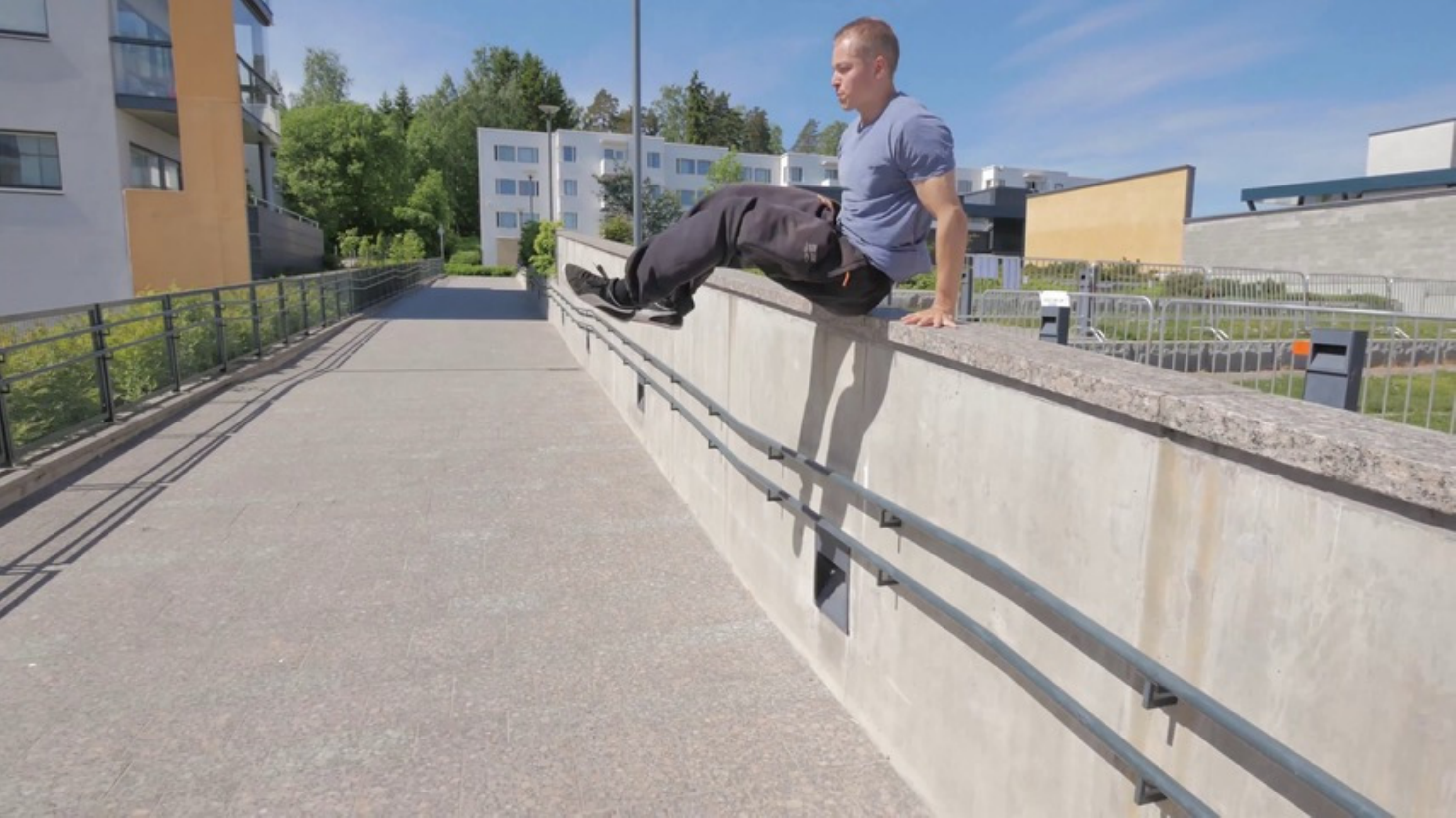}\,
	        \includegraphics[width=0.19\textwidth]{1_p_1_00042.pdf}\,
	        \includegraphics[width=0.19\textwidth]{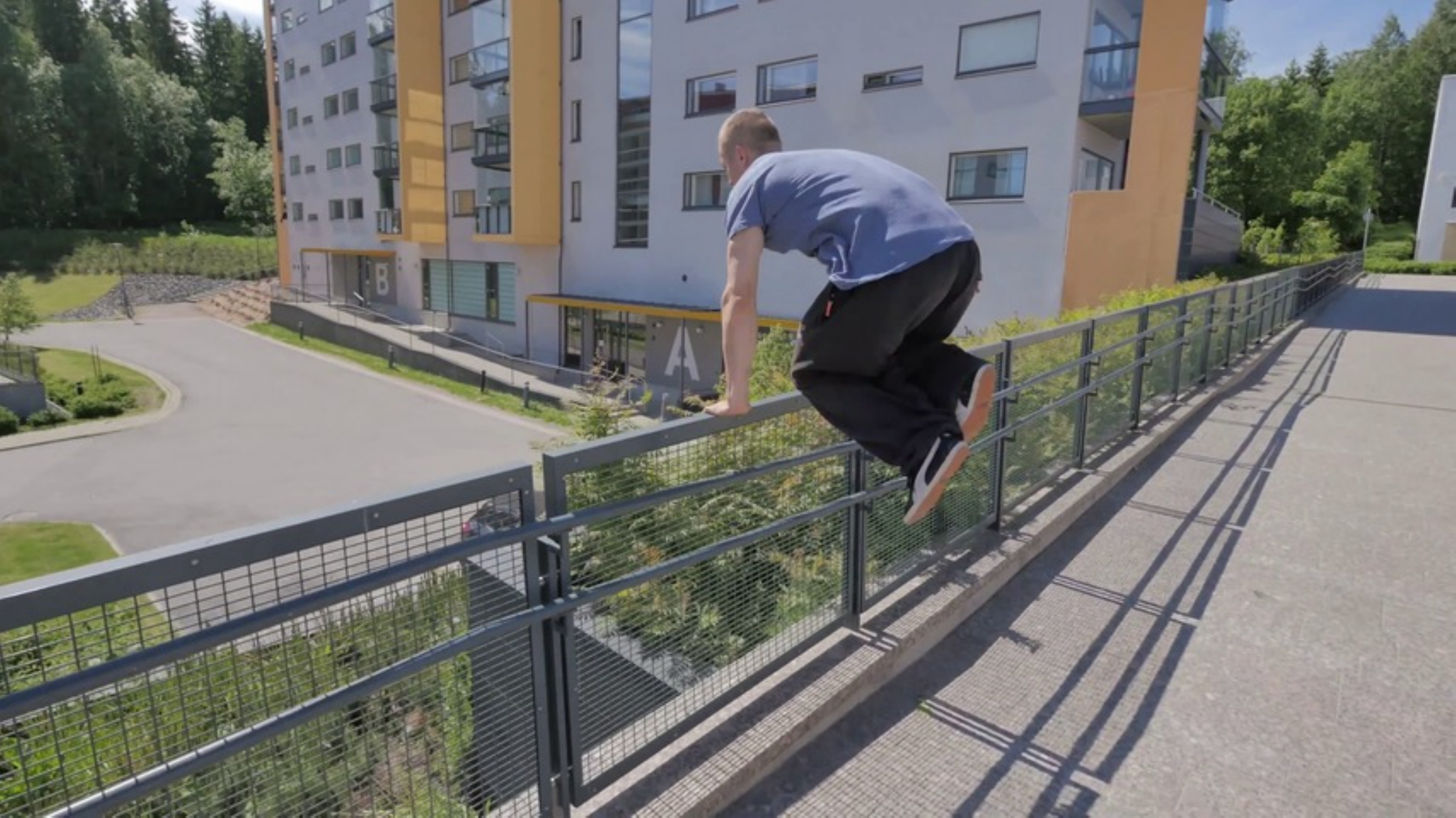}\,
	        \includegraphics[width=0.19\textwidth]{1_p_1_00092.pdf}\,\\
	        \includegraphics[width=0.19\textwidth]{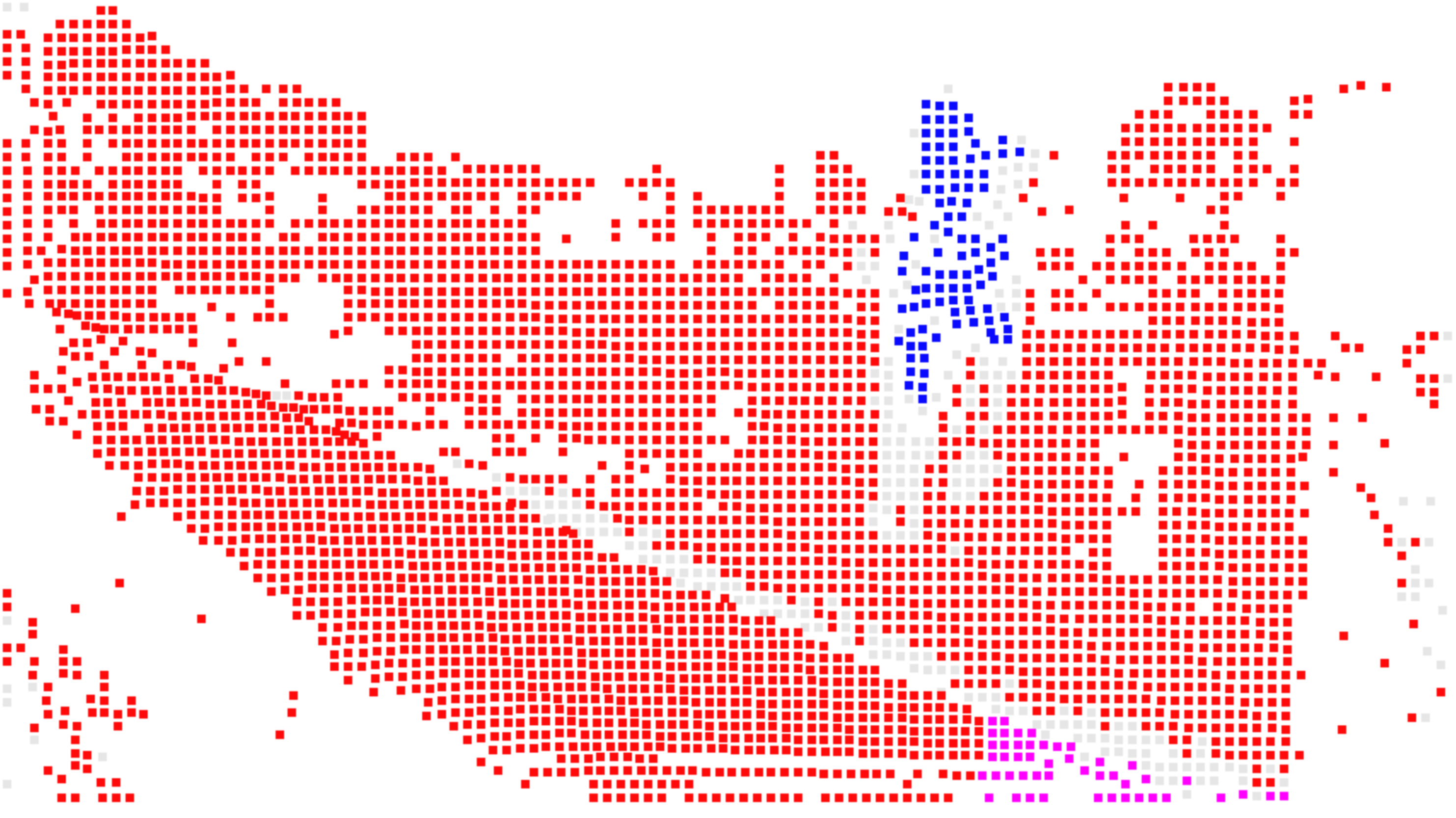}\,
	        \includegraphics[width=0.19\textwidth]{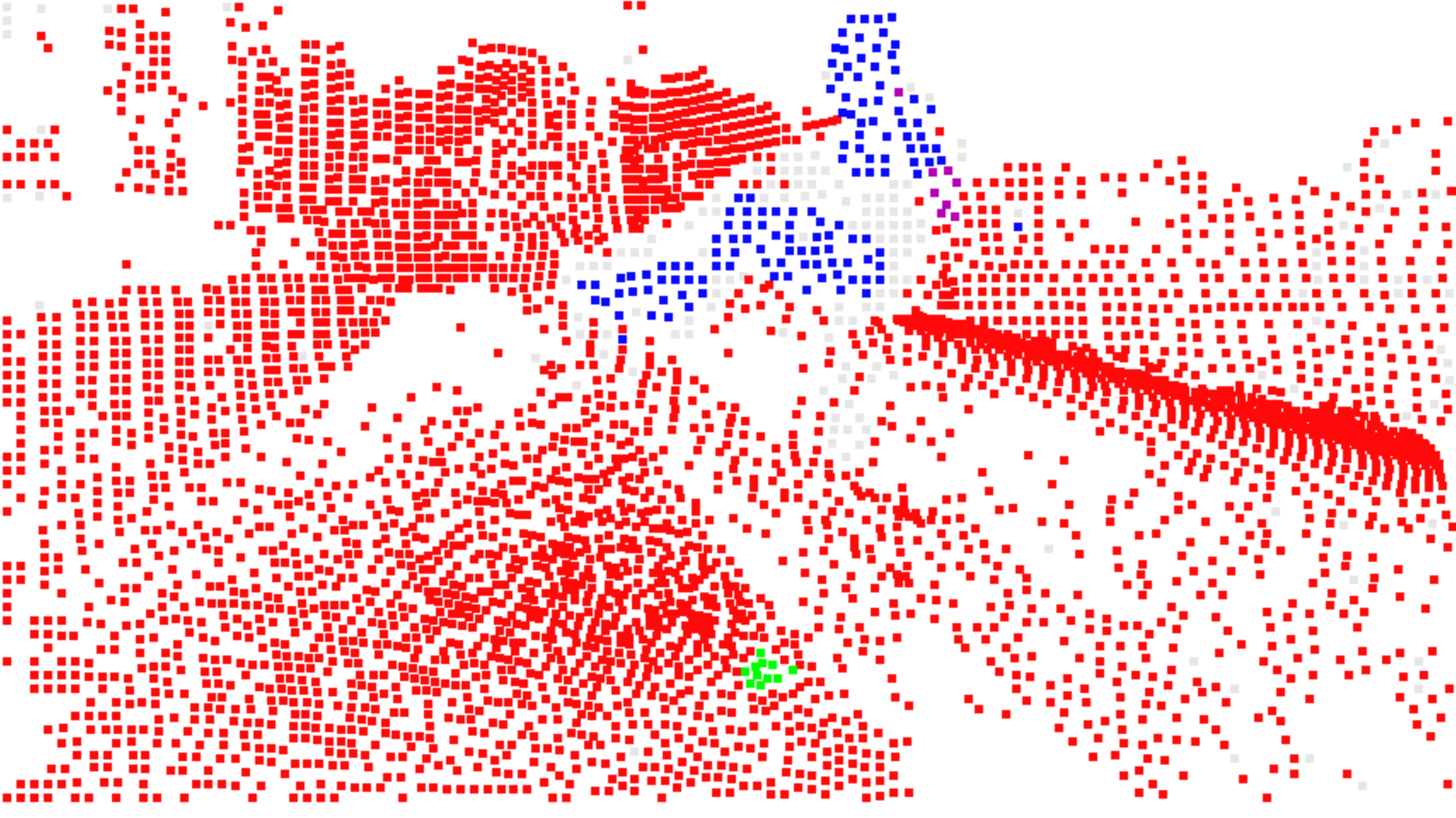}\,
	        \includegraphics[width=0.19\textwidth]{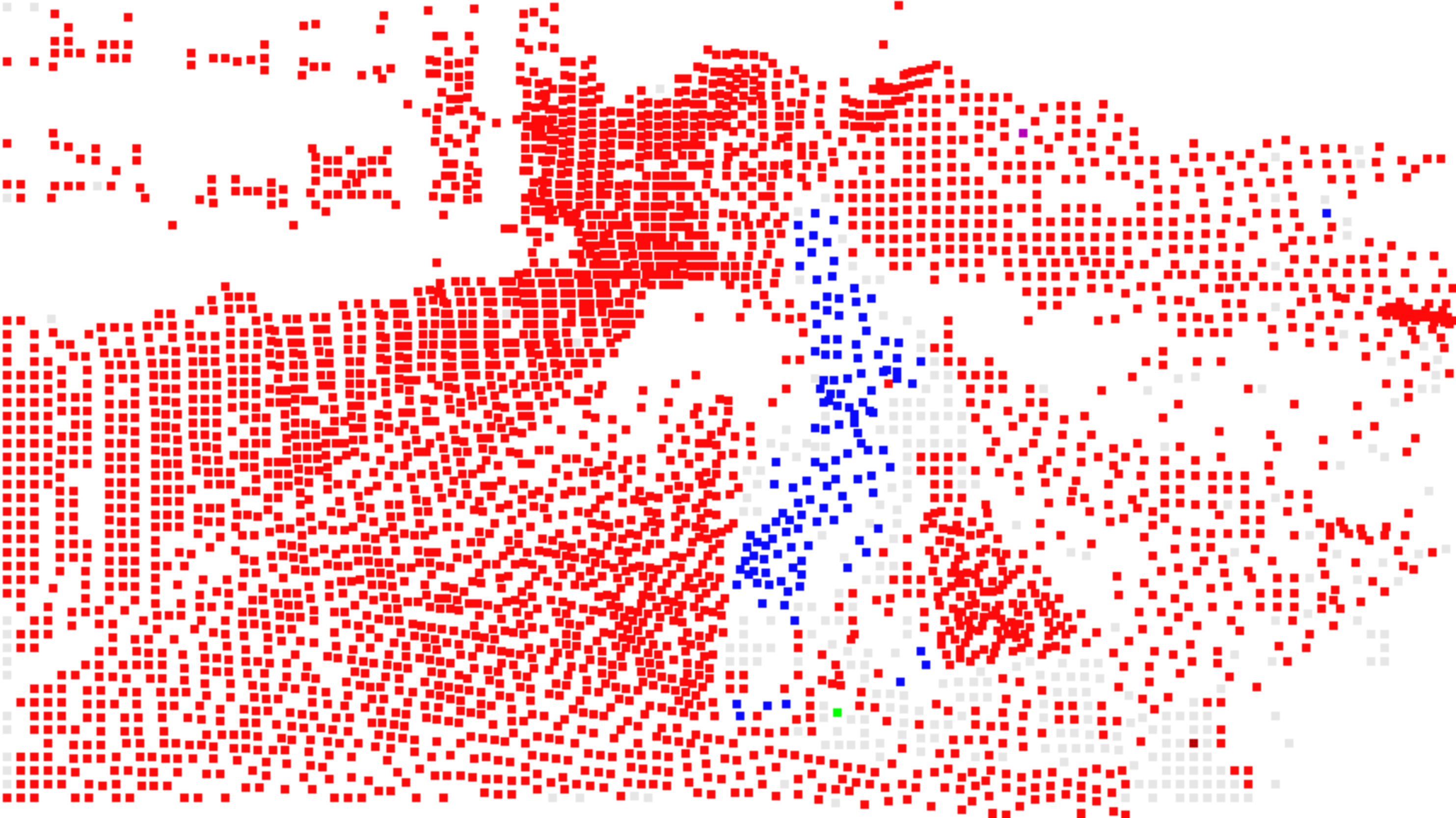}\,
	        \includegraphics[width=0.19\textwidth]{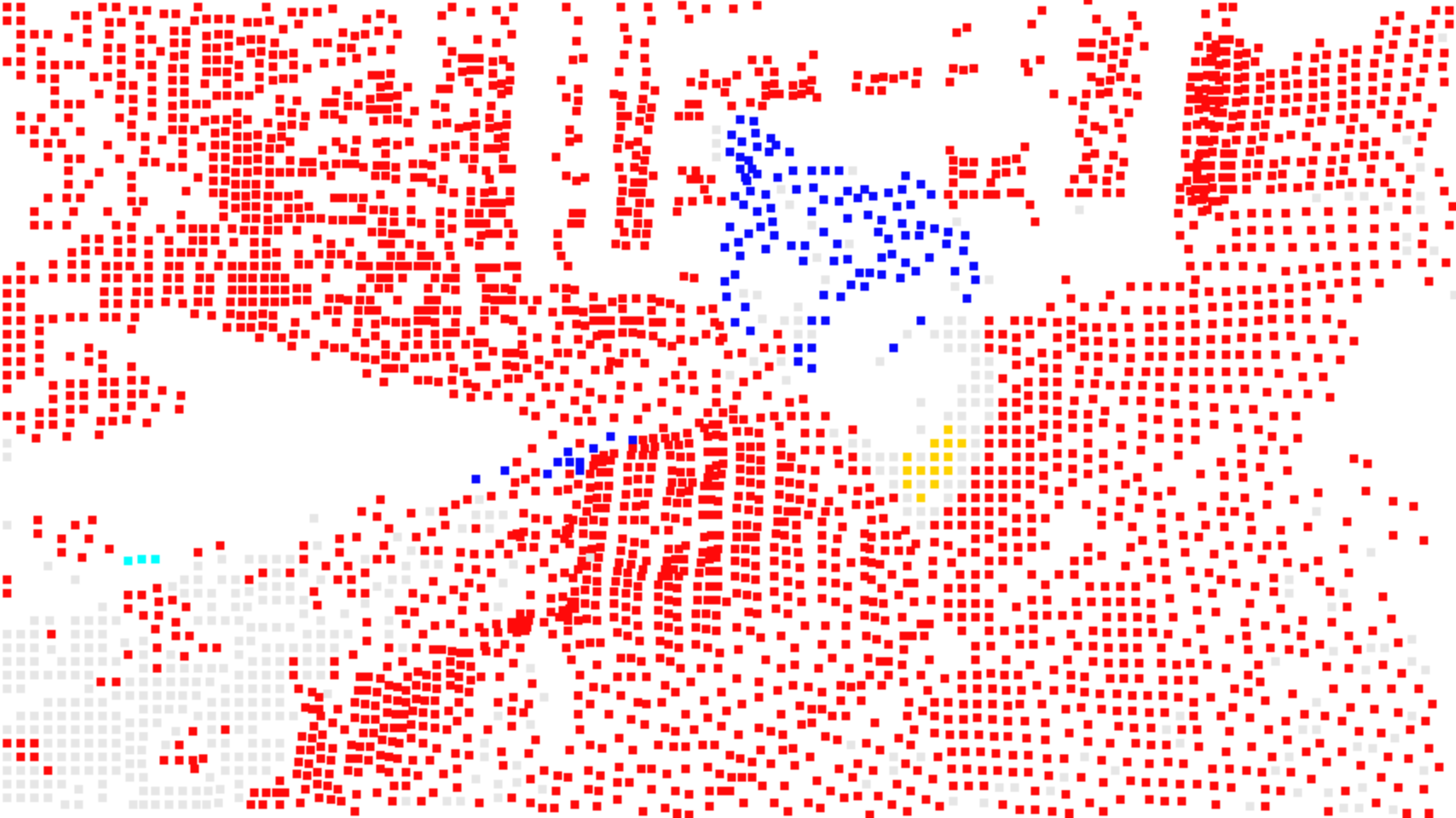}\,
	        \includegraphics[width=0.19\textwidth]{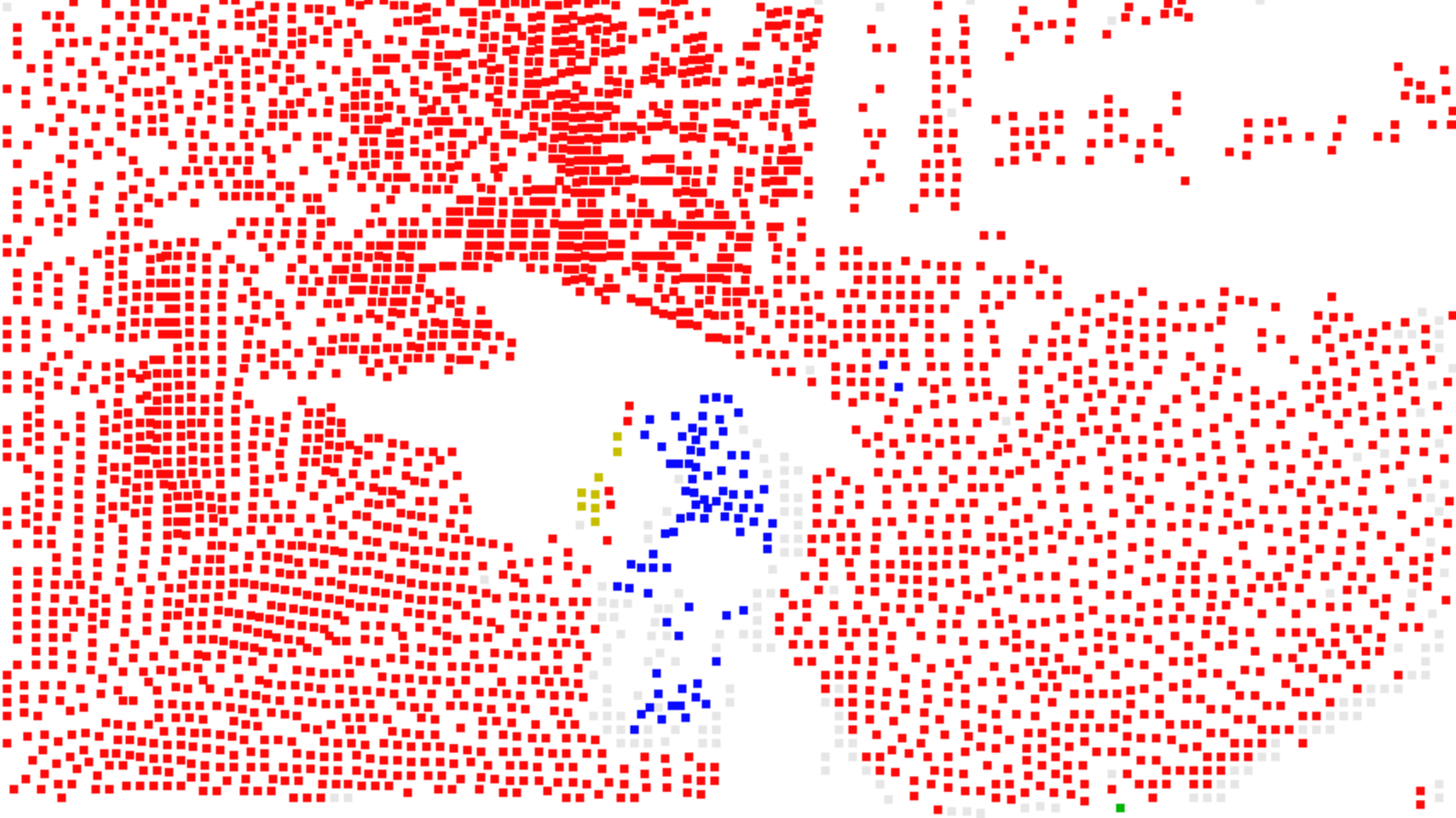}\,\\
	        \includegraphics[width=0.19\textwidth]{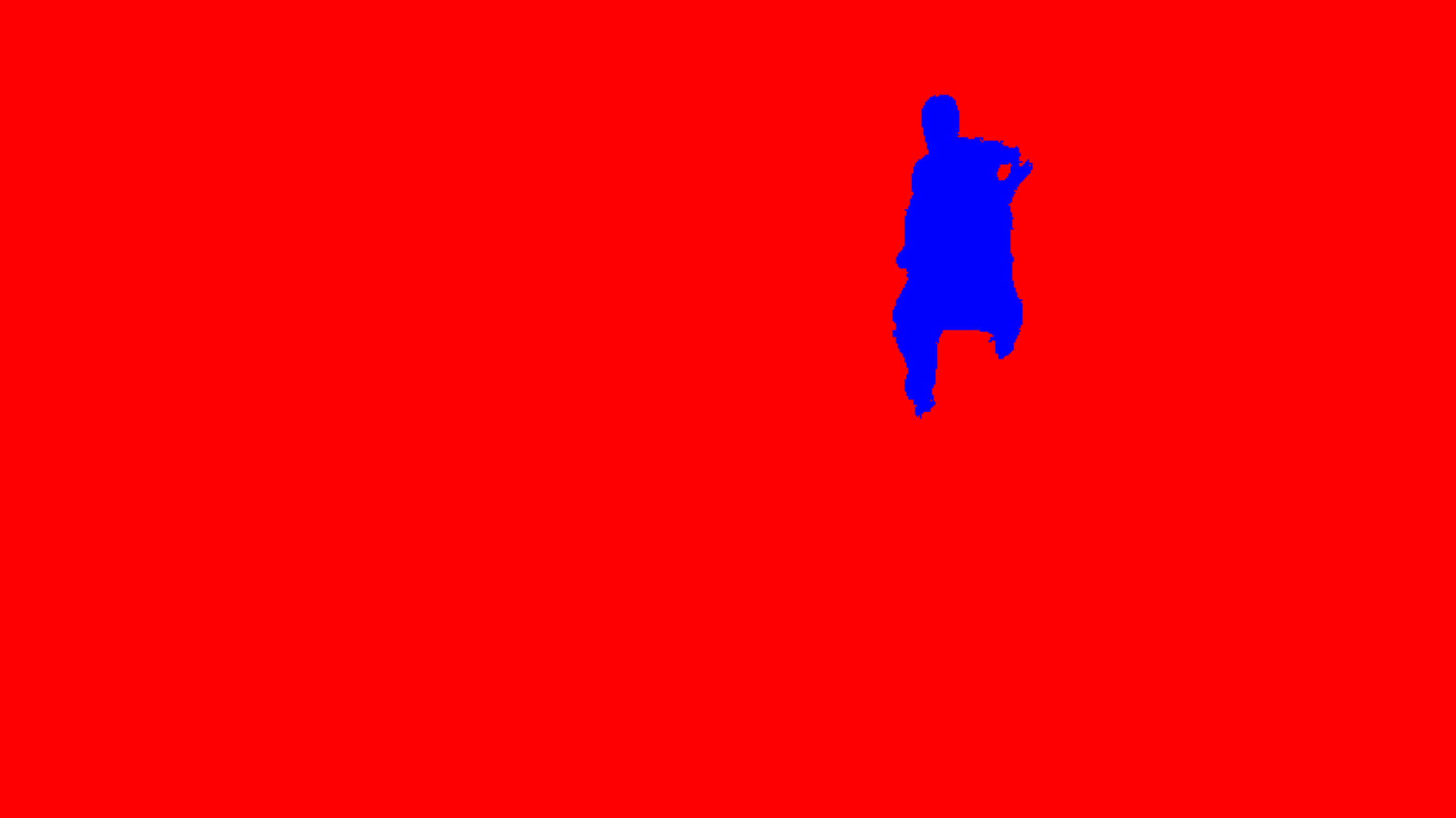}\,
	        \includegraphics[width=0.19\textwidth]{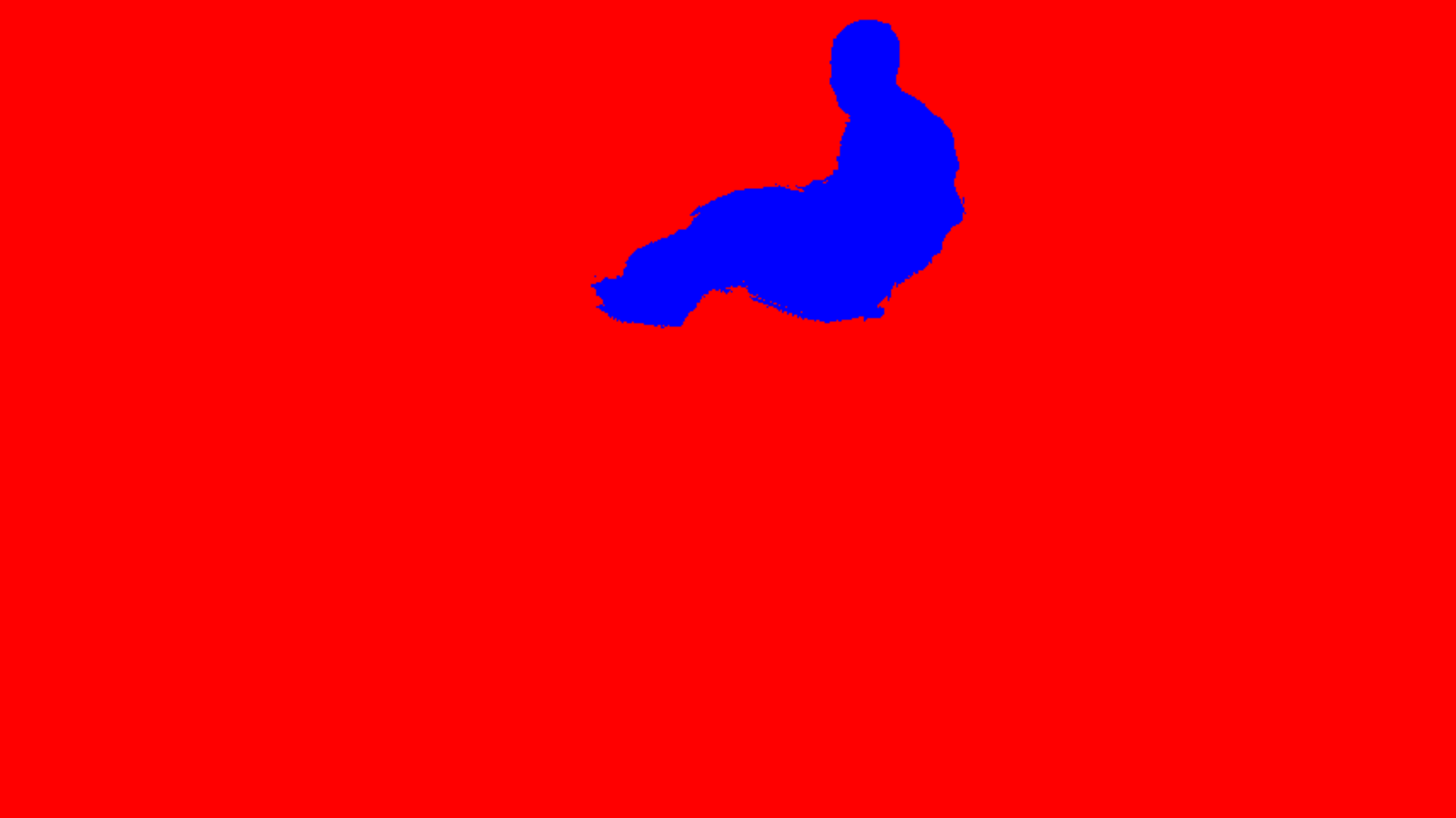}\,
	        \includegraphics[width=0.19\textwidth]{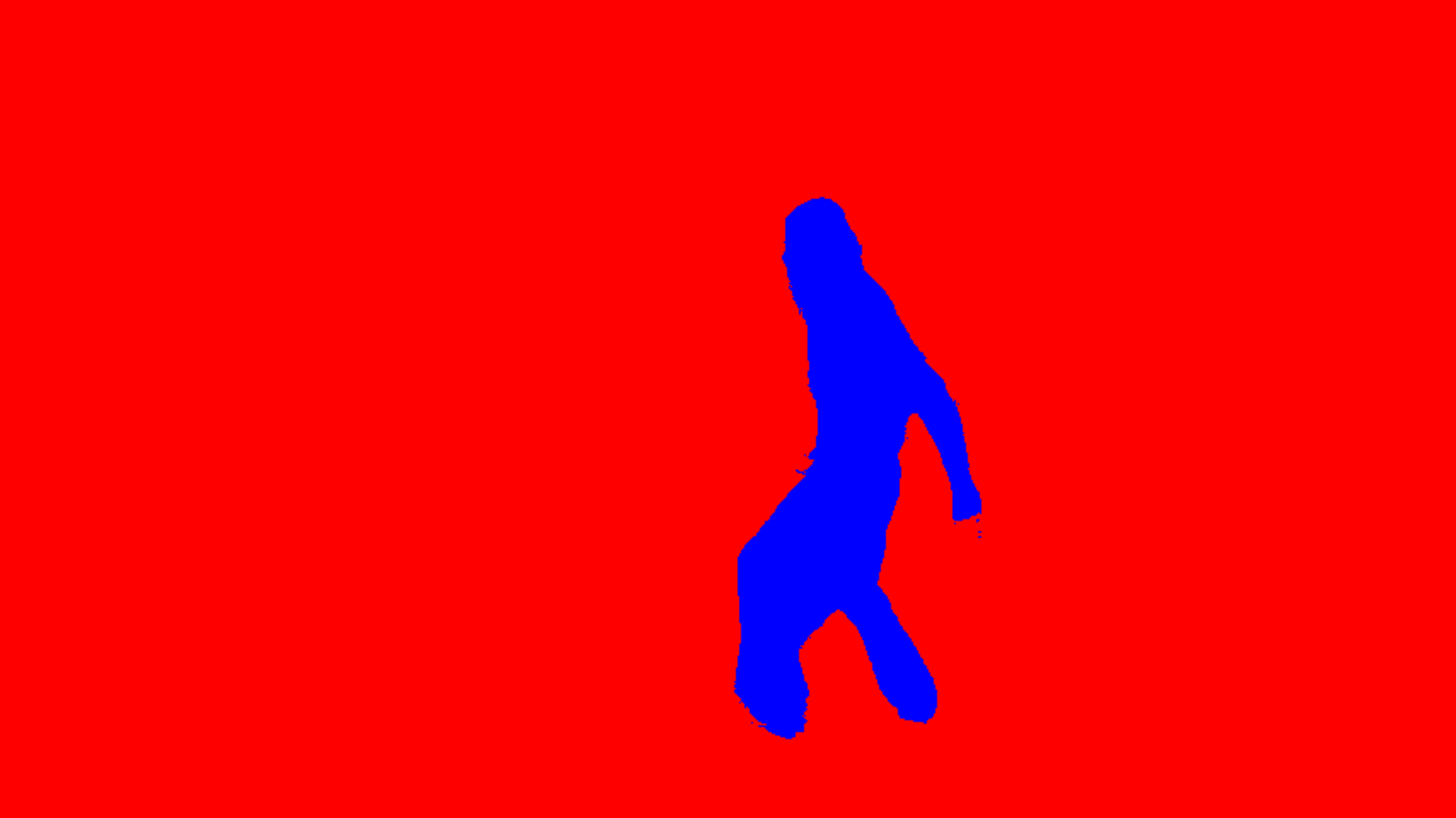}\,
	        \includegraphics[width=0.19\textwidth]{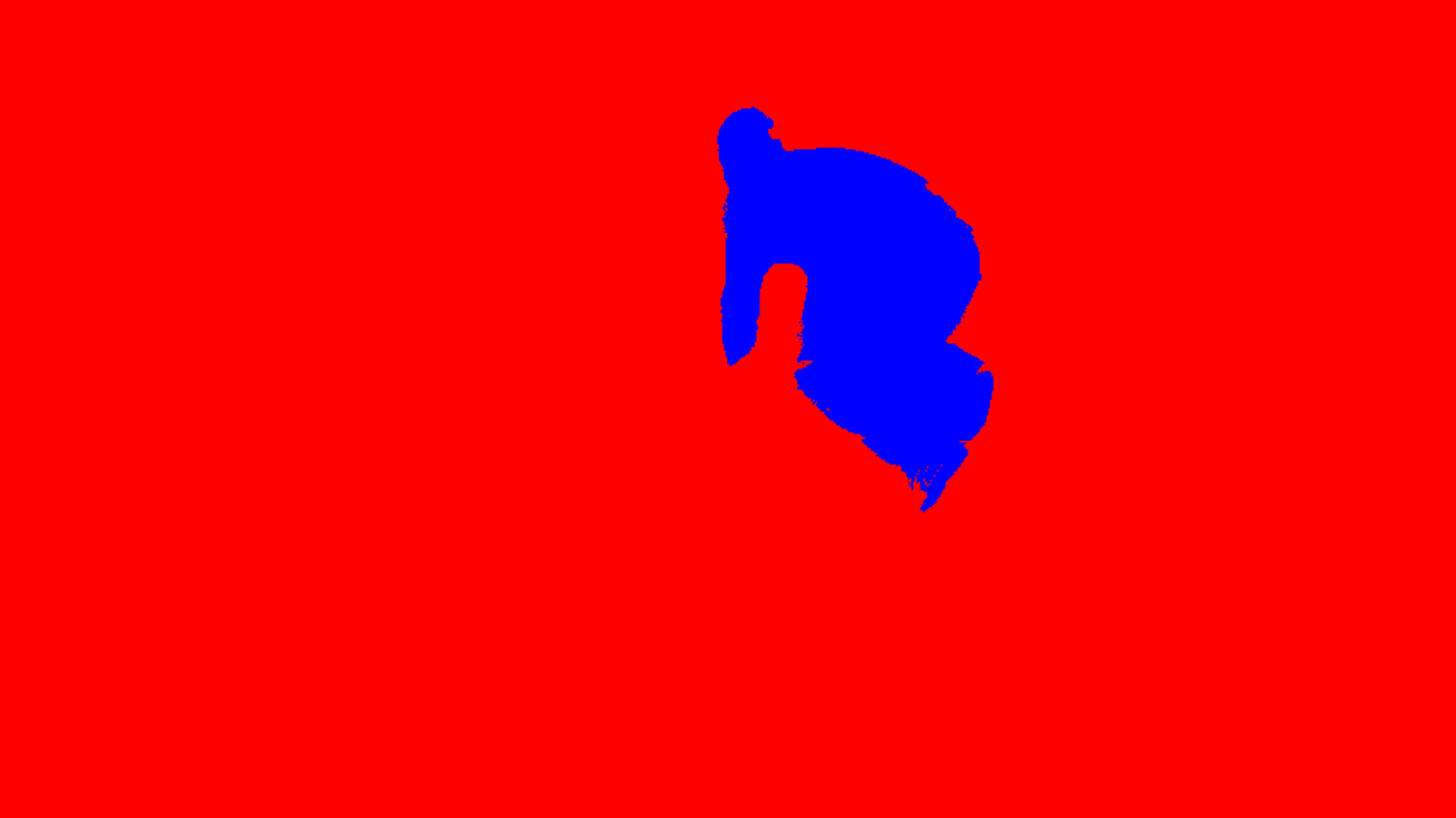}\,
	        \includegraphics[width=0.19\textwidth]{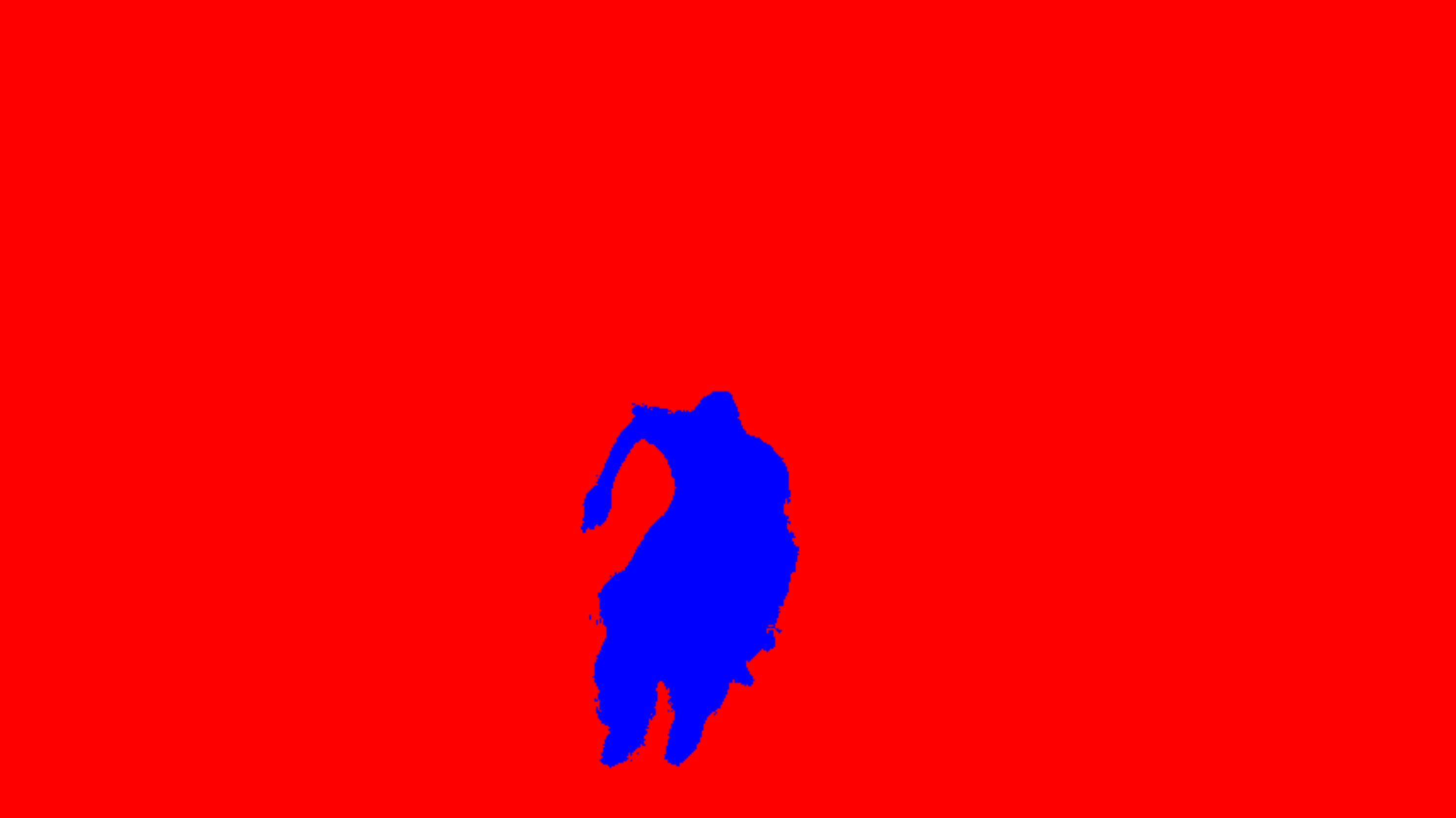}\,\\
	        
	        \includegraphics[width=0.19\textwidth]{1_d_1_00000.pdf}\,
	        \includegraphics[width=0.19\textwidth]{1_d_1_00014.pdf}\,
	        \includegraphics[width=0.19\textwidth]{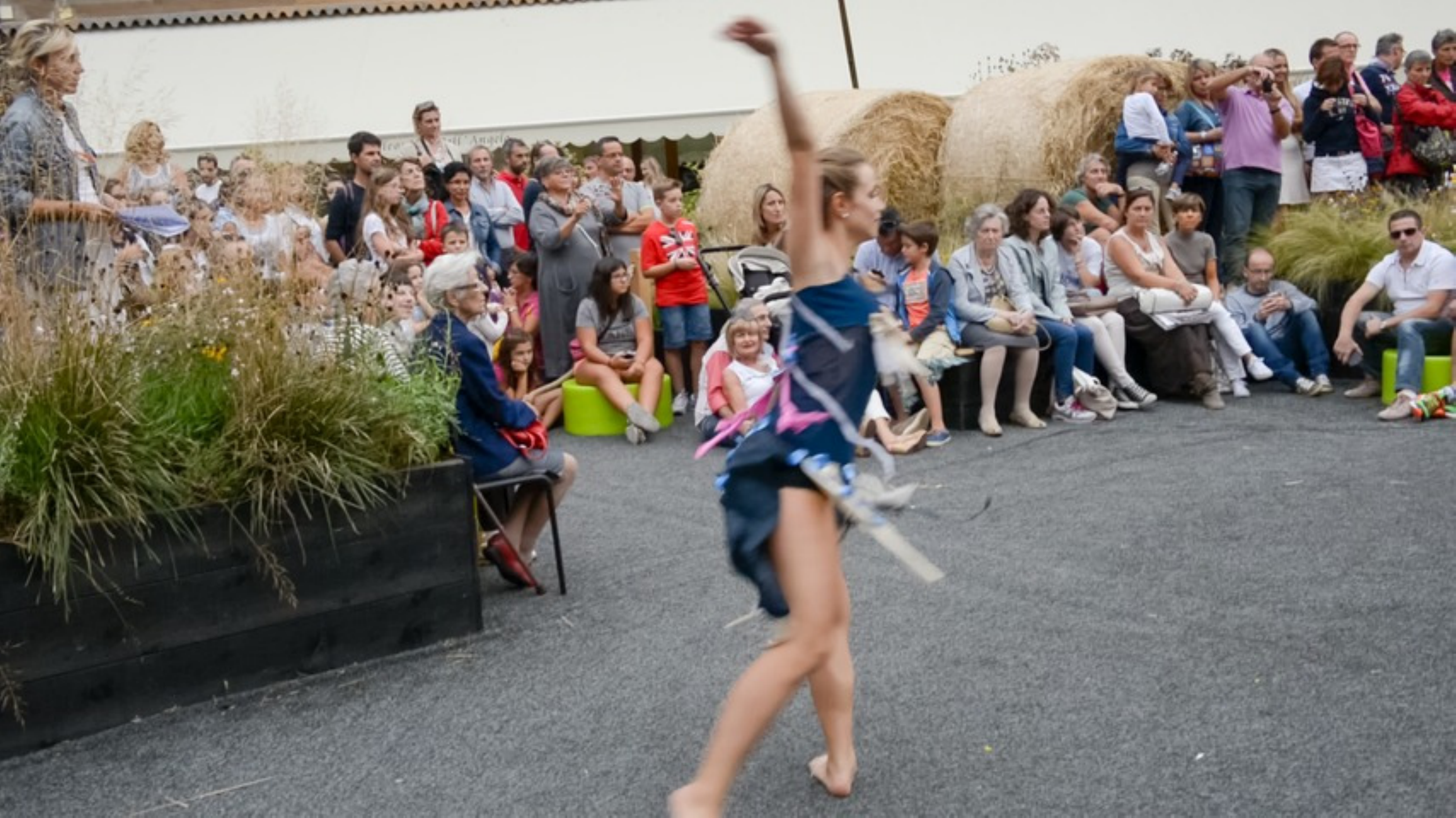}\,
	        \includegraphics[width=0.19\textwidth]{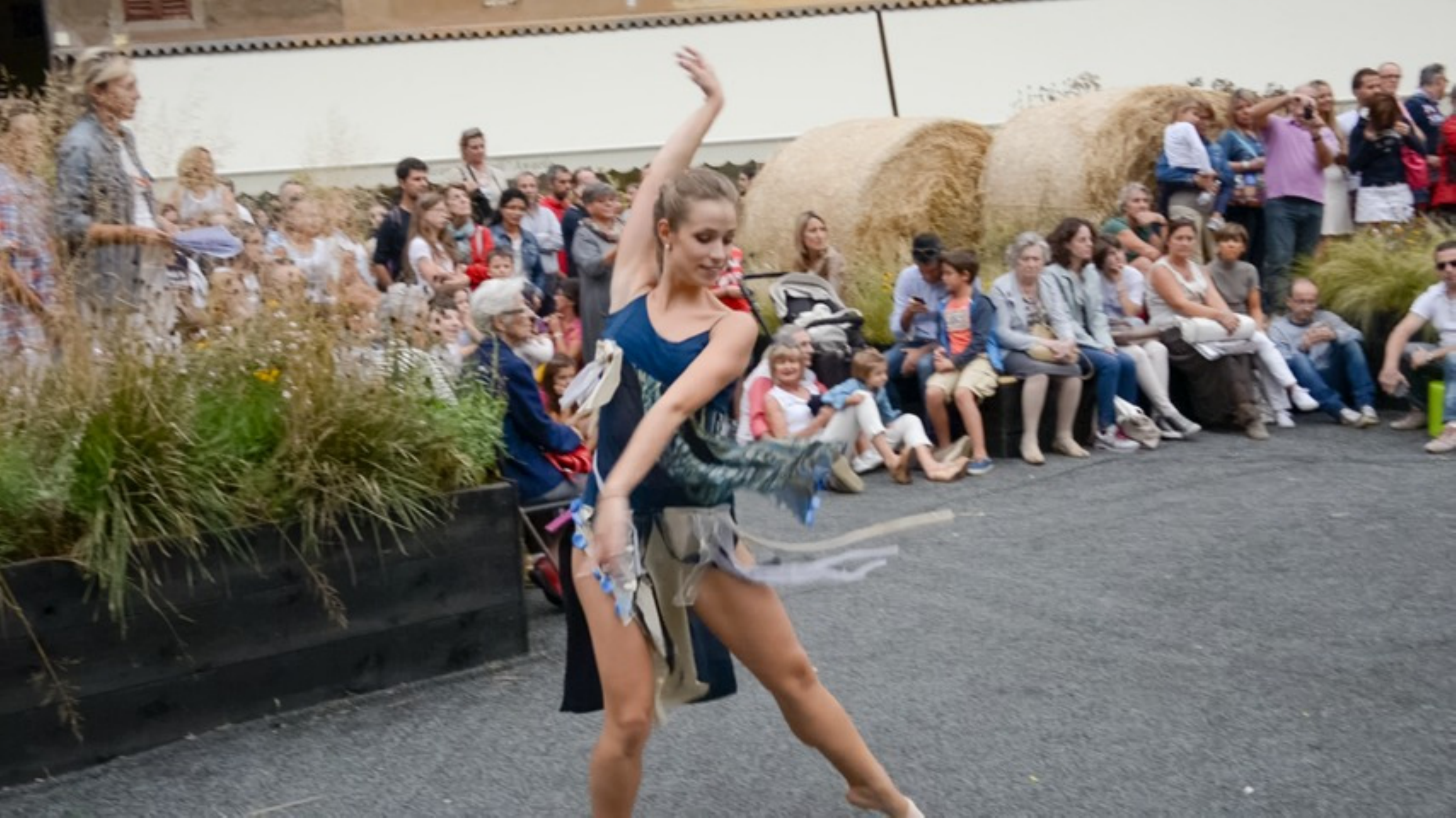}\,
	        \includegraphics[width=0.19\textwidth]{1_d_1_00088.pdf}\,\\
	        \includegraphics[width=0.19\textwidth]{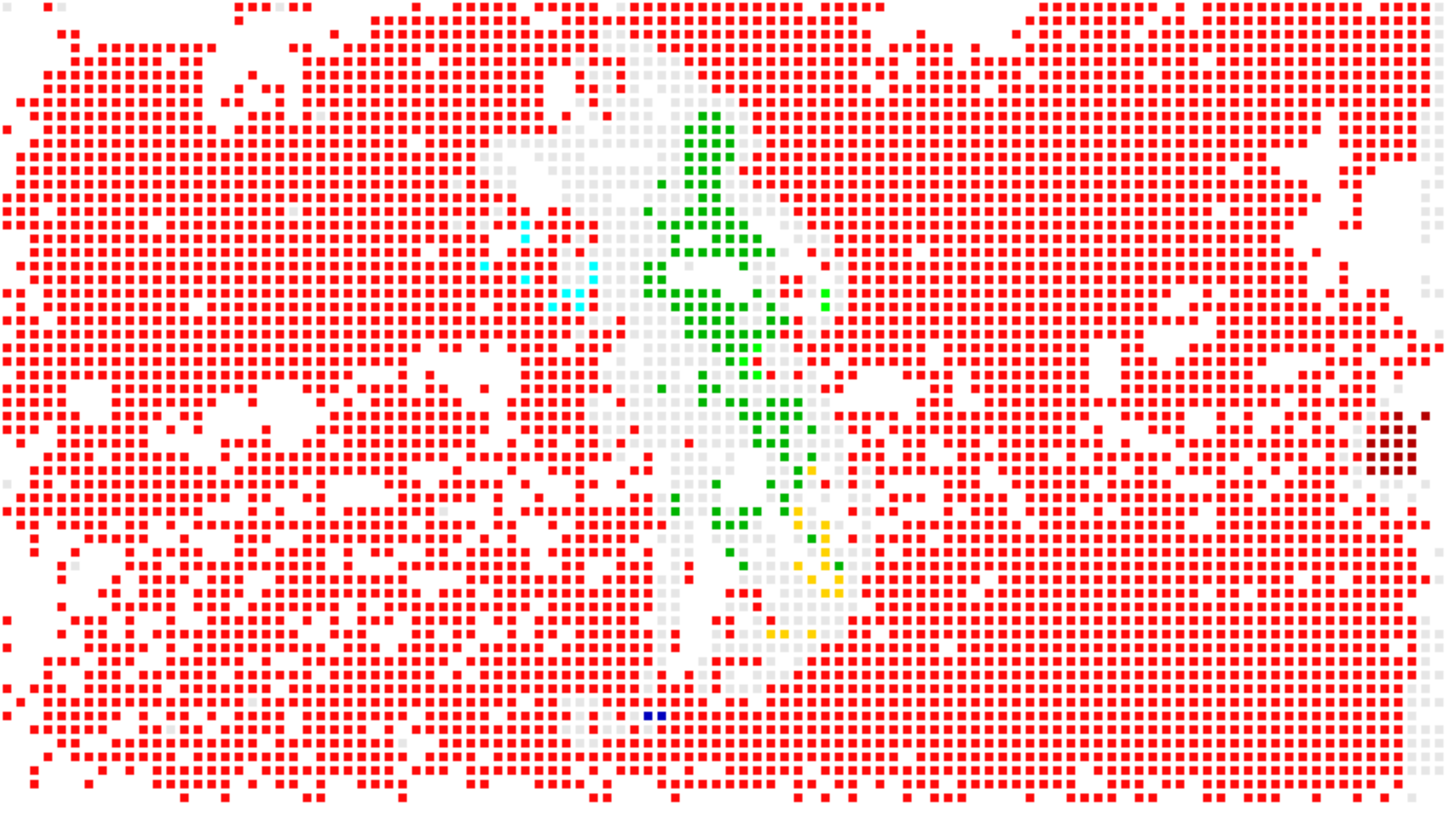}\,
	        \includegraphics[width=0.19\textwidth]{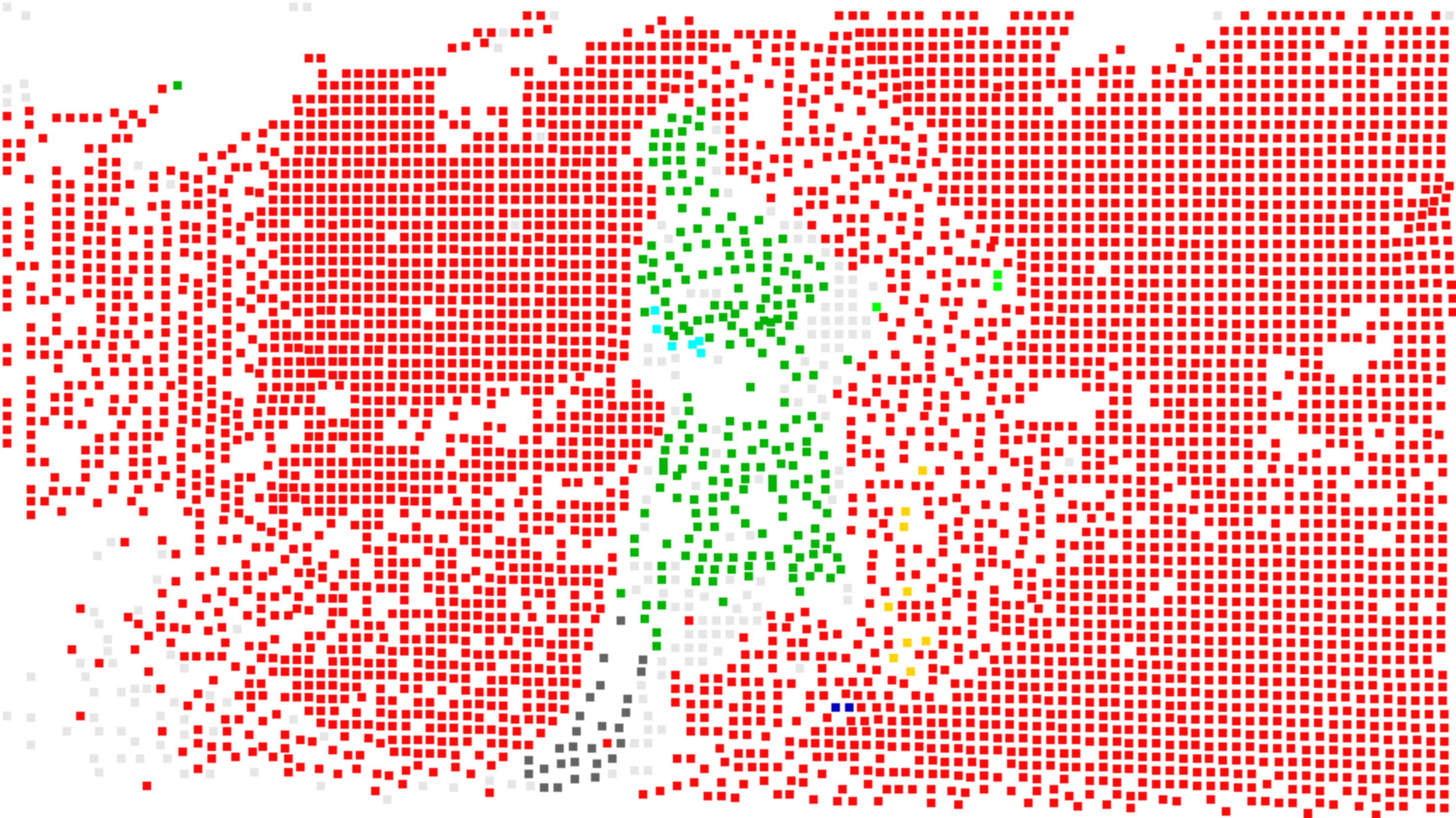}\,
	        \includegraphics[width=0.19\textwidth]{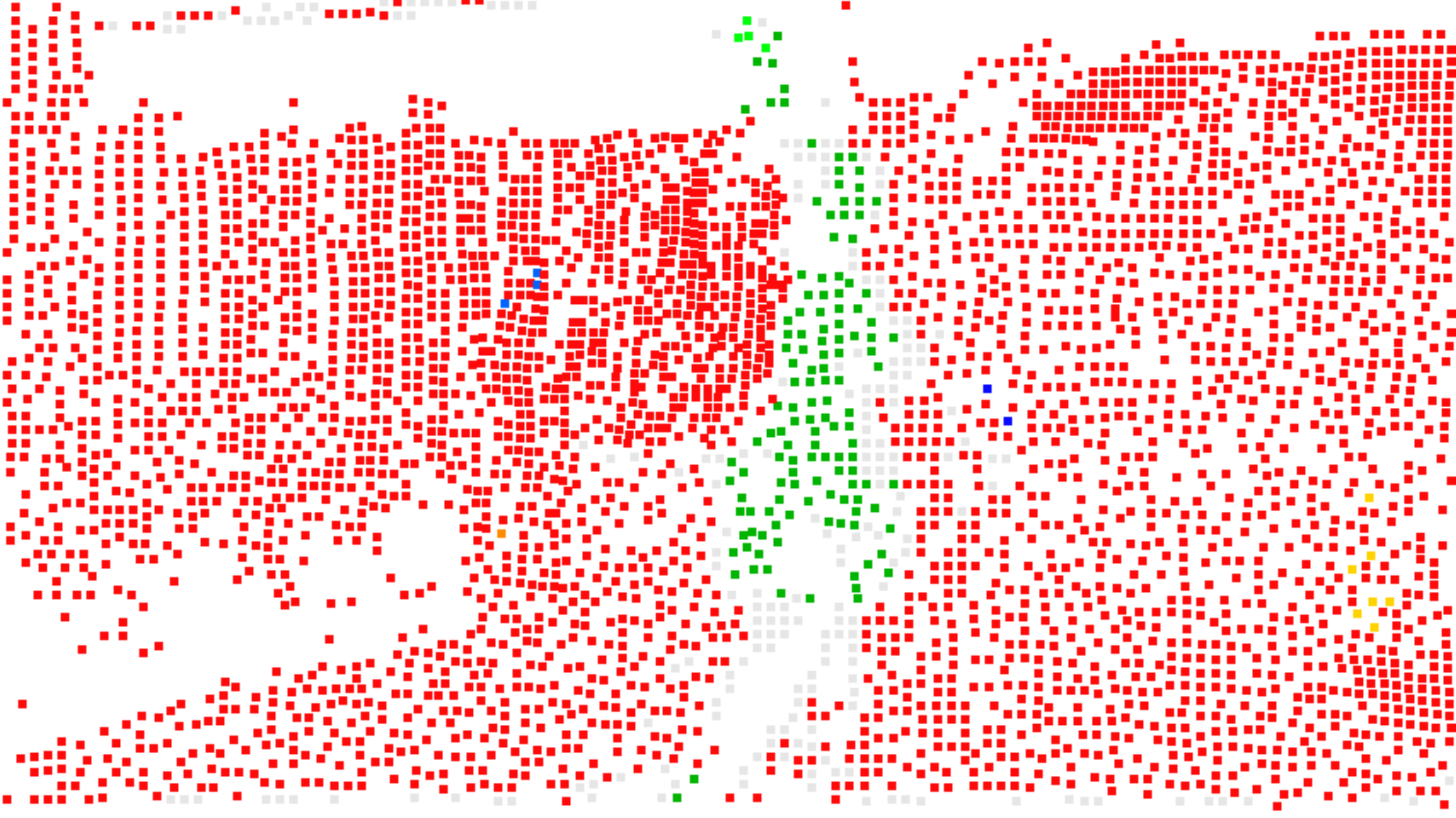}\,
	        \includegraphics[width=0.19\textwidth]{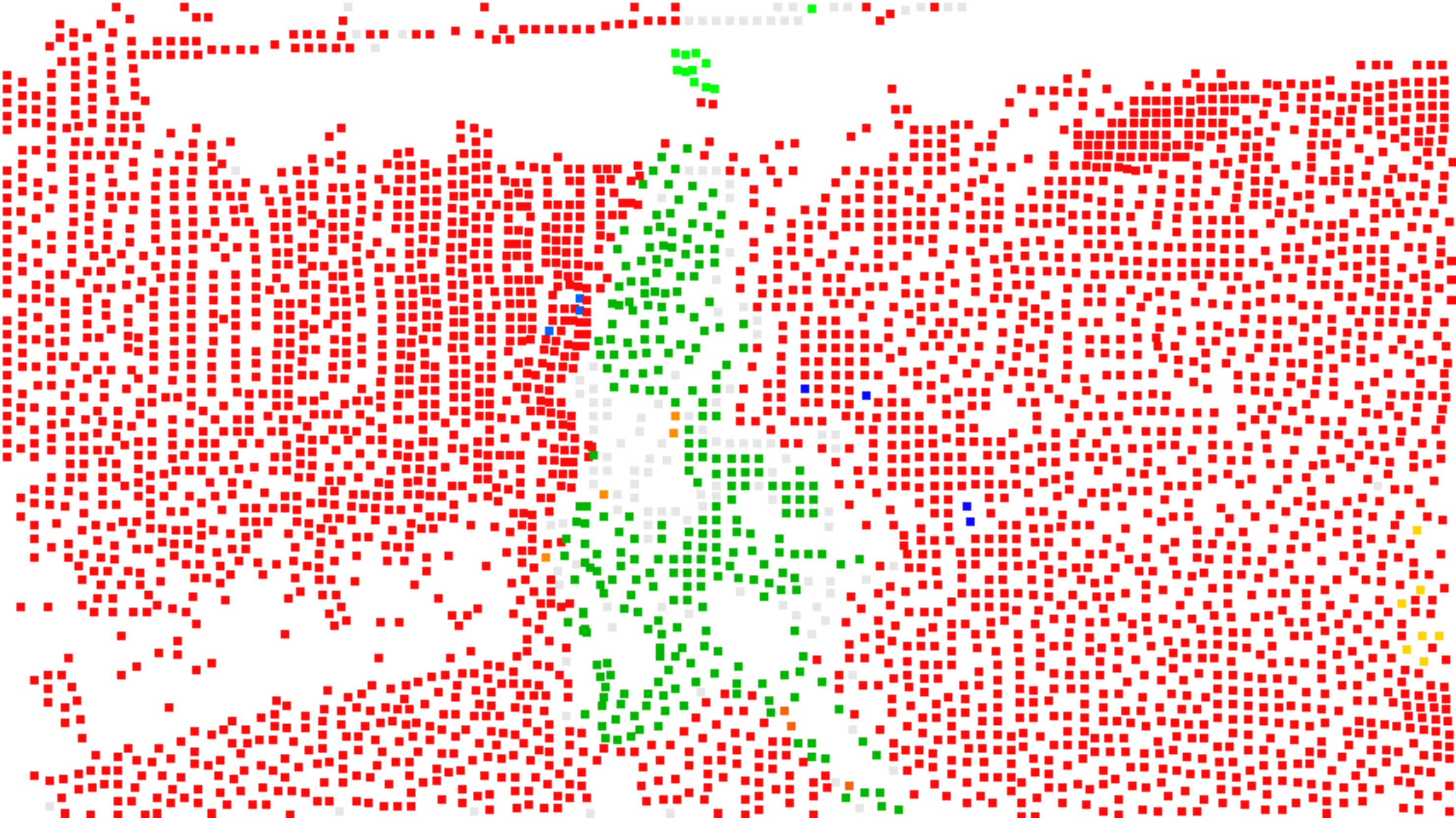}\,
	        \includegraphics[width=0.19\textwidth]{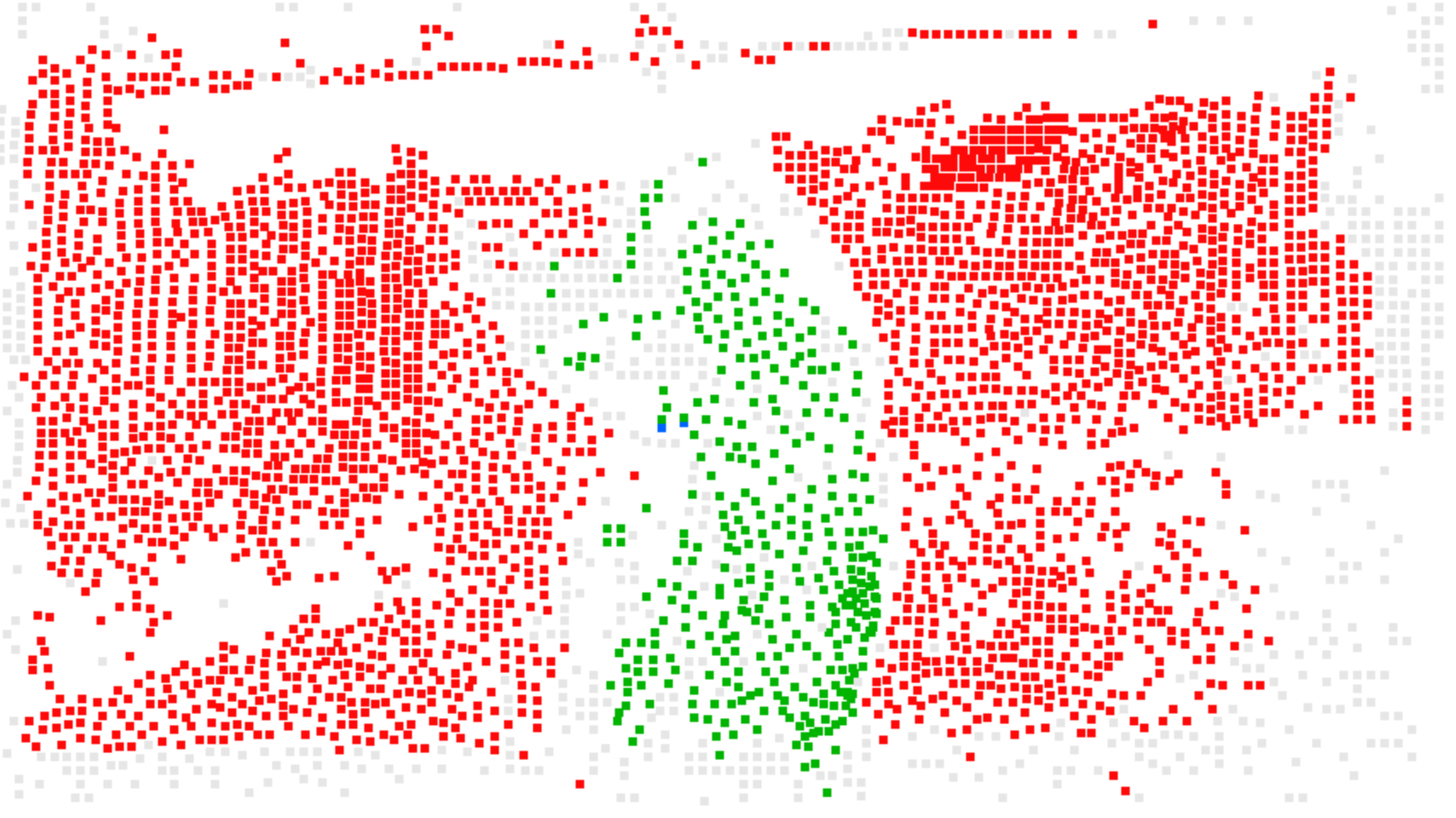}\,\\
	        \includegraphics[width=0.19\textwidth]{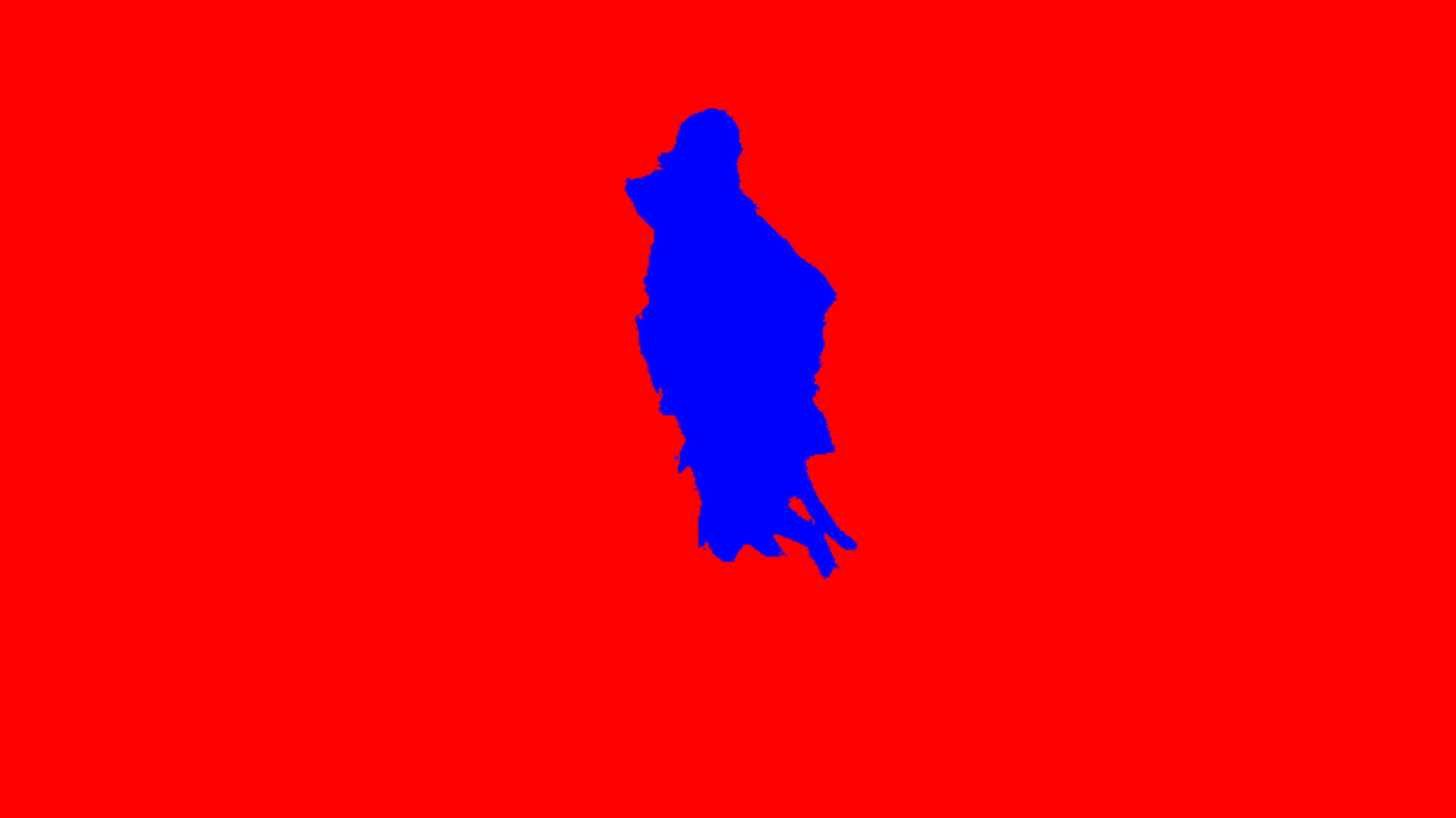}\,
	        \includegraphics[width=0.19\textwidth]{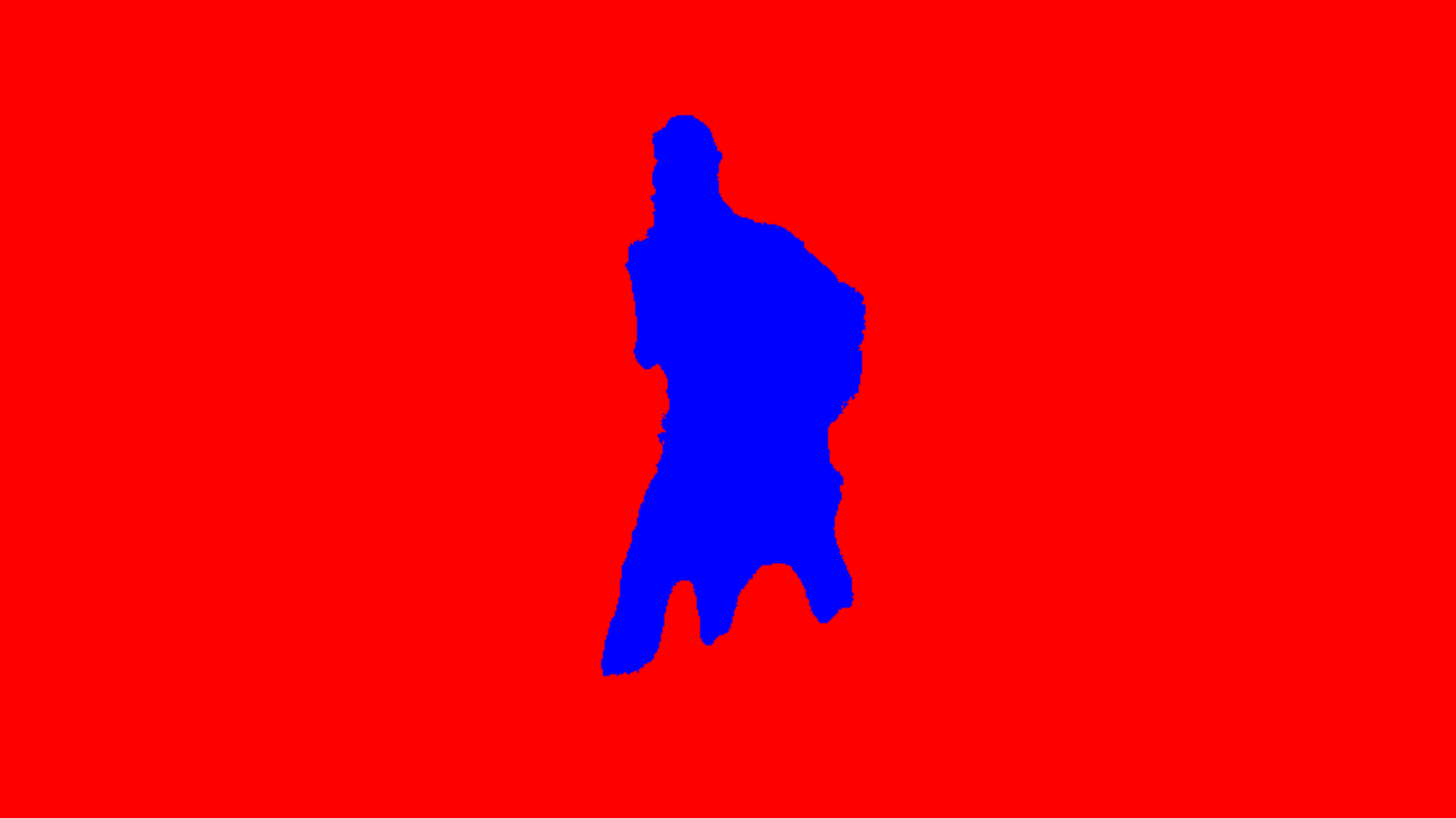}\,
	        \includegraphics[width=0.19\textwidth]{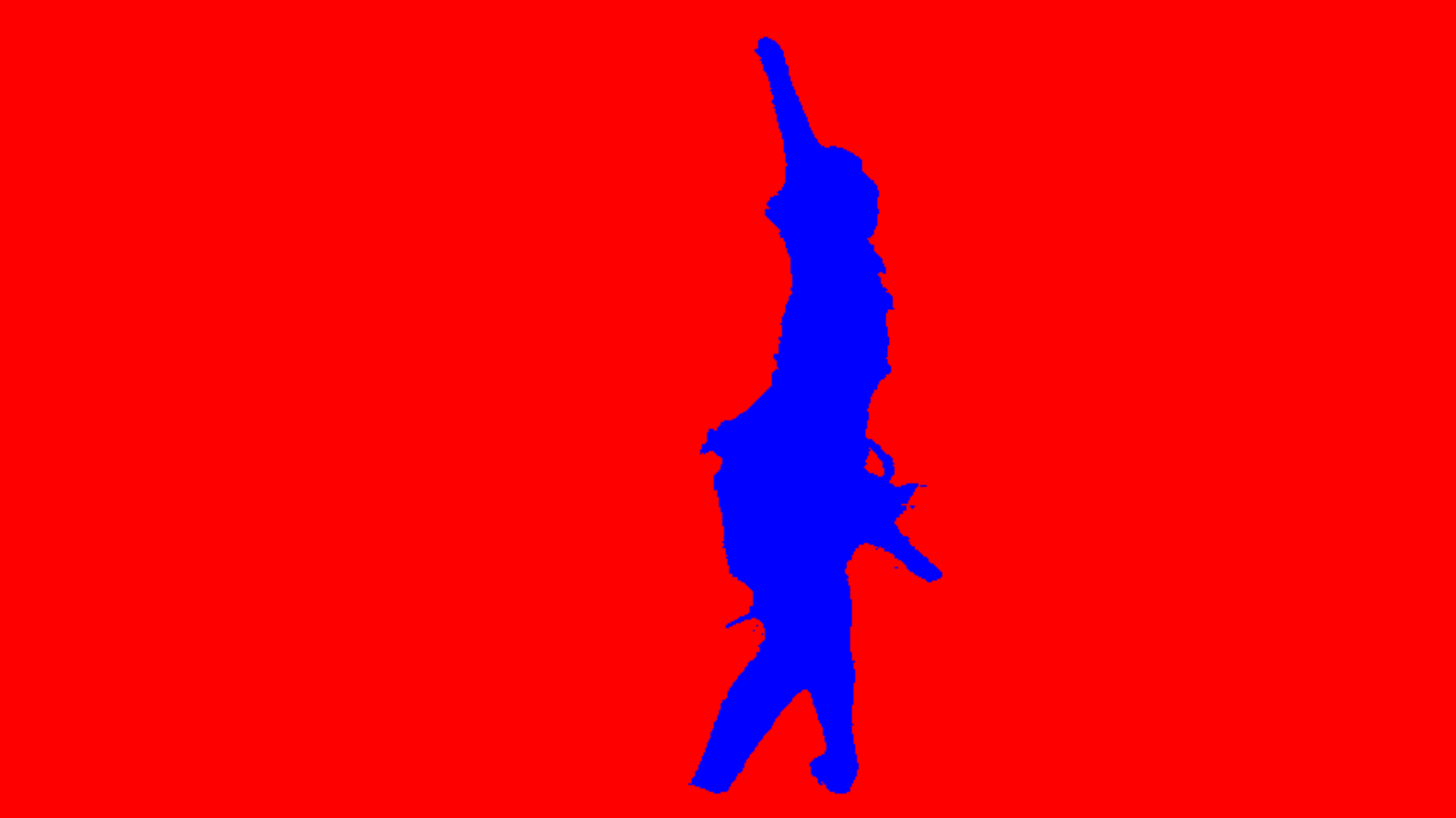}\,
	        \includegraphics[width=0.19\textwidth]{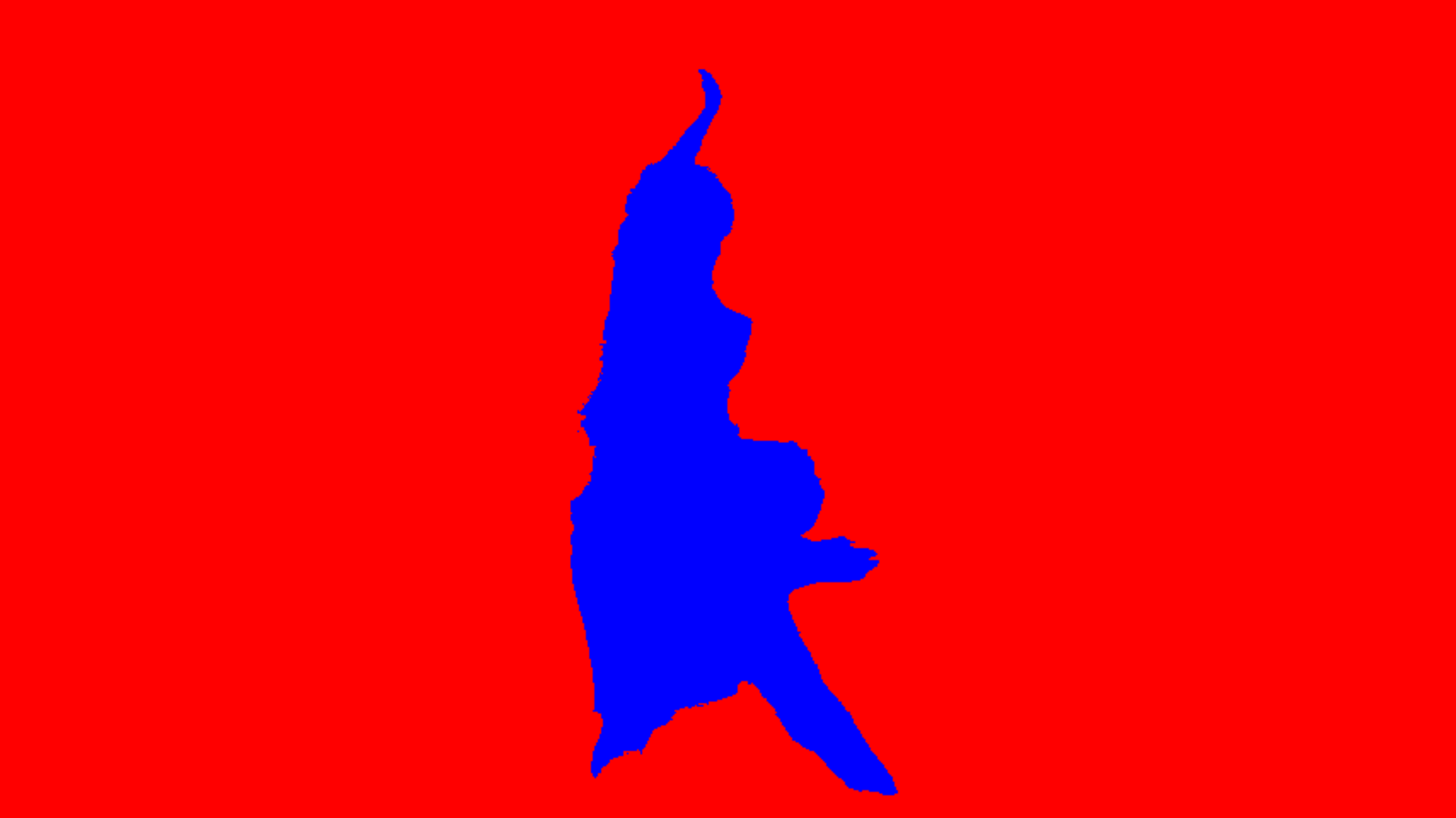}\,
	        \includegraphics[width=0.19\textwidth]{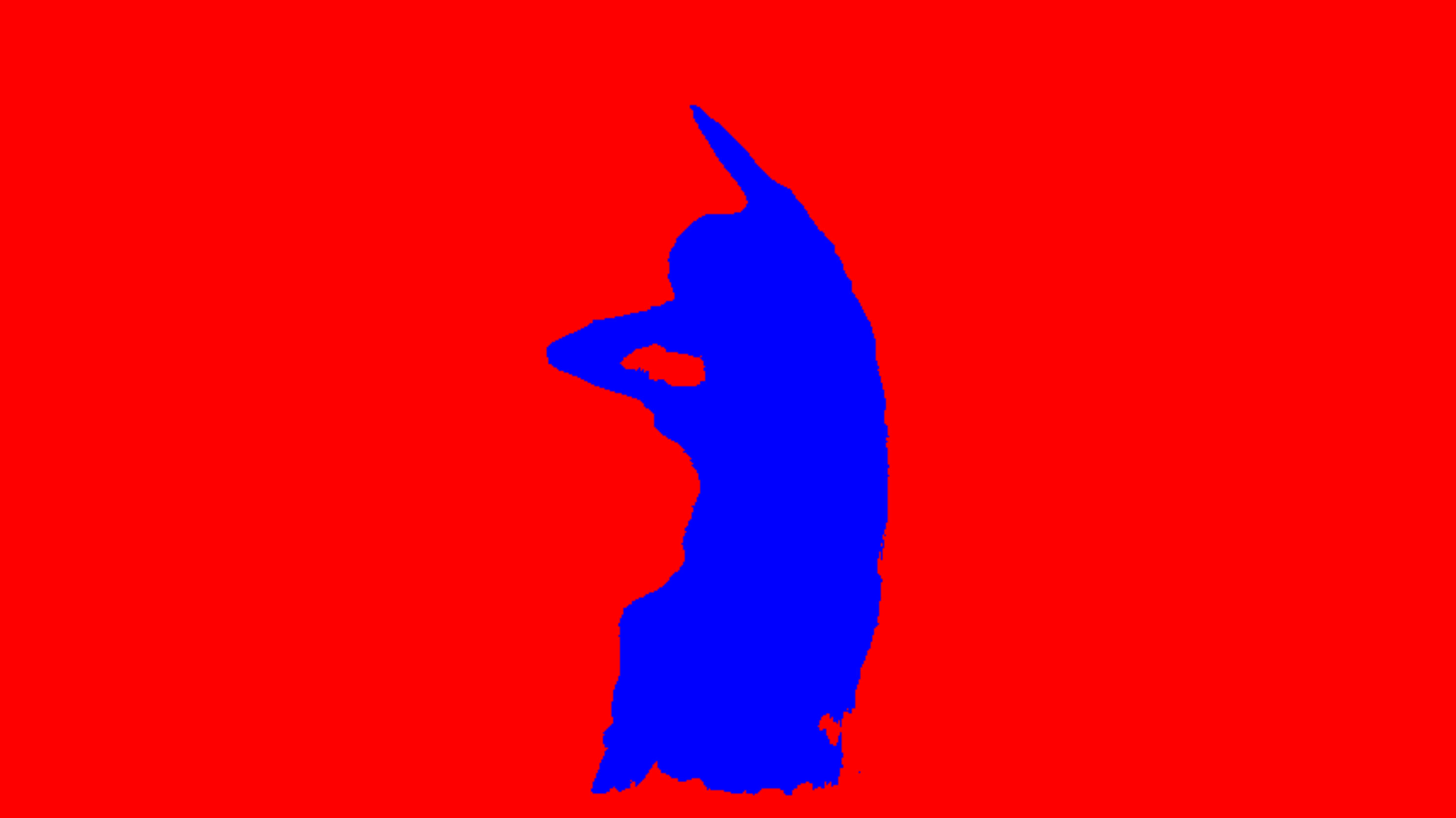}\,

        \end{tabular}
    \caption{Exemplary single-label motion segmentation results showing the five frames and their sparse and dense segmentation for two different sequences, generated using the proposed U-Net model. 
    The images are from the sequences on the validation set of $\mathrm{DAVIS}_{16}$ \cite{davis_16} dataset.}
    \label{examples_fig_2}
    \end{center}
\end{figure}

\begin{figure}[t!]
    \begin{center}
	    \small
	    \begin{tabular}{@{}c@{}c@{}c@{}c@{}c@{}}
            \includegraphics[width=0.19\textwidth]{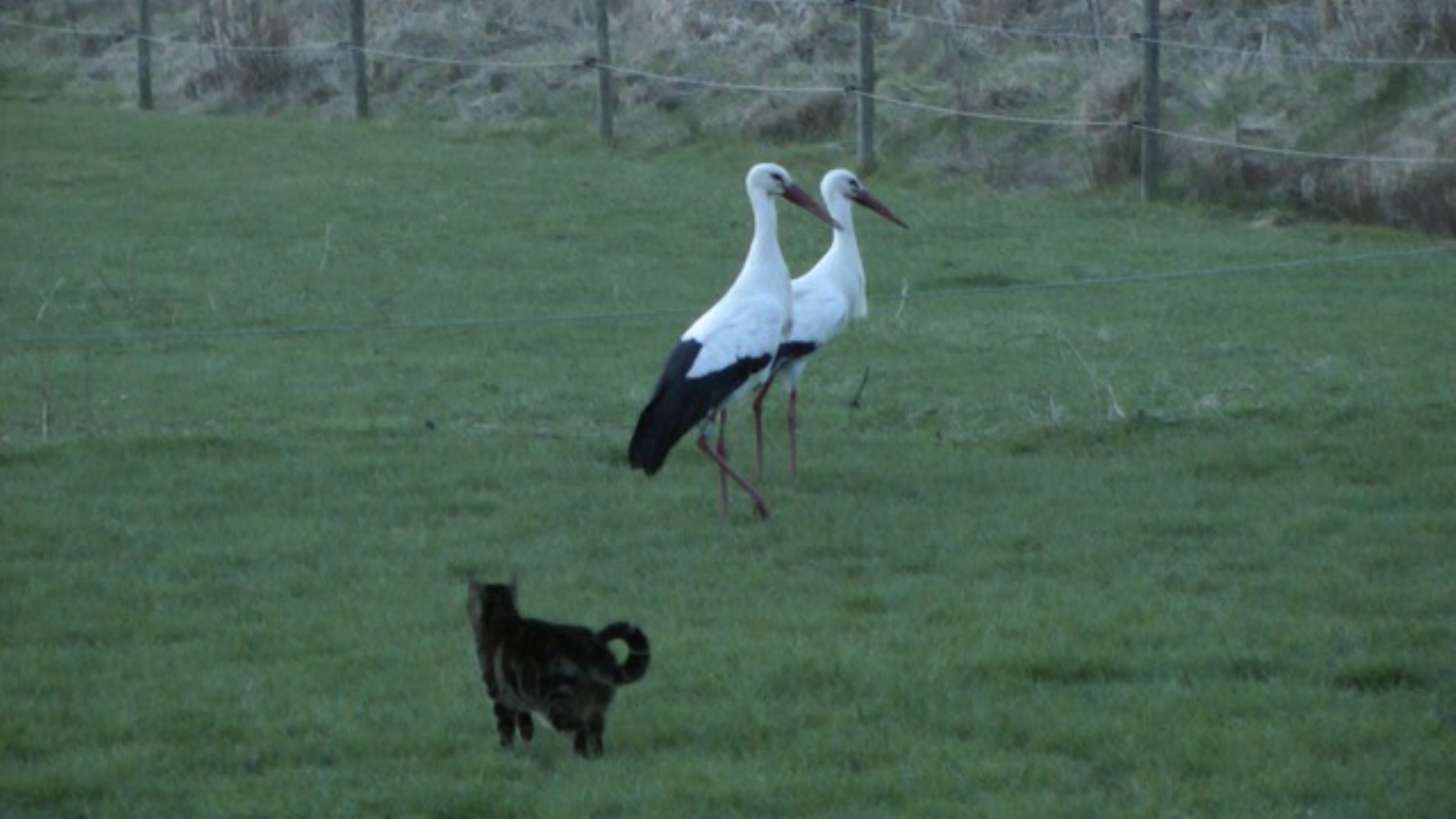}\,
            \includegraphics[width=0.19\textwidth]{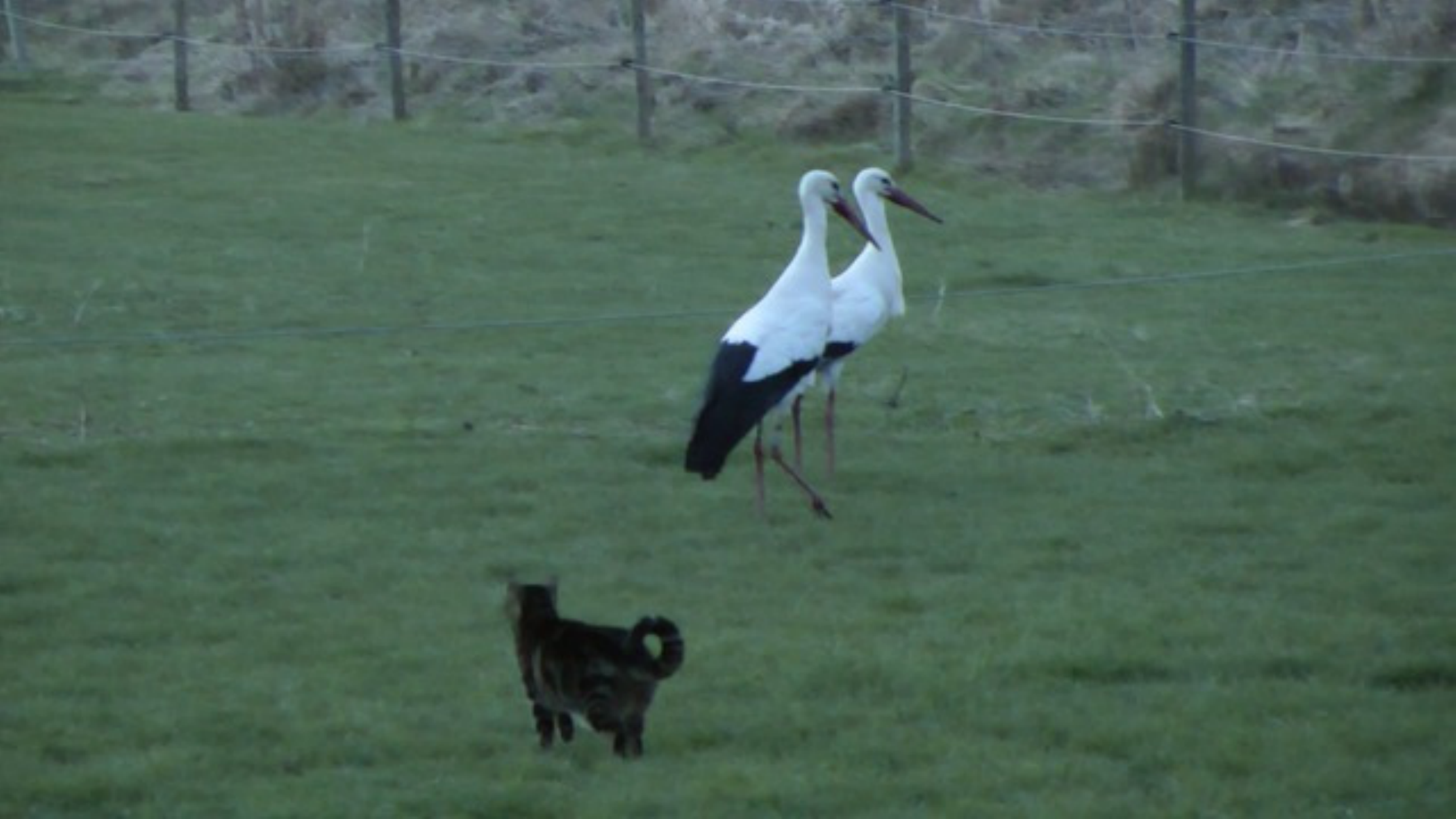}\,
            \includegraphics[width=0.19\textwidth]{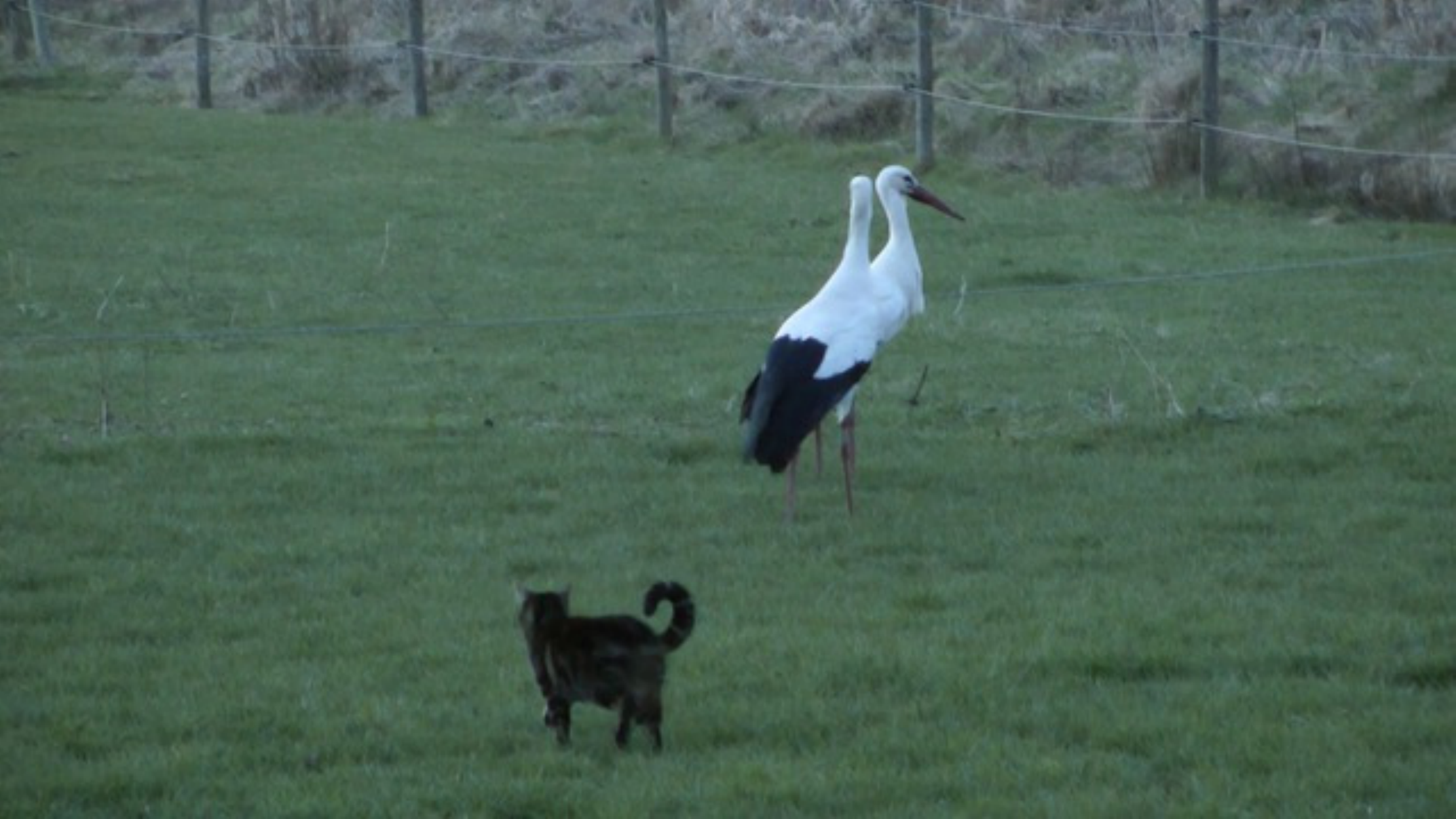}\,
            \includegraphics[width=0.19\textwidth]{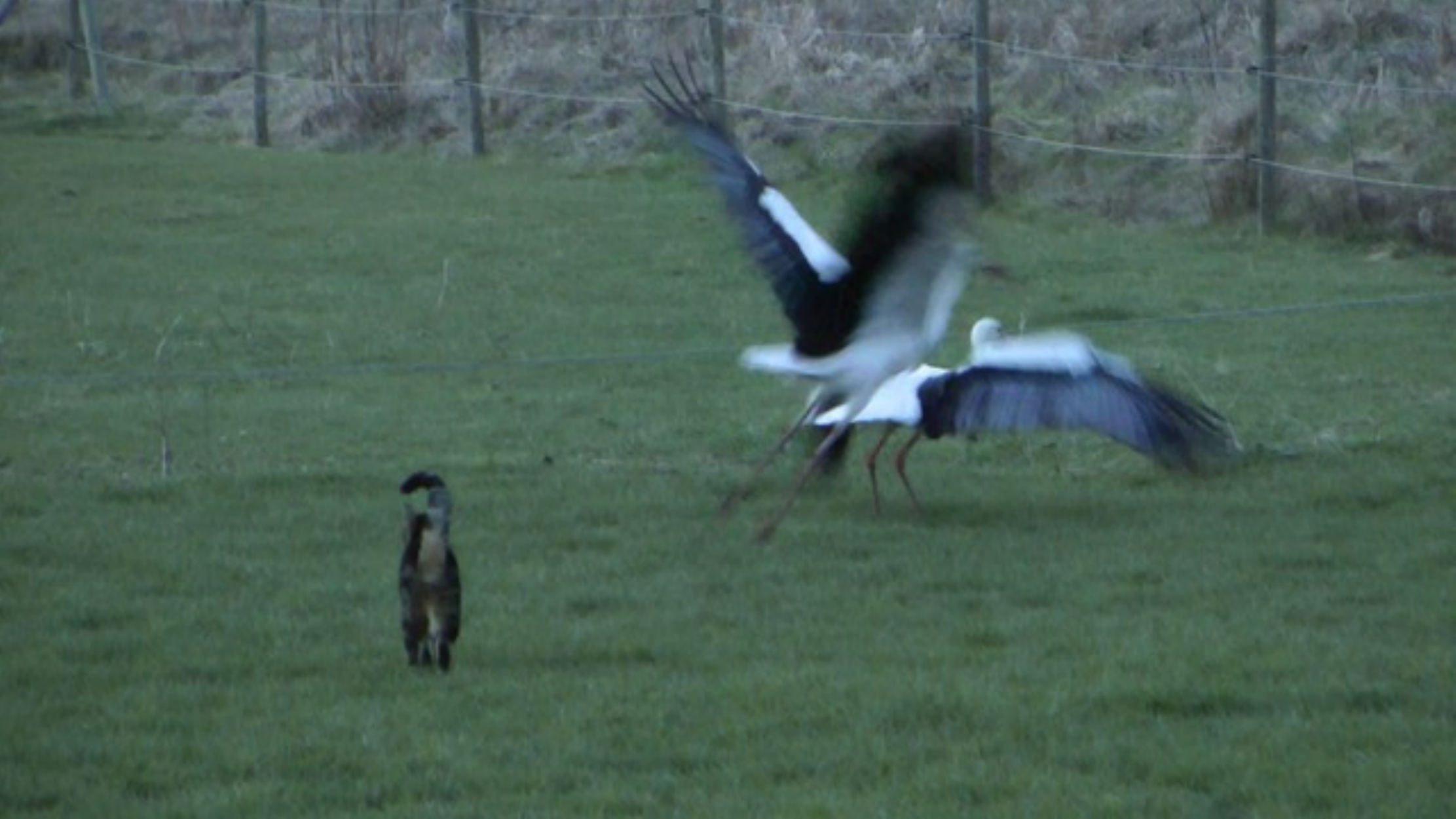}\,
            \includegraphics[width=0.19\textwidth]{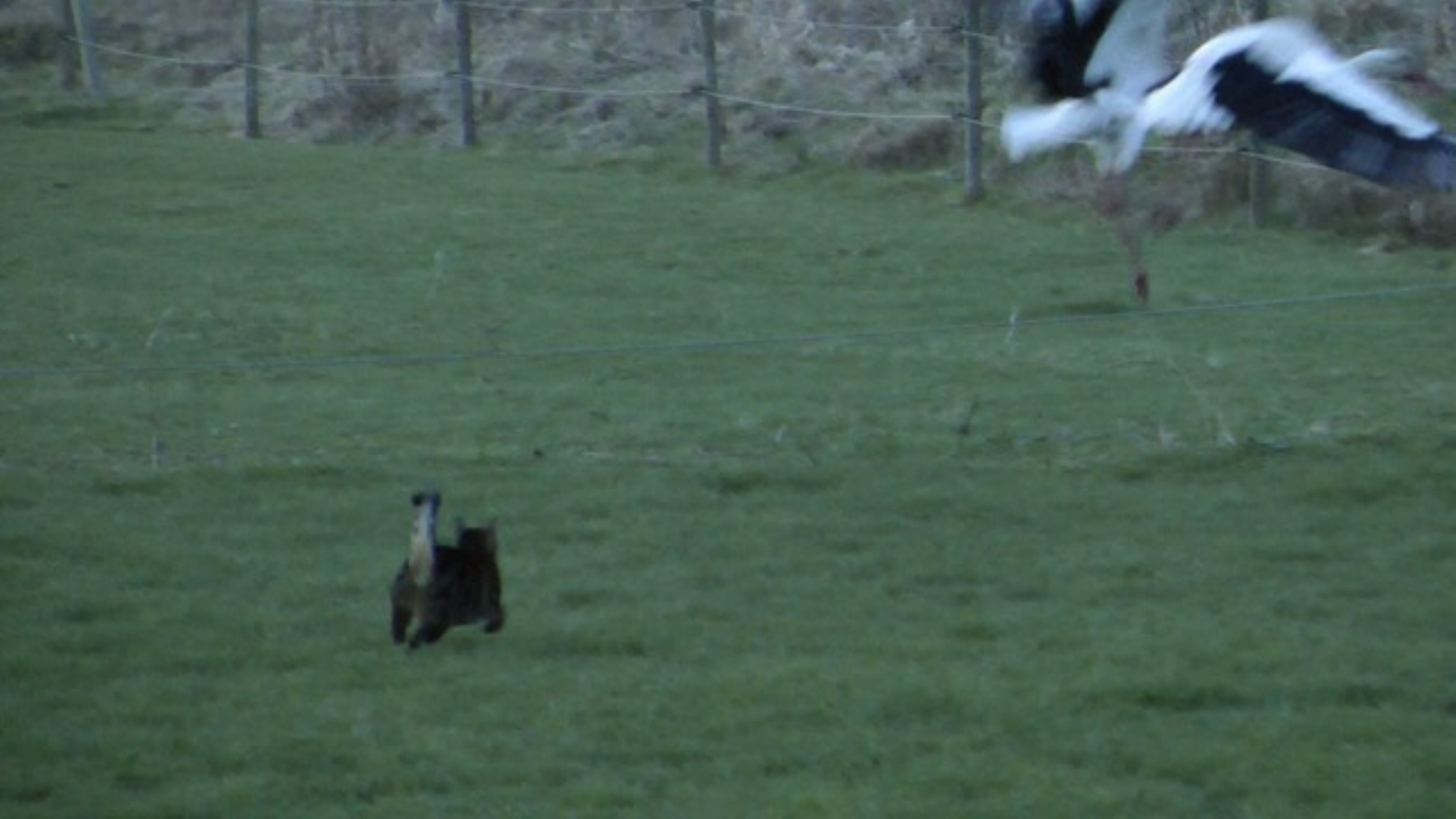}\,\\
            \includegraphics[width=0.19\textwidth]{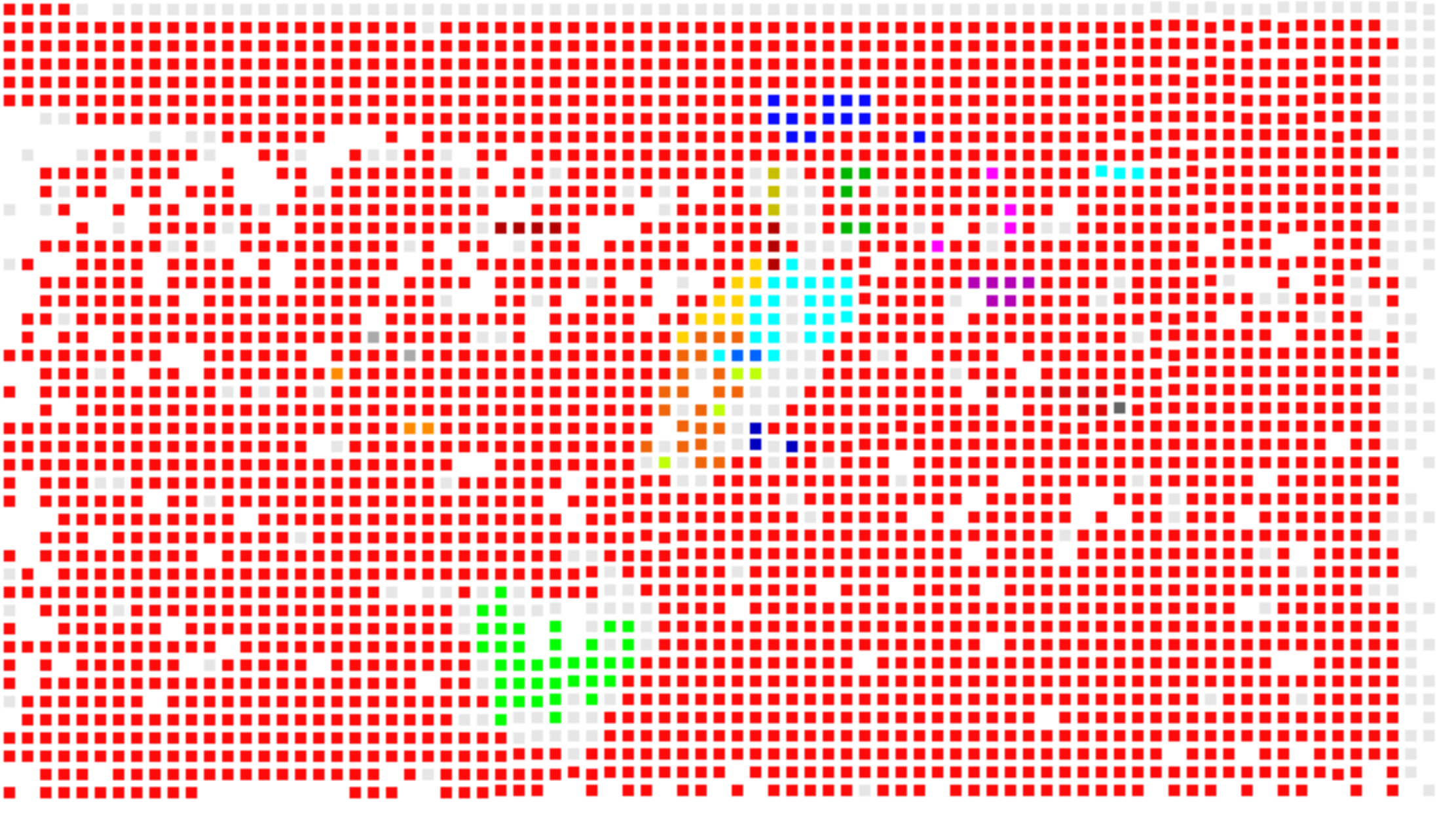}\,
            \includegraphics[width=0.19\textwidth]{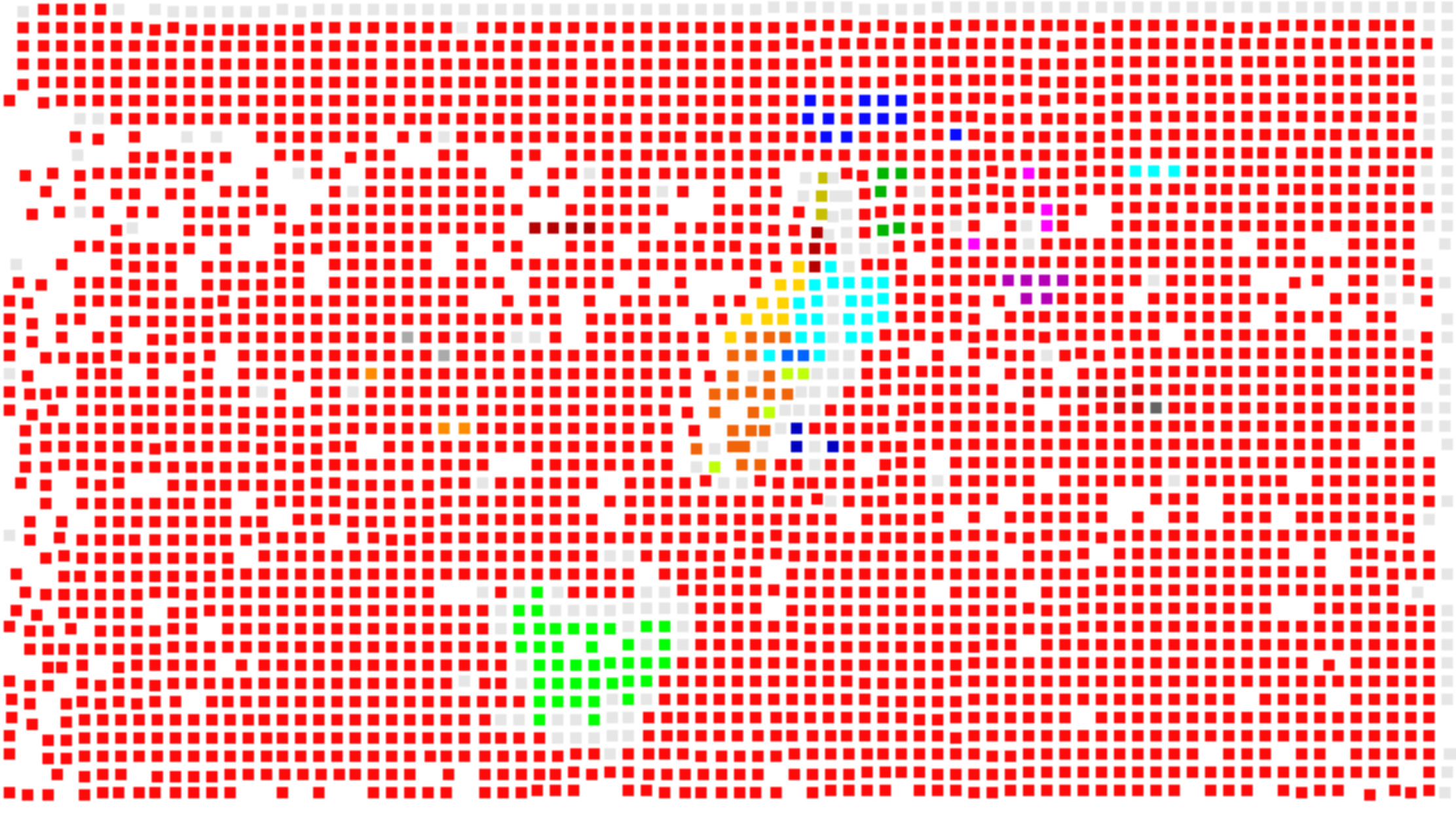}\,
            \includegraphics[width=0.19\textwidth]{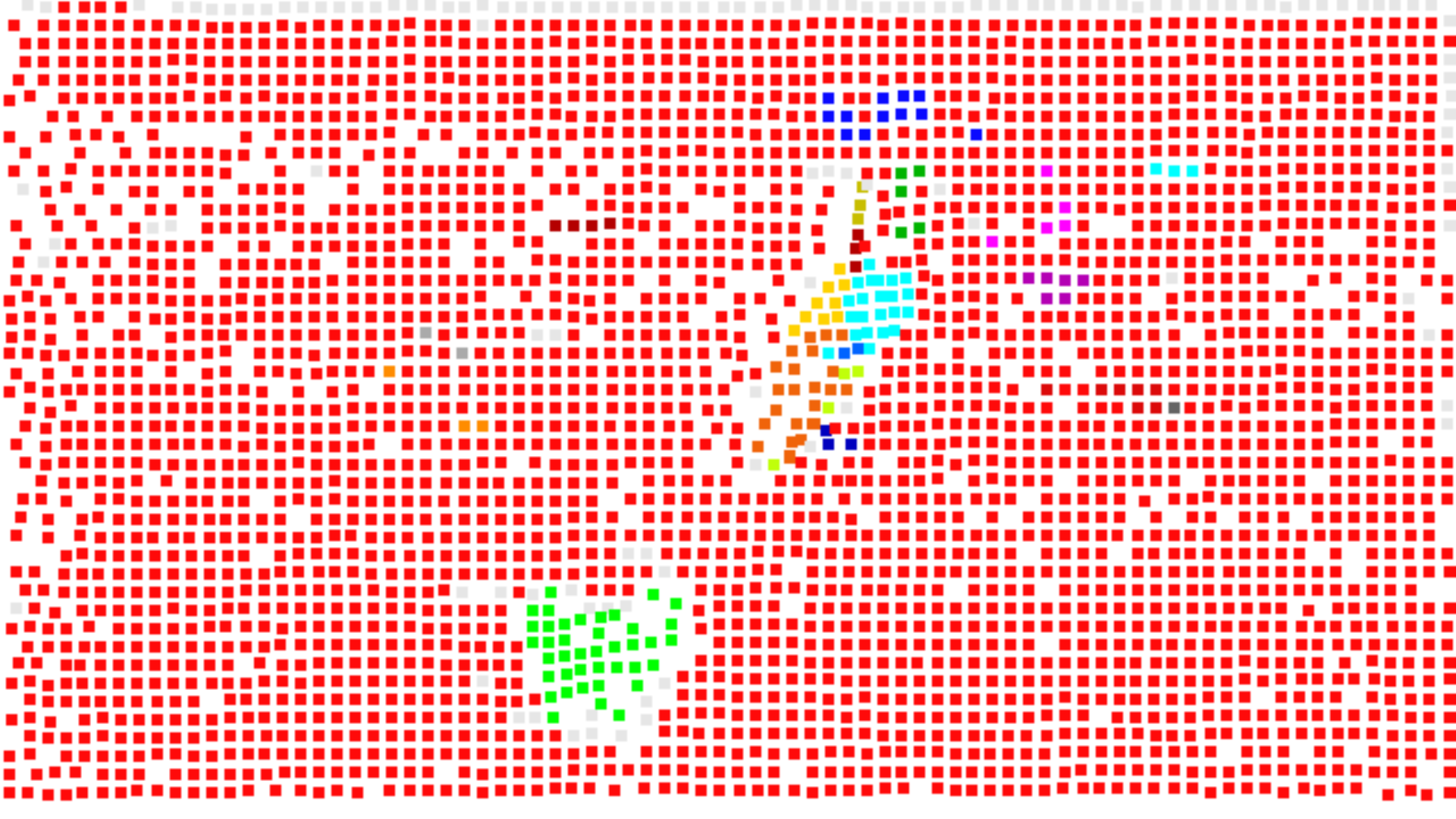}\,
            \includegraphics[width=0.19\textwidth]{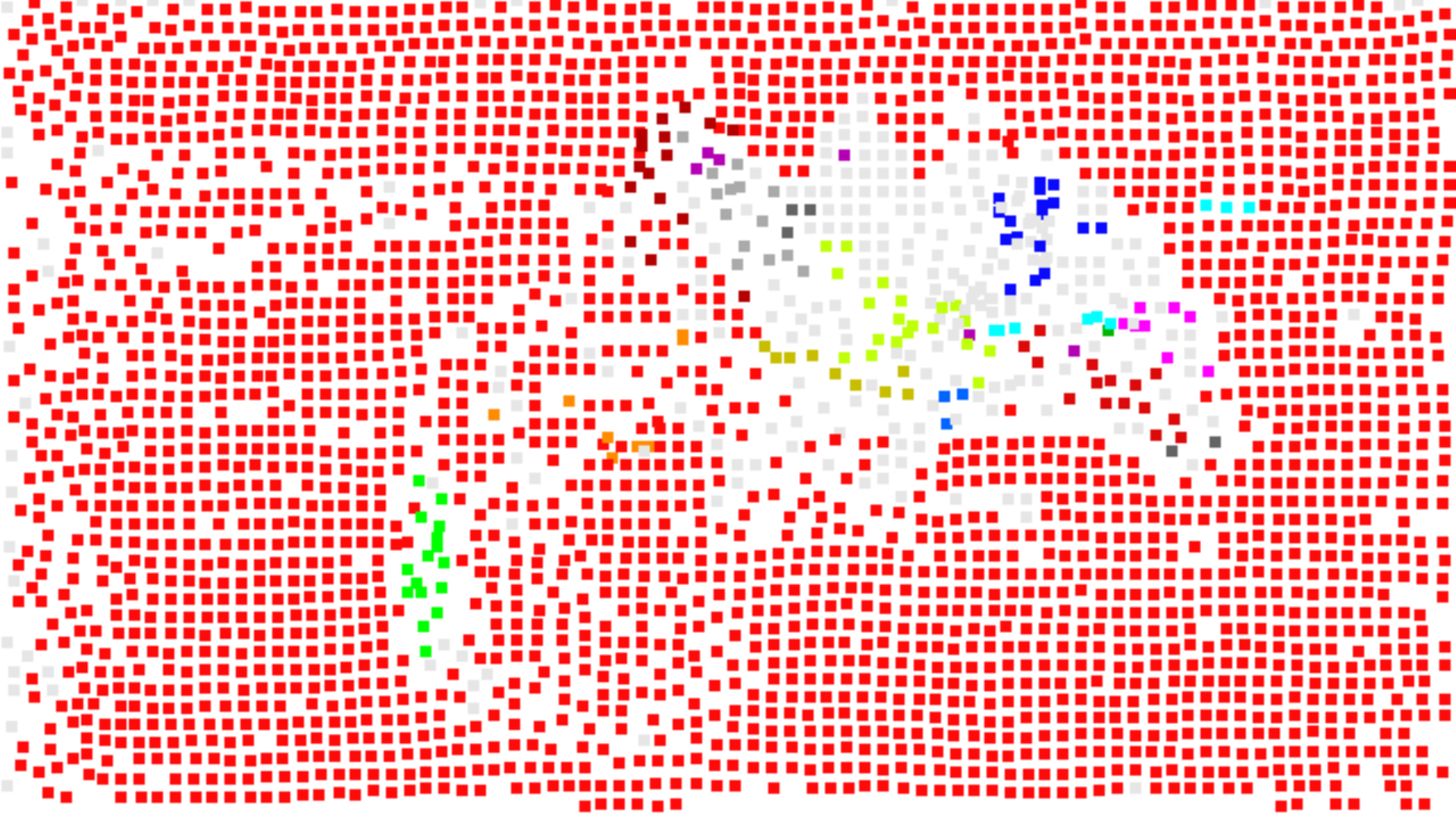}\,
            \includegraphics[width=0.19\textwidth]{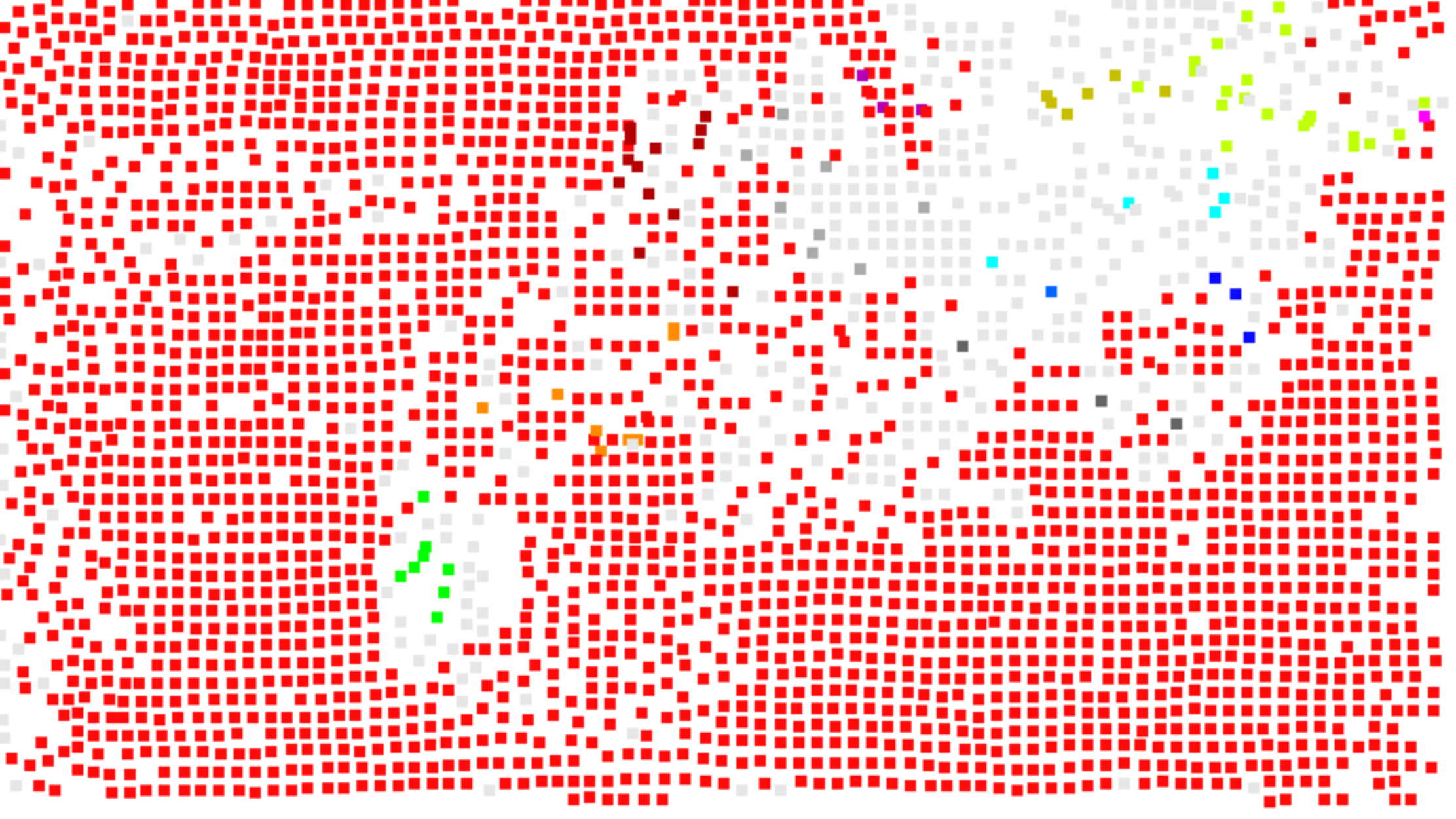}\,\\
            \includegraphics[width=0.19\textwidth]{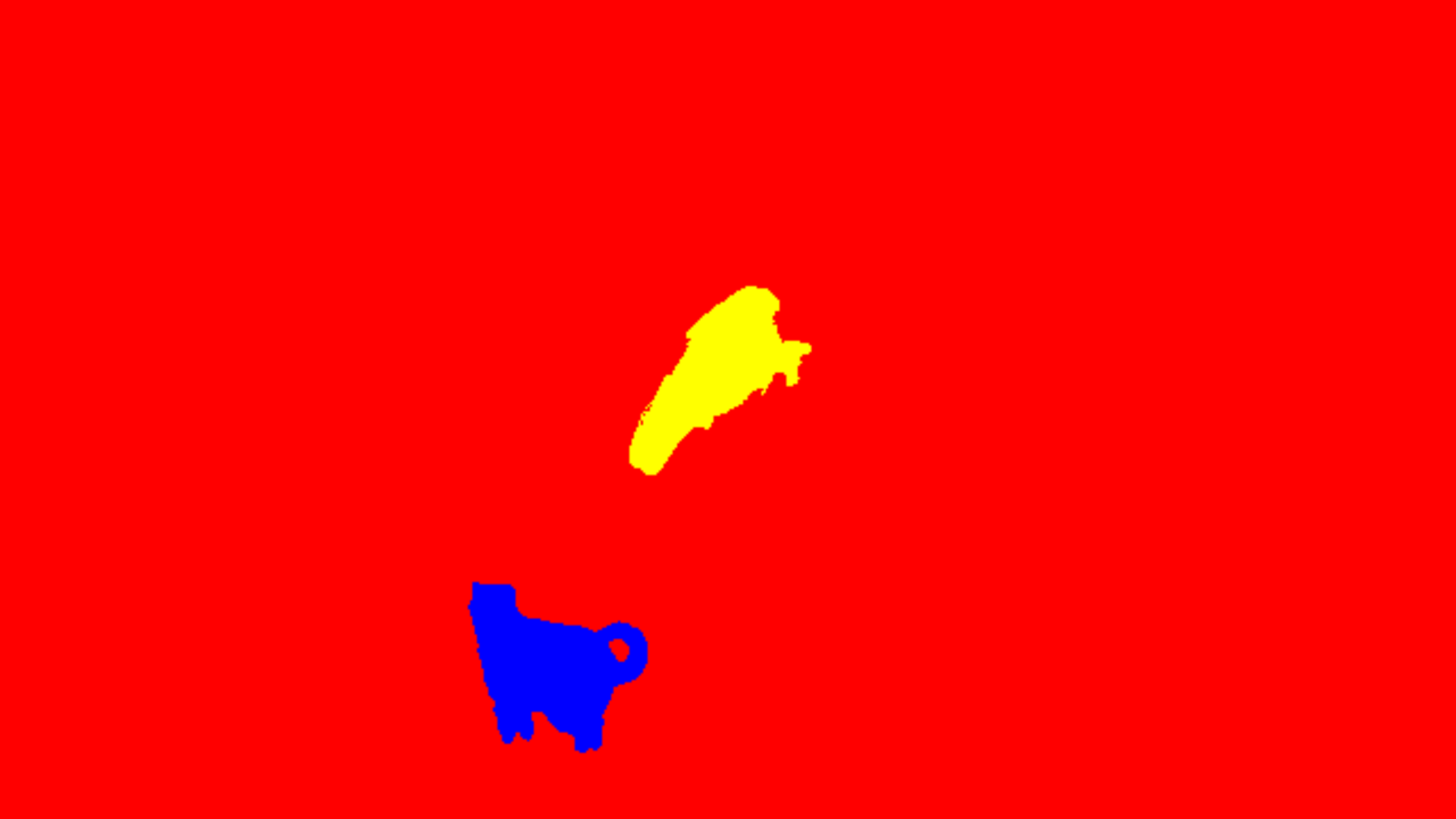}\,
            \includegraphics[width=0.19\textwidth]{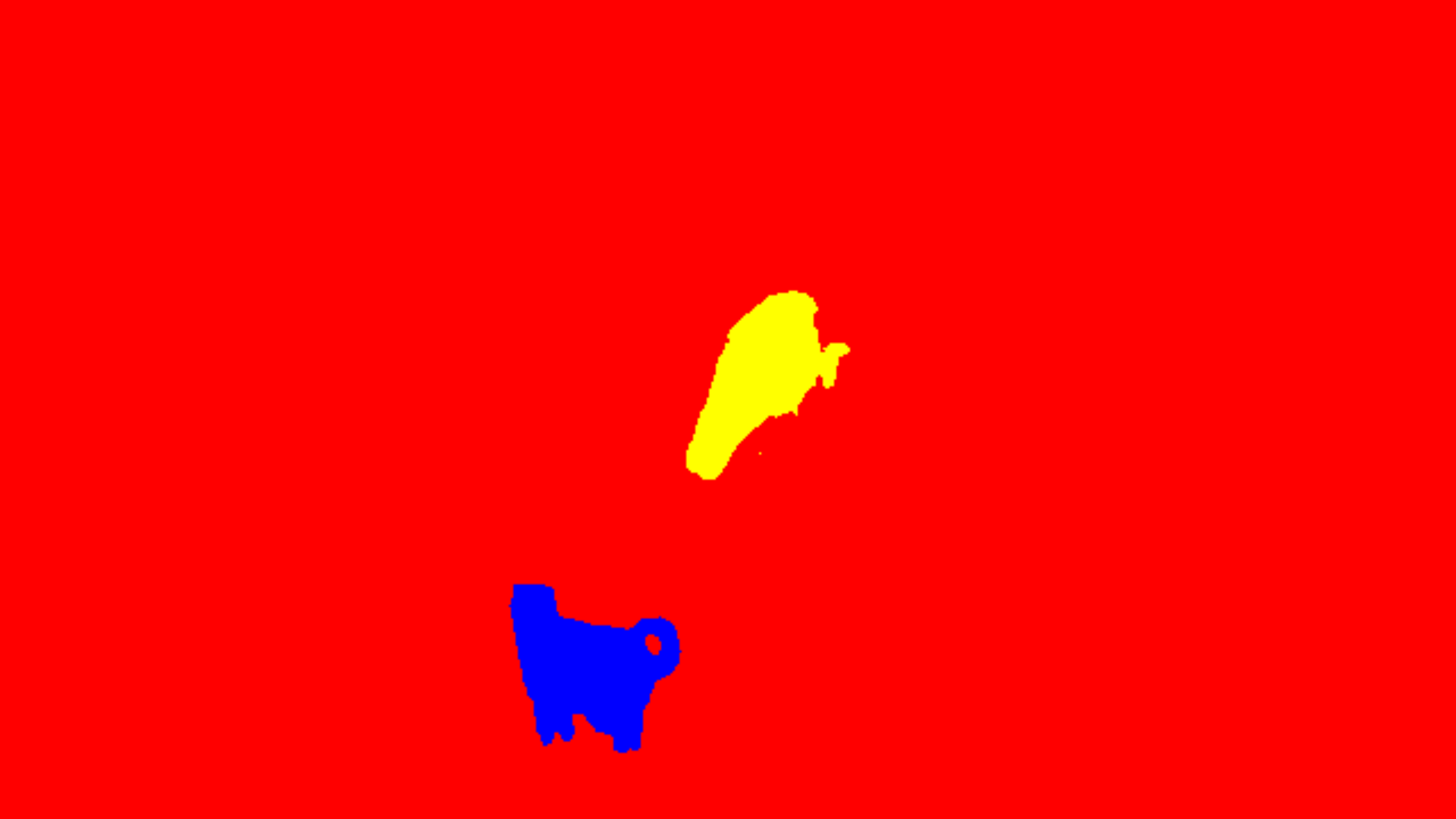}\,
            \includegraphics[width=0.19\textwidth]{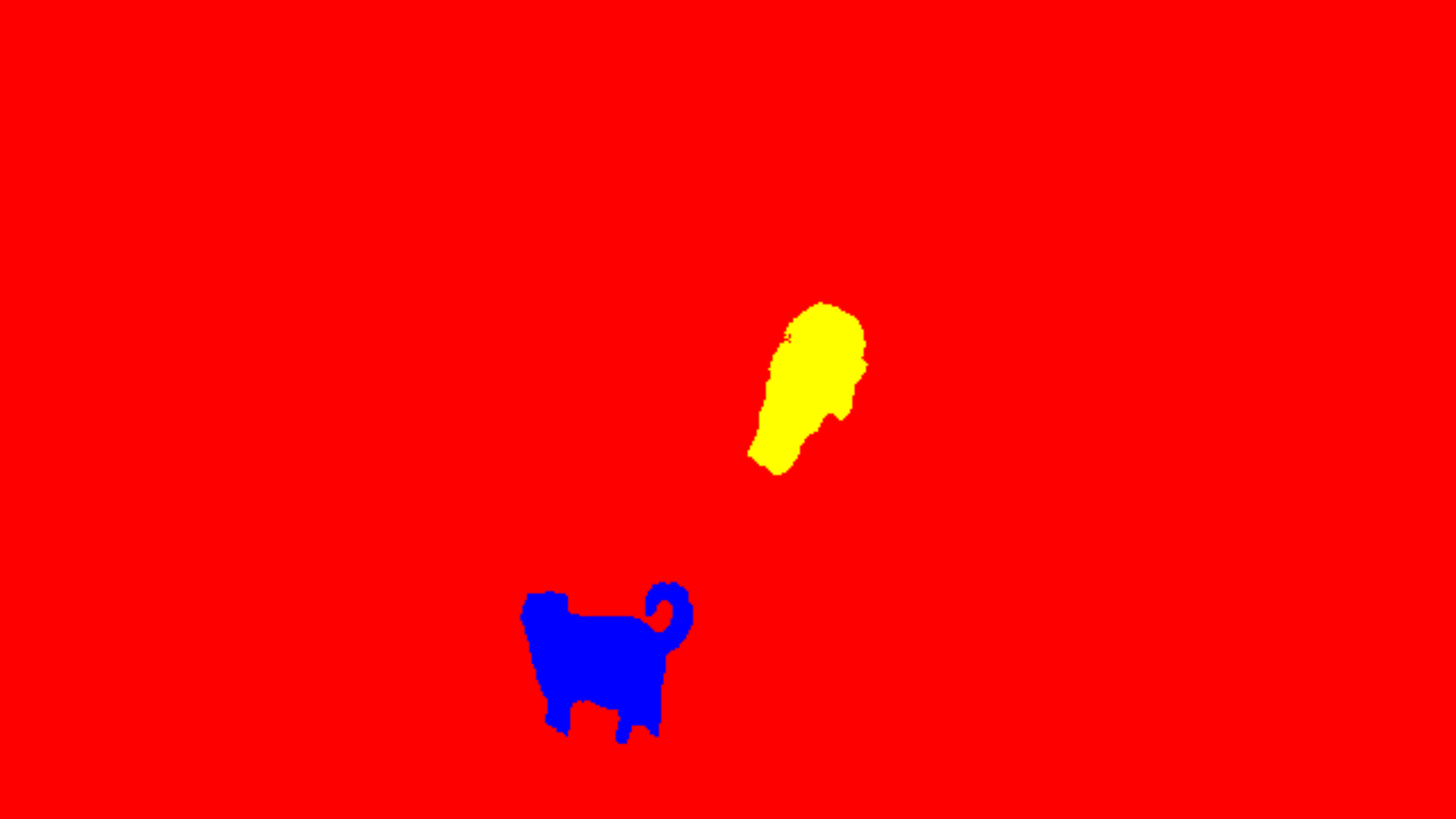}\,
            \includegraphics[width=0.19\textwidth]{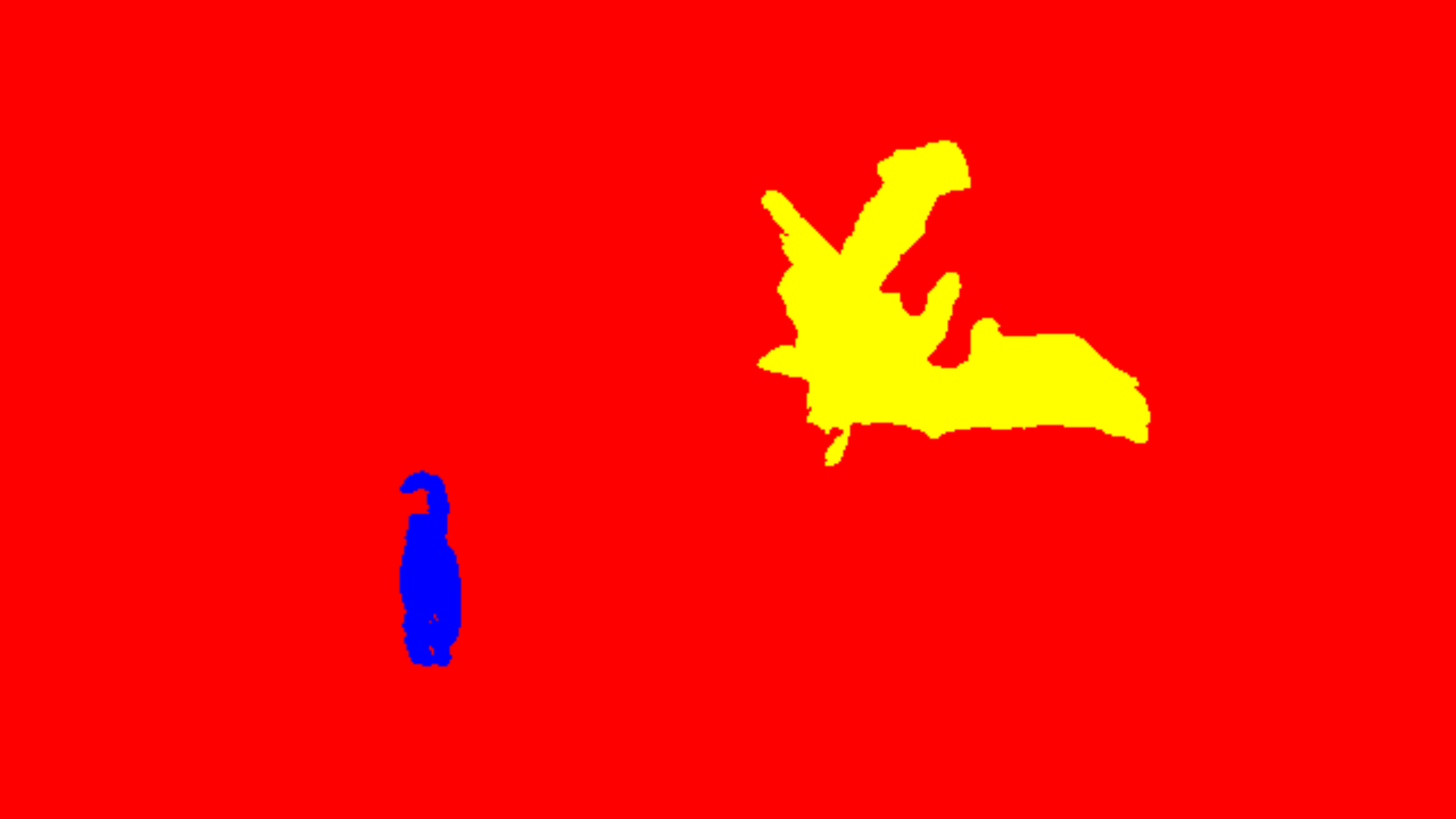}\,
            \includegraphics[width=0.19\textwidth]{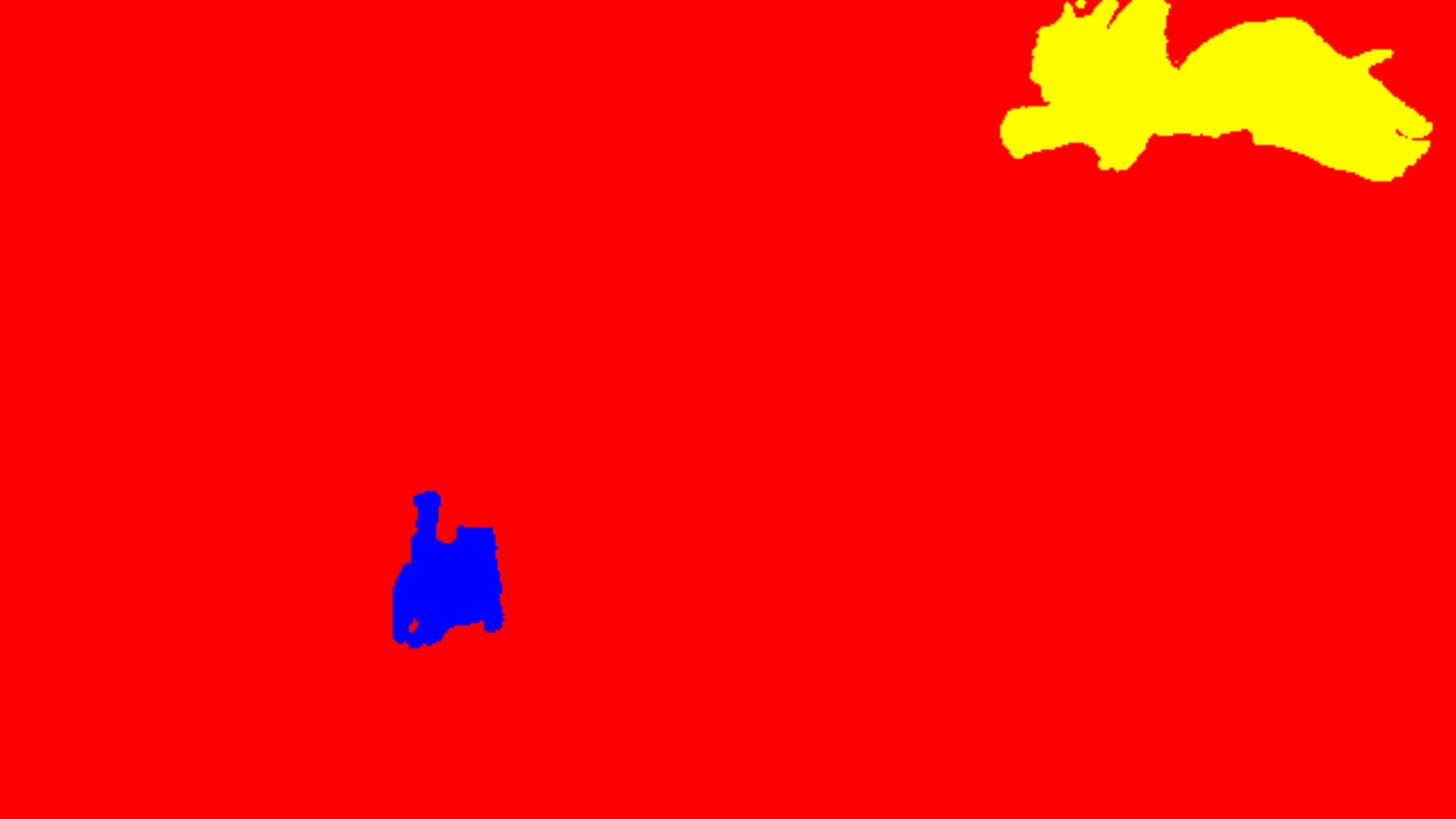}\,\\
            
            \includegraphics[width=0.19\textwidth]{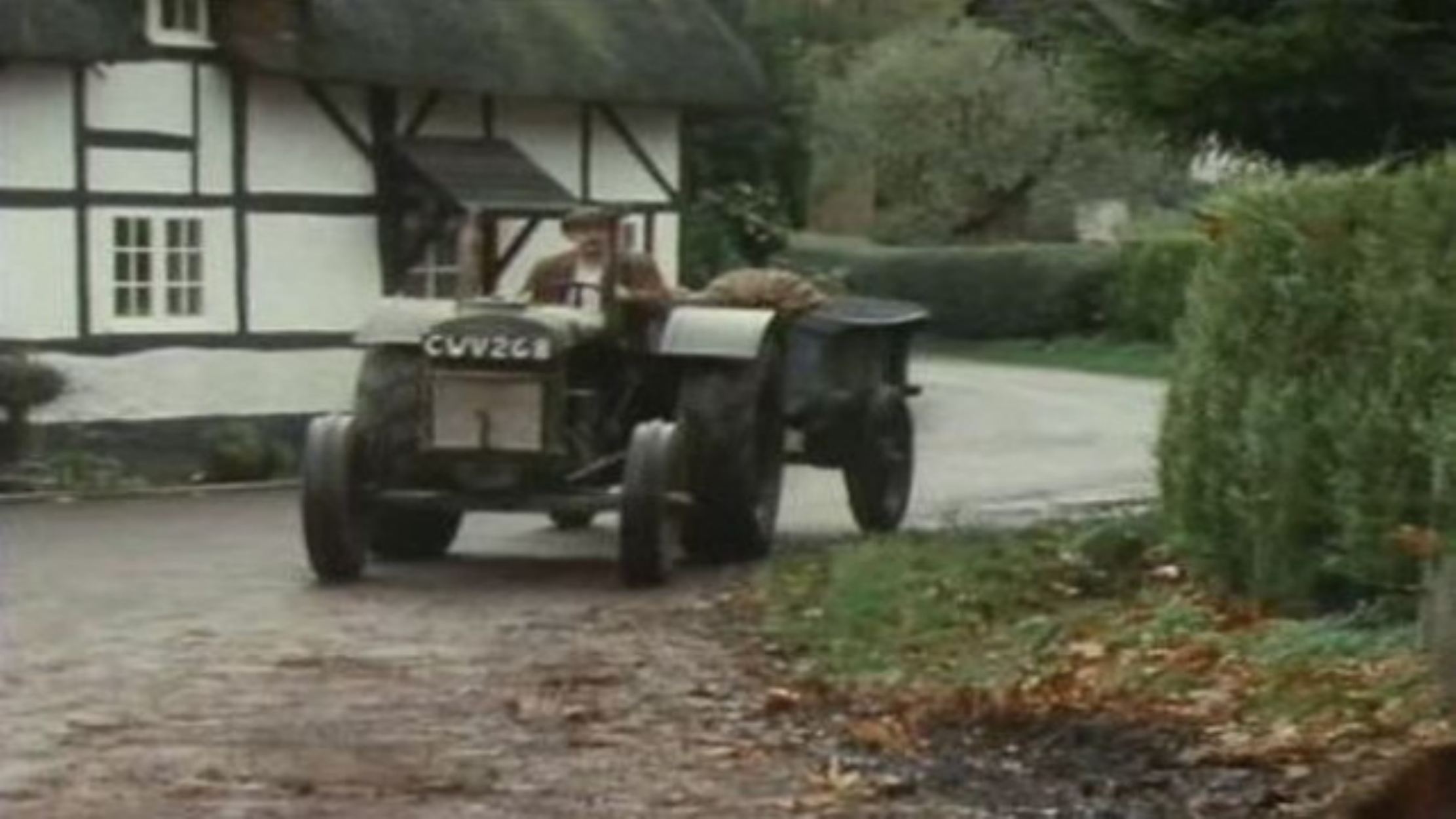}\,
            \includegraphics[width=0.19\textwidth]{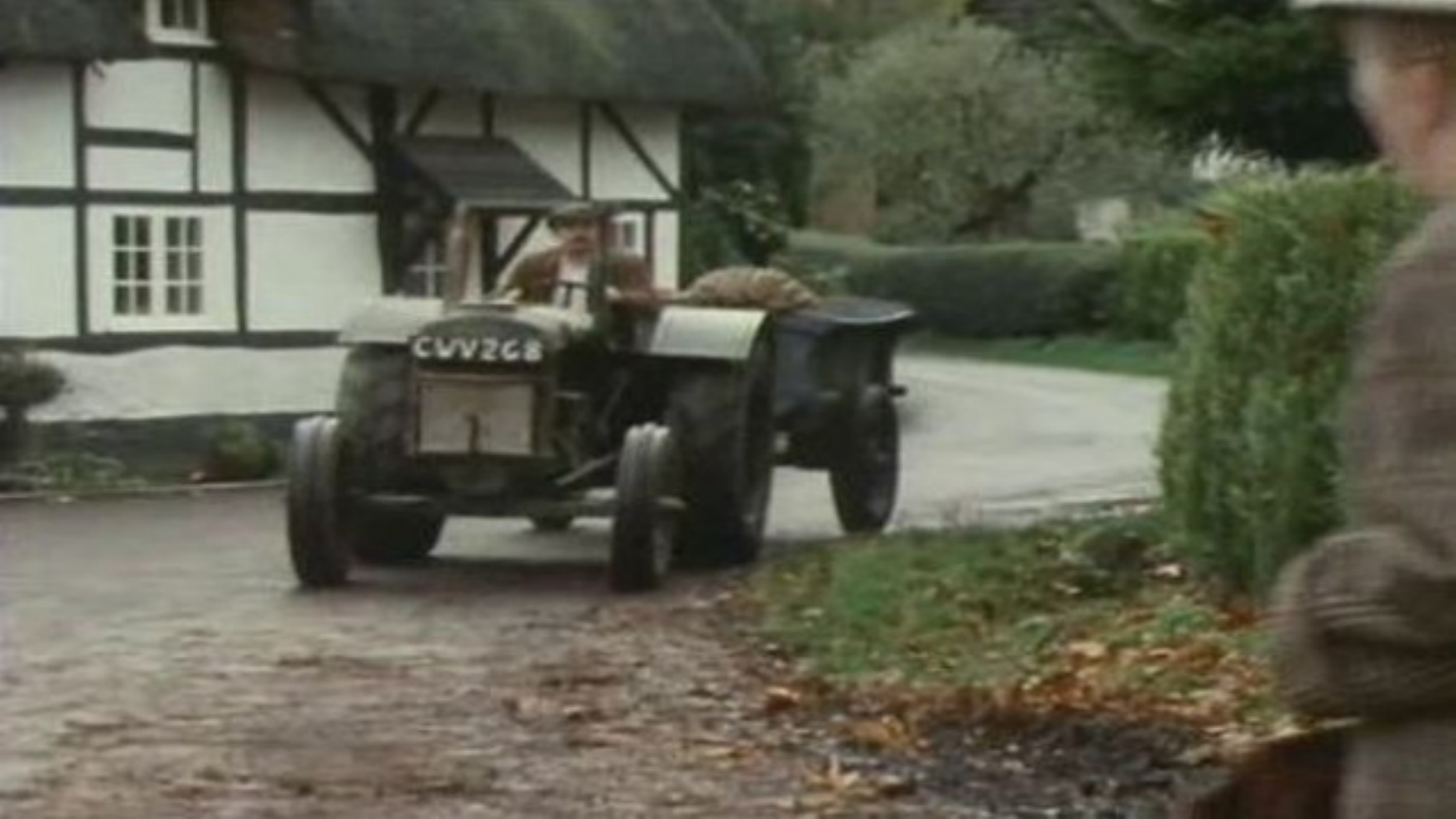}\,
            \includegraphics[width=0.19\textwidth]{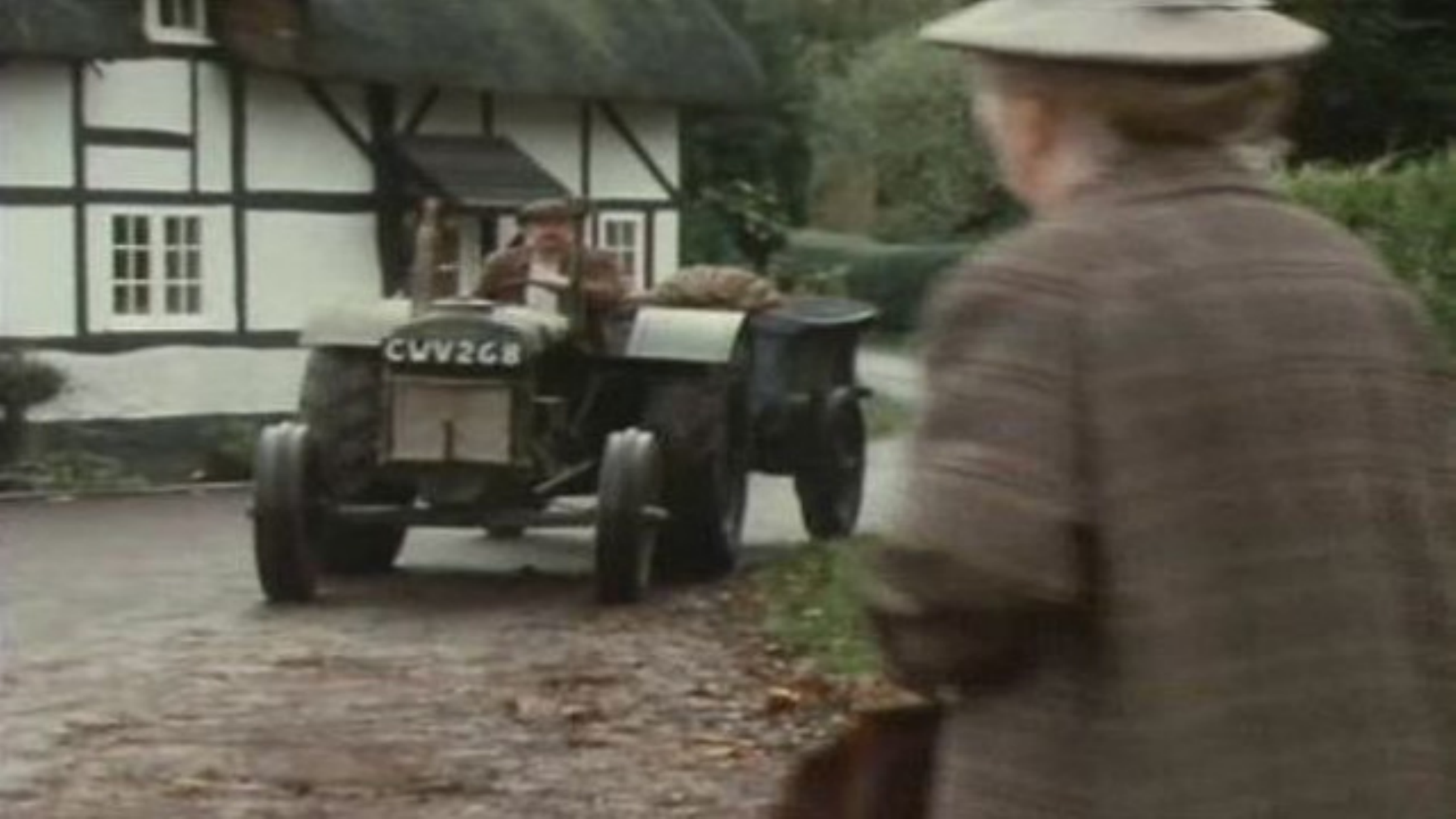}\,
            \includegraphics[width=0.19\textwidth]{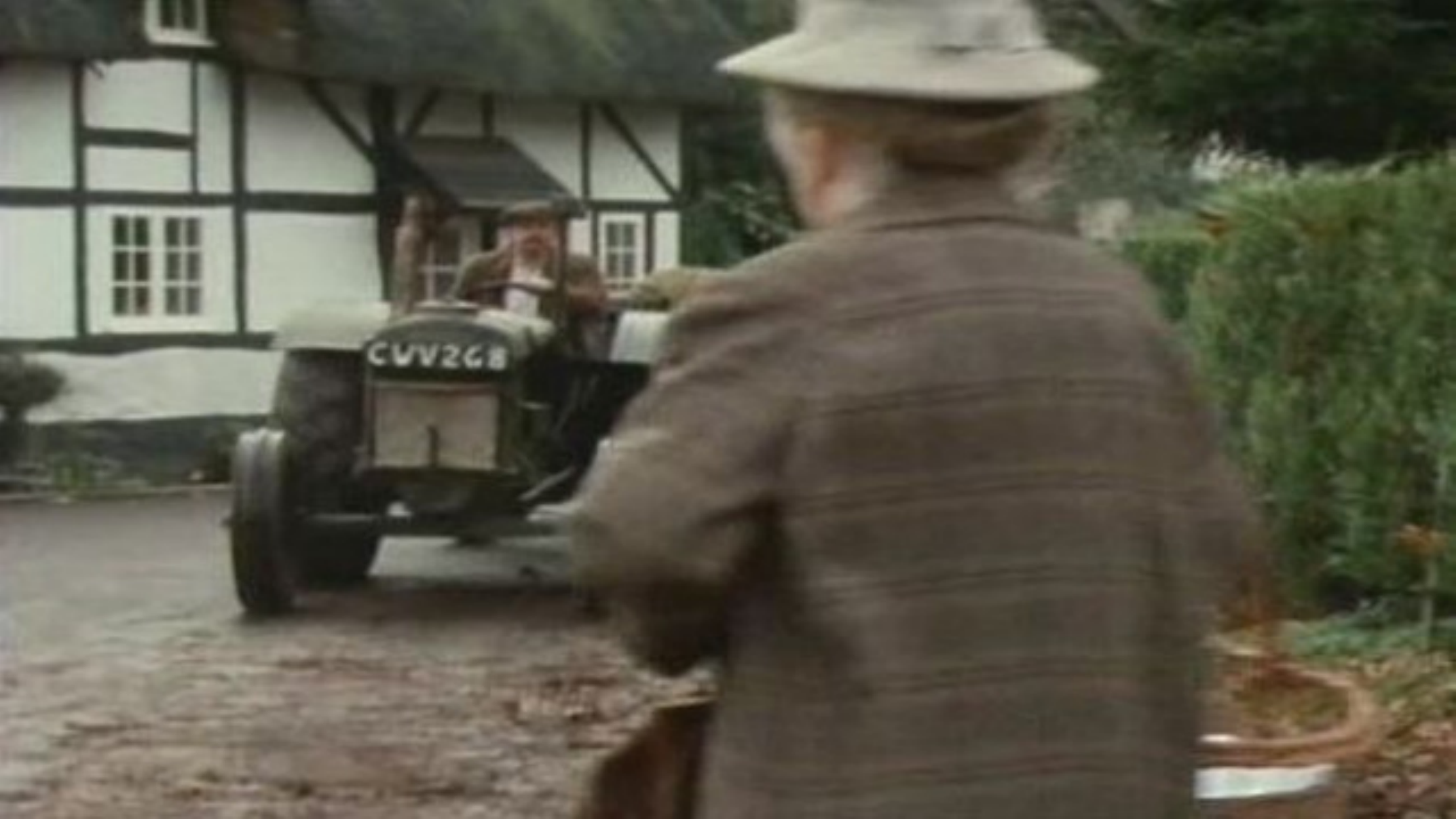}\,
            \includegraphics[width=0.19\textwidth]{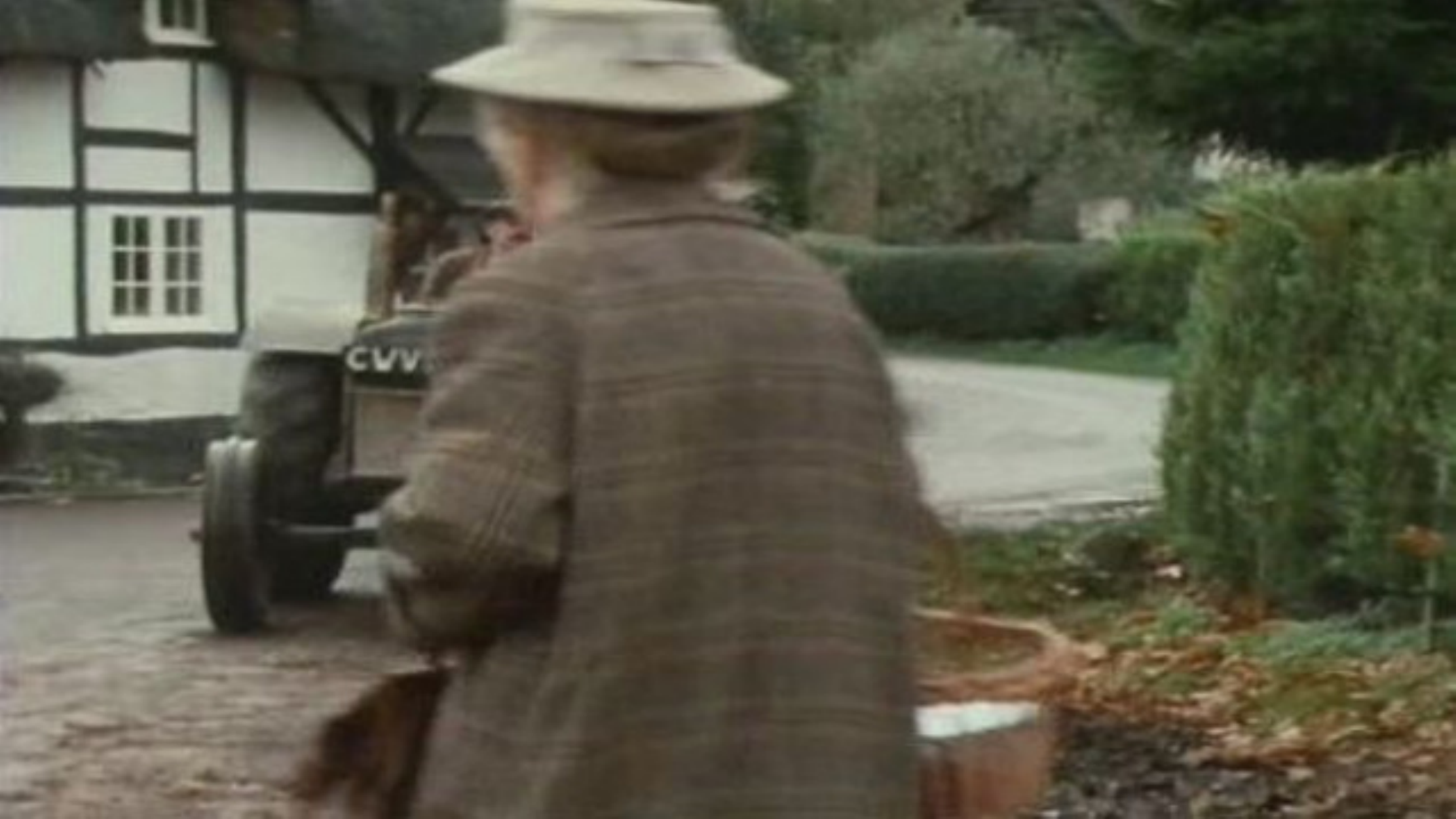}\,\\
            \includegraphics[width=0.19\textwidth]{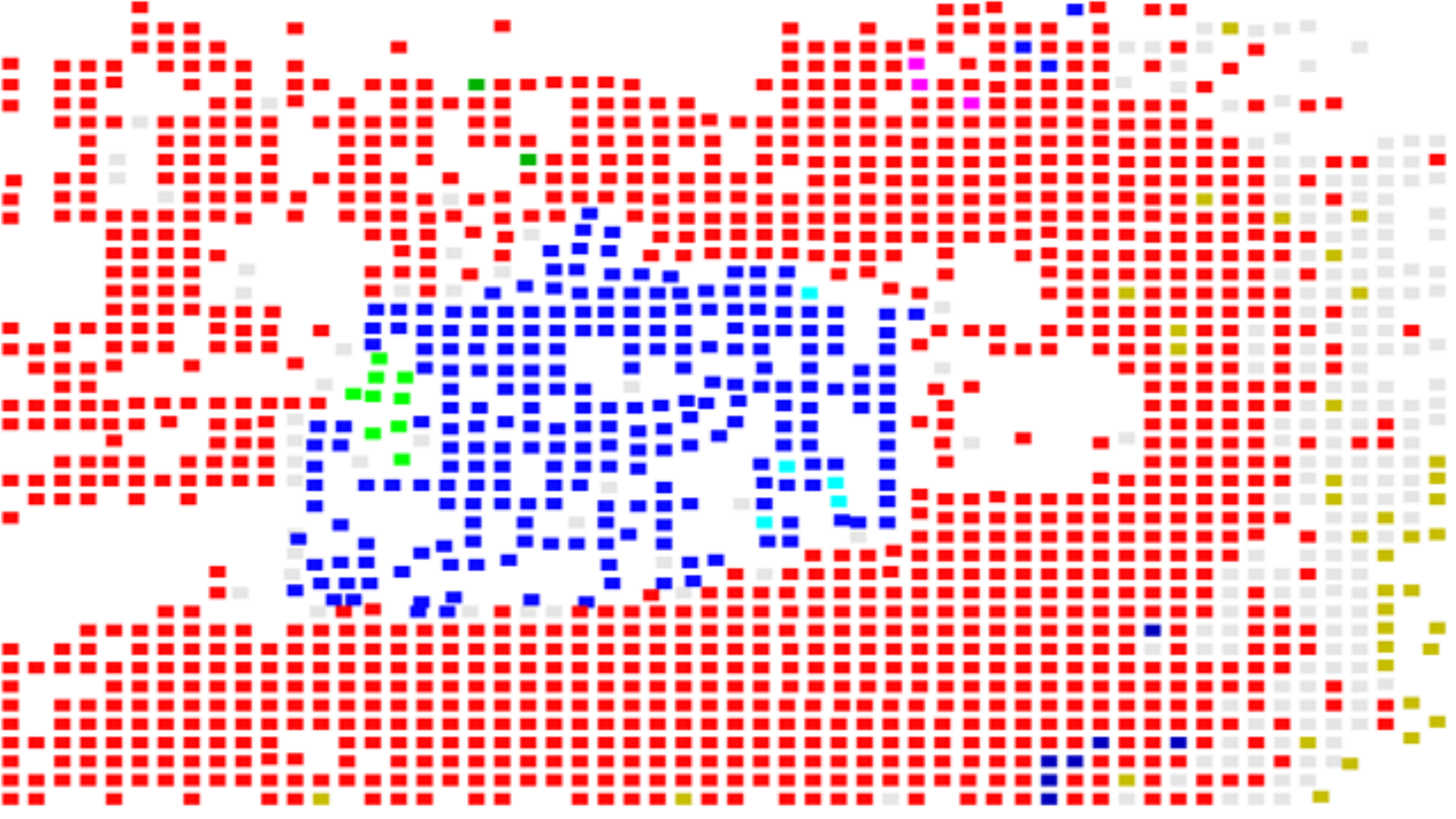}\,
            \includegraphics[width=0.19\textwidth]{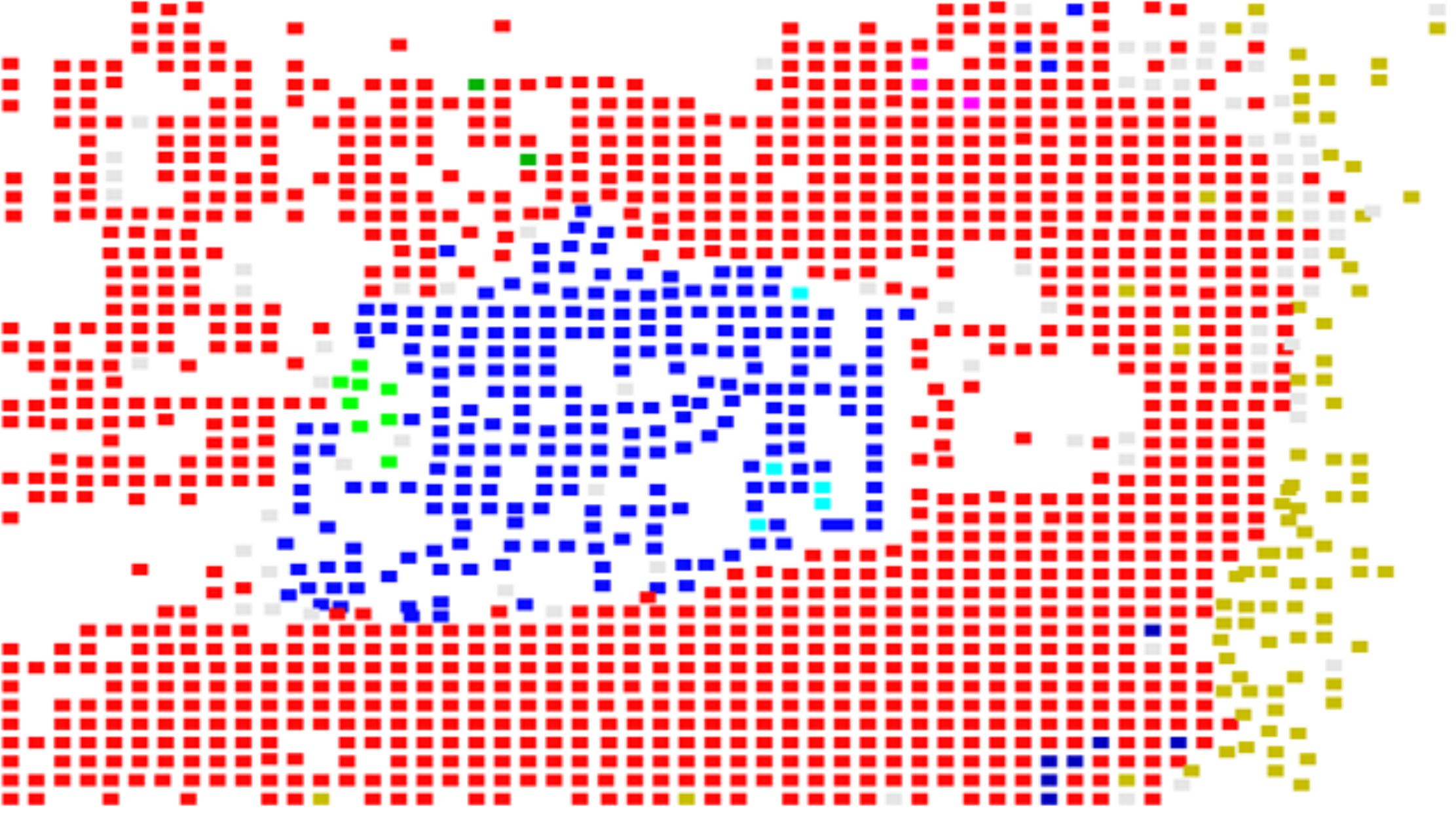}\,
            \includegraphics[width=0.19\textwidth]{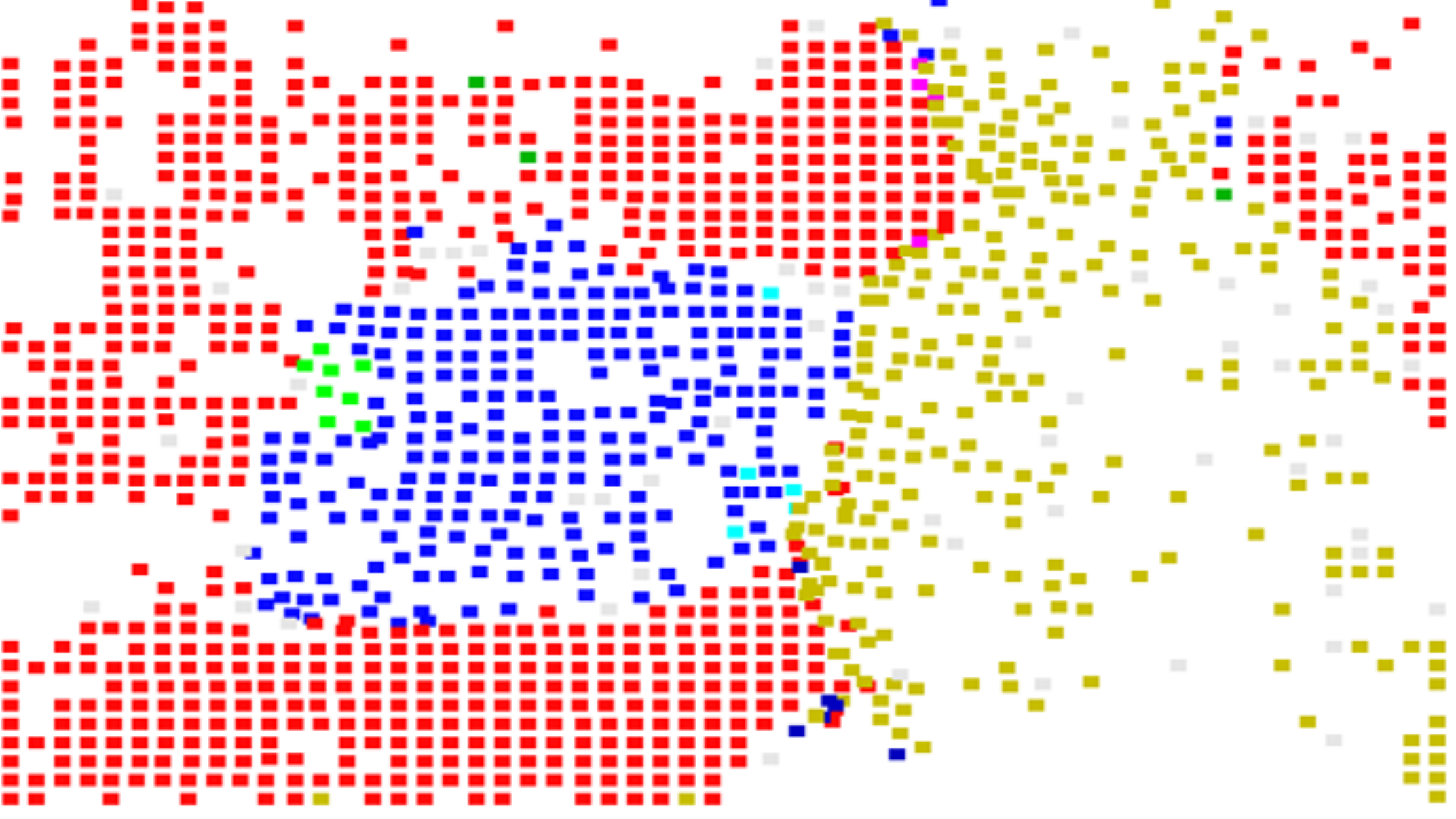}\,
            \includegraphics[width=0.19\textwidth]{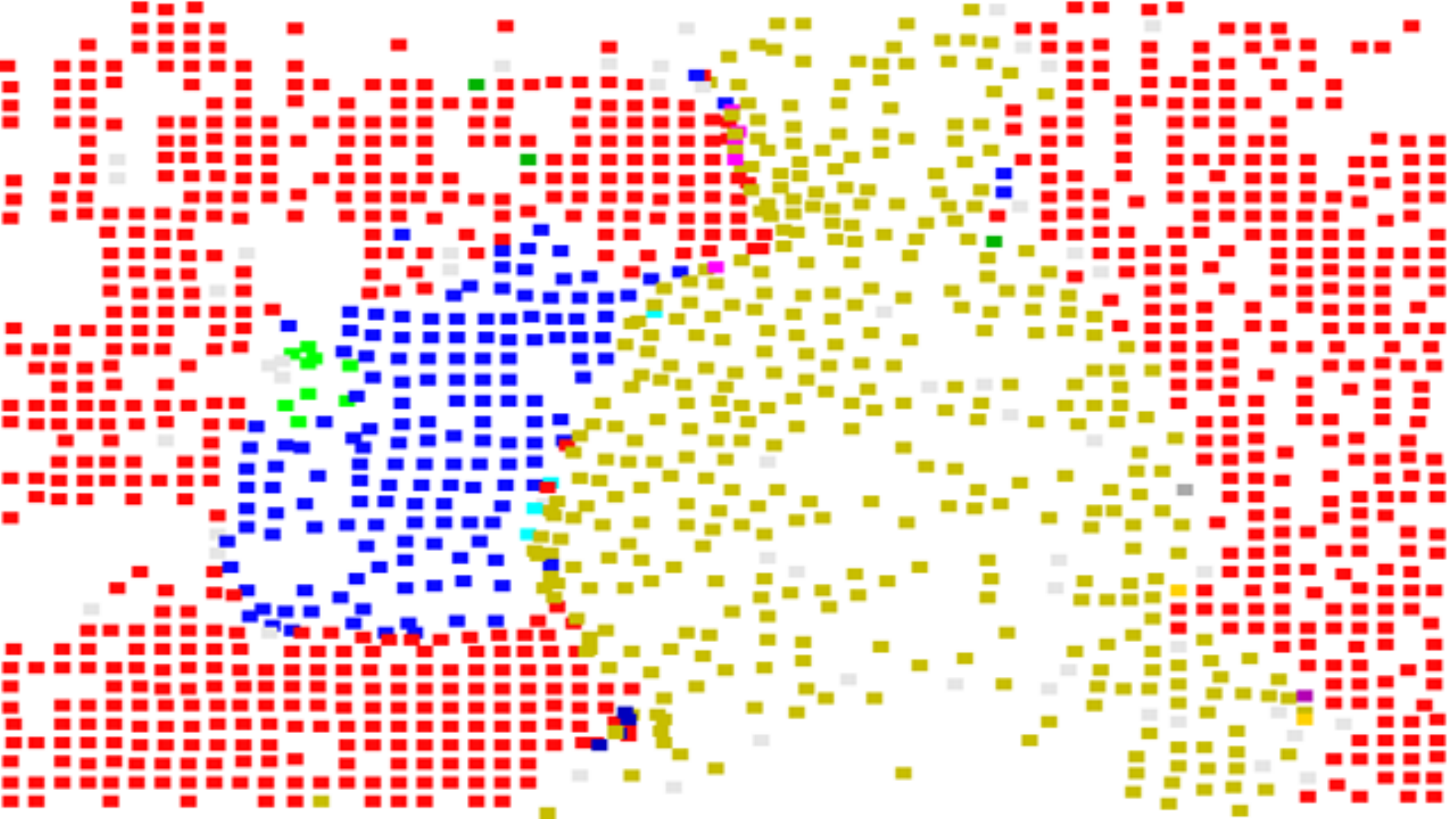}\,
            \includegraphics[width=0.19\textwidth]{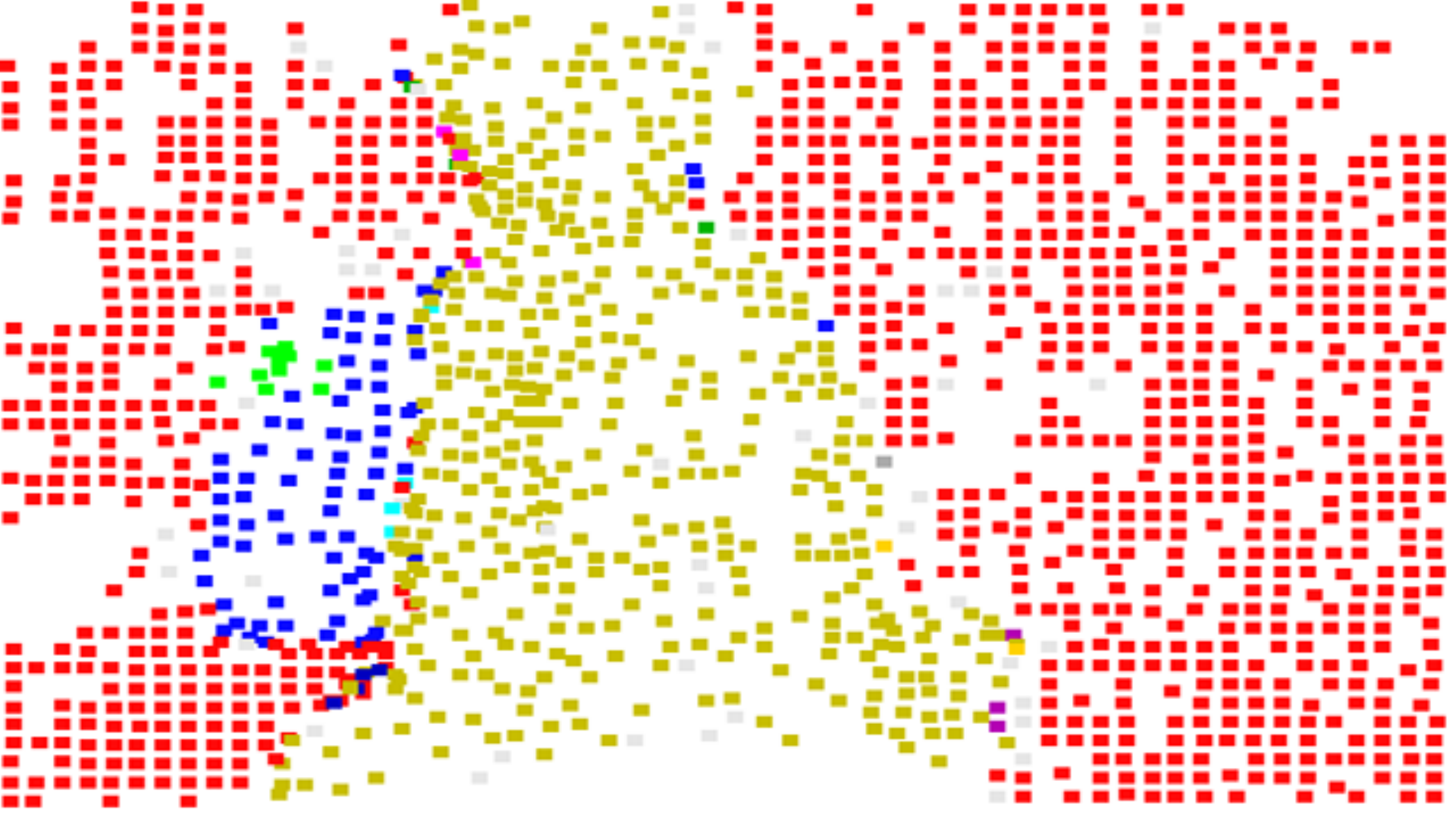}\,\\
            \includegraphics[width=0.19\textwidth]{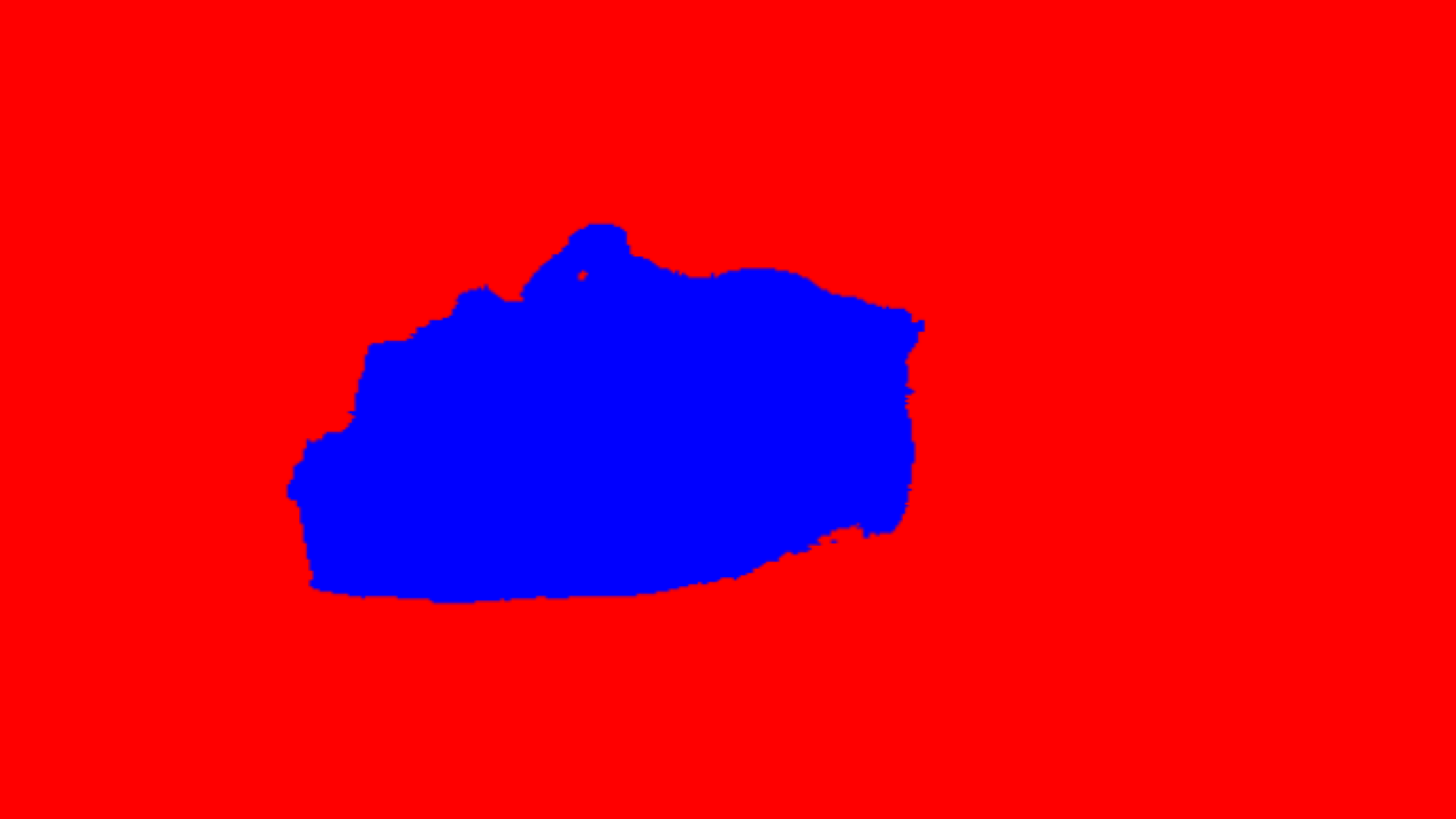}\,
            \includegraphics[width=0.19\textwidth]{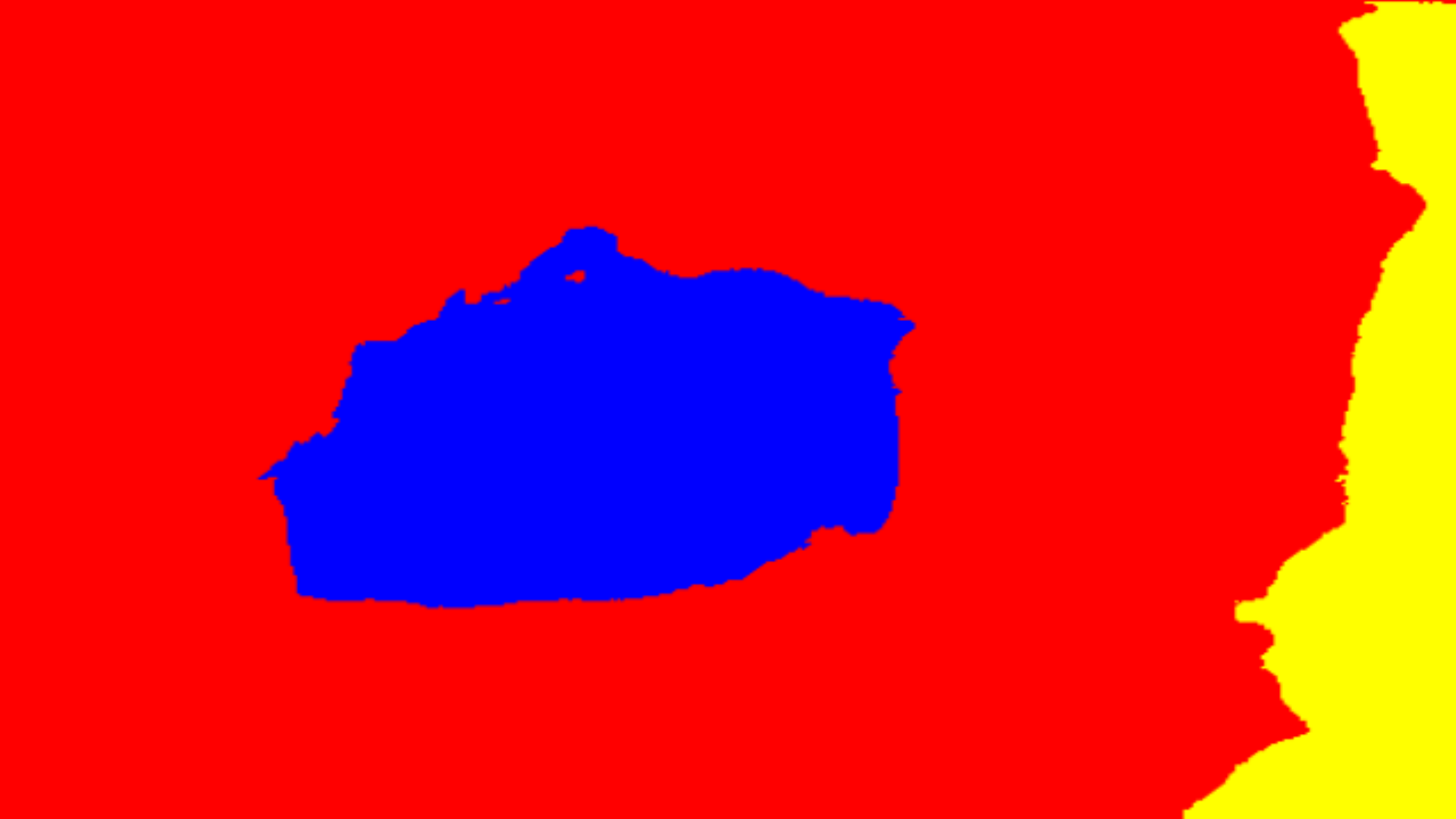}\,
            \includegraphics[width=0.19\textwidth]{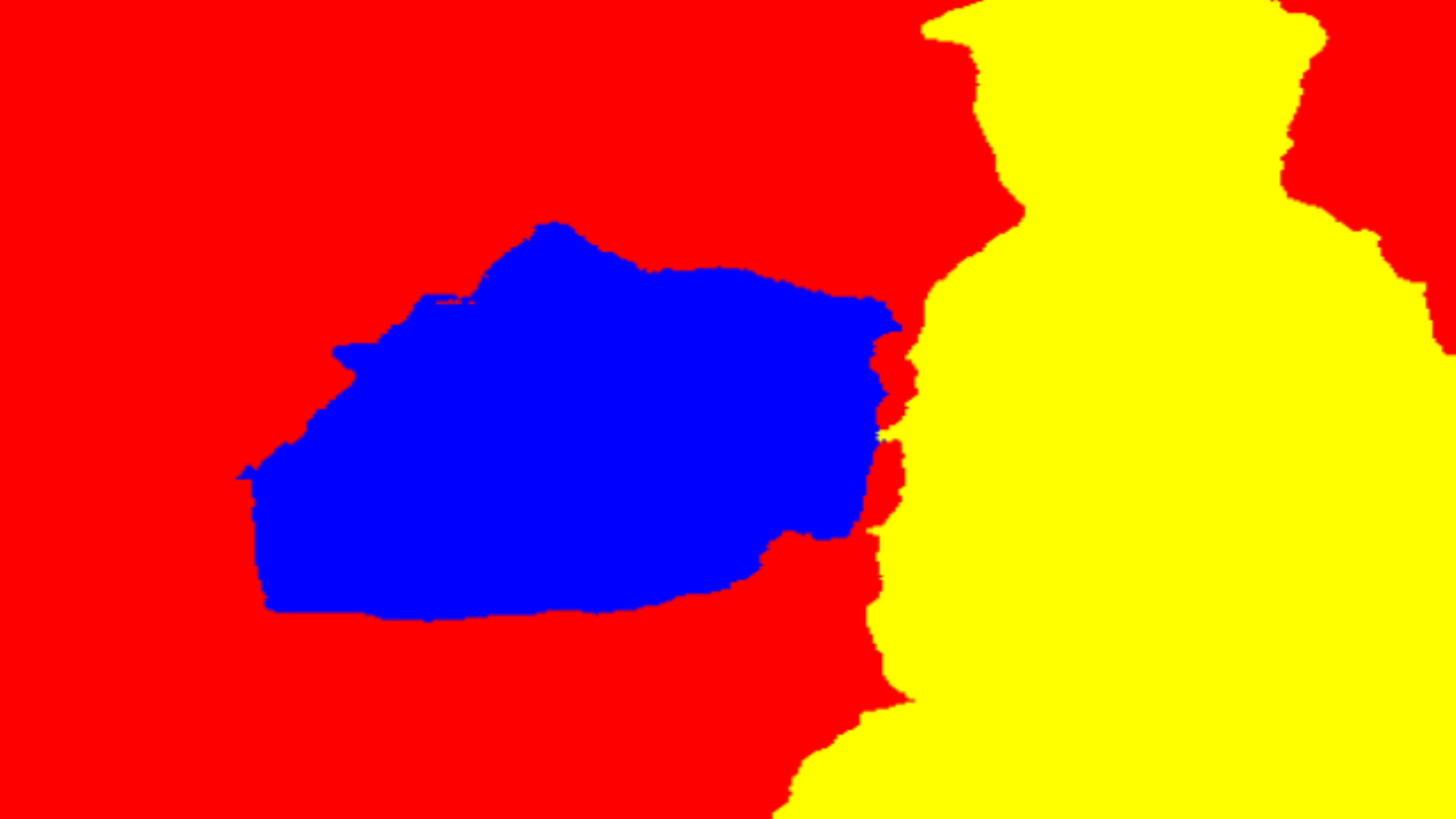}\,
            \includegraphics[width=0.19\textwidth]{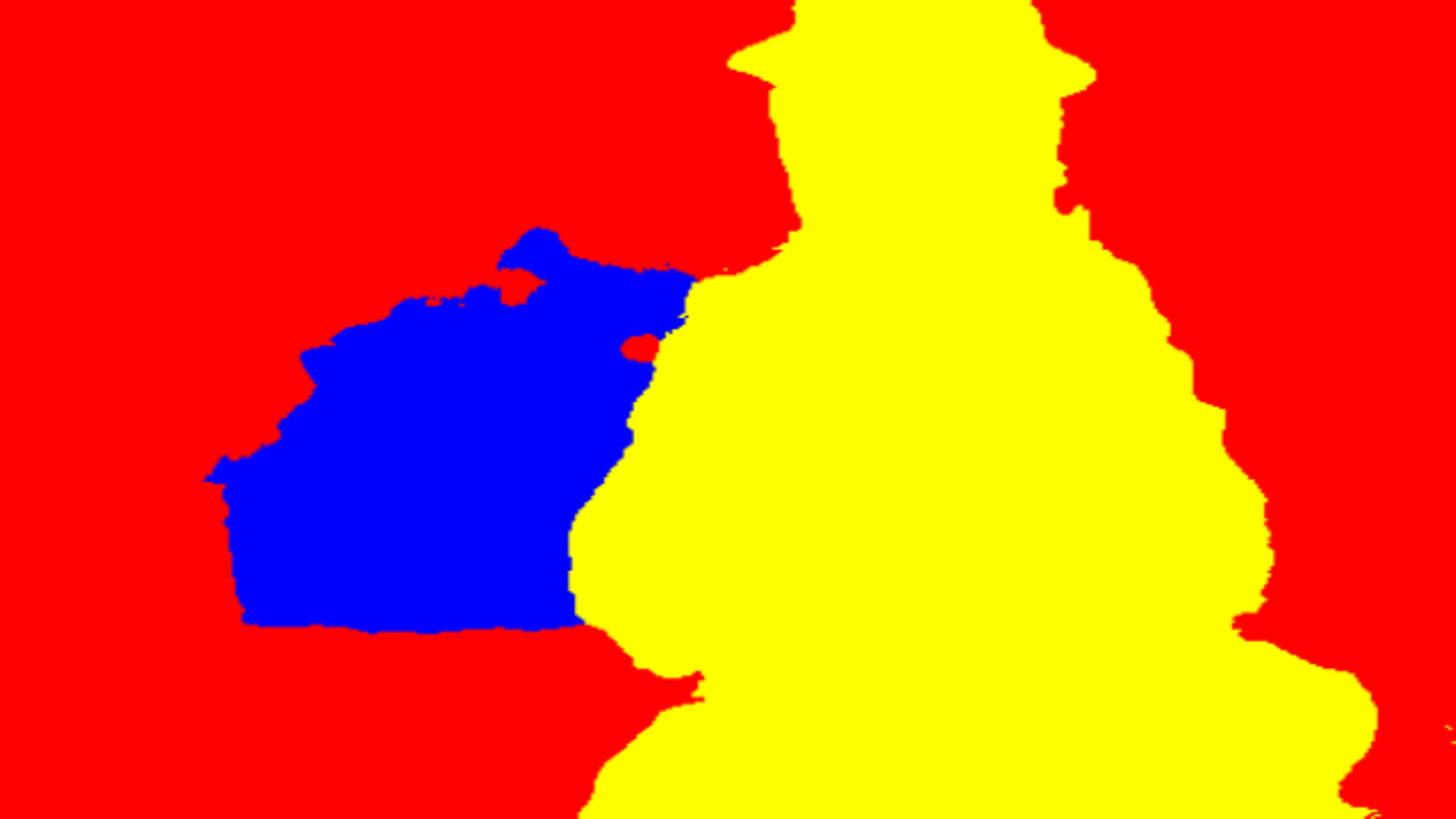}\,
            \includegraphics[width=0.19\textwidth]{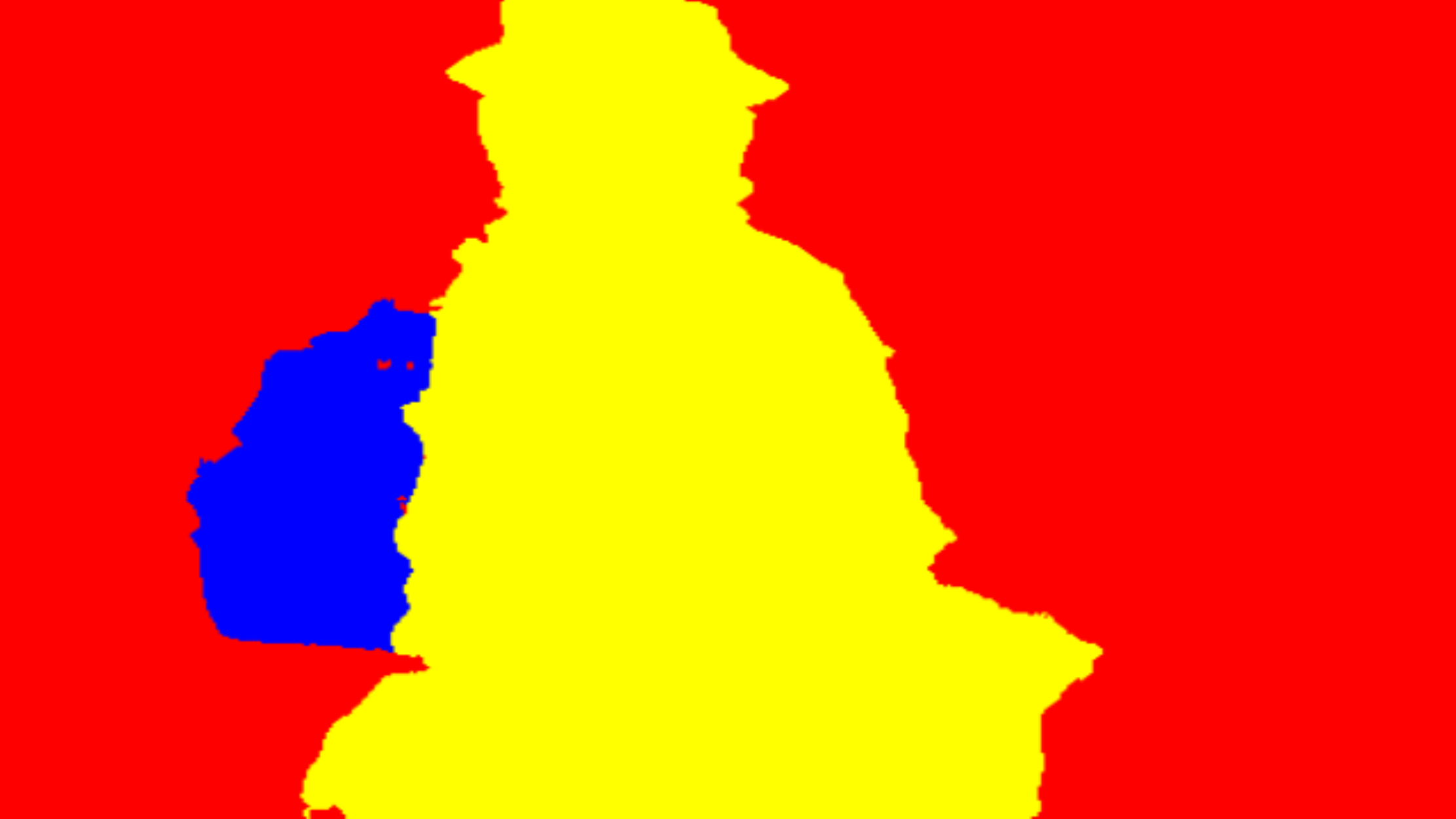}\,\\
            
            \includegraphics[width=0.19\textwidth]{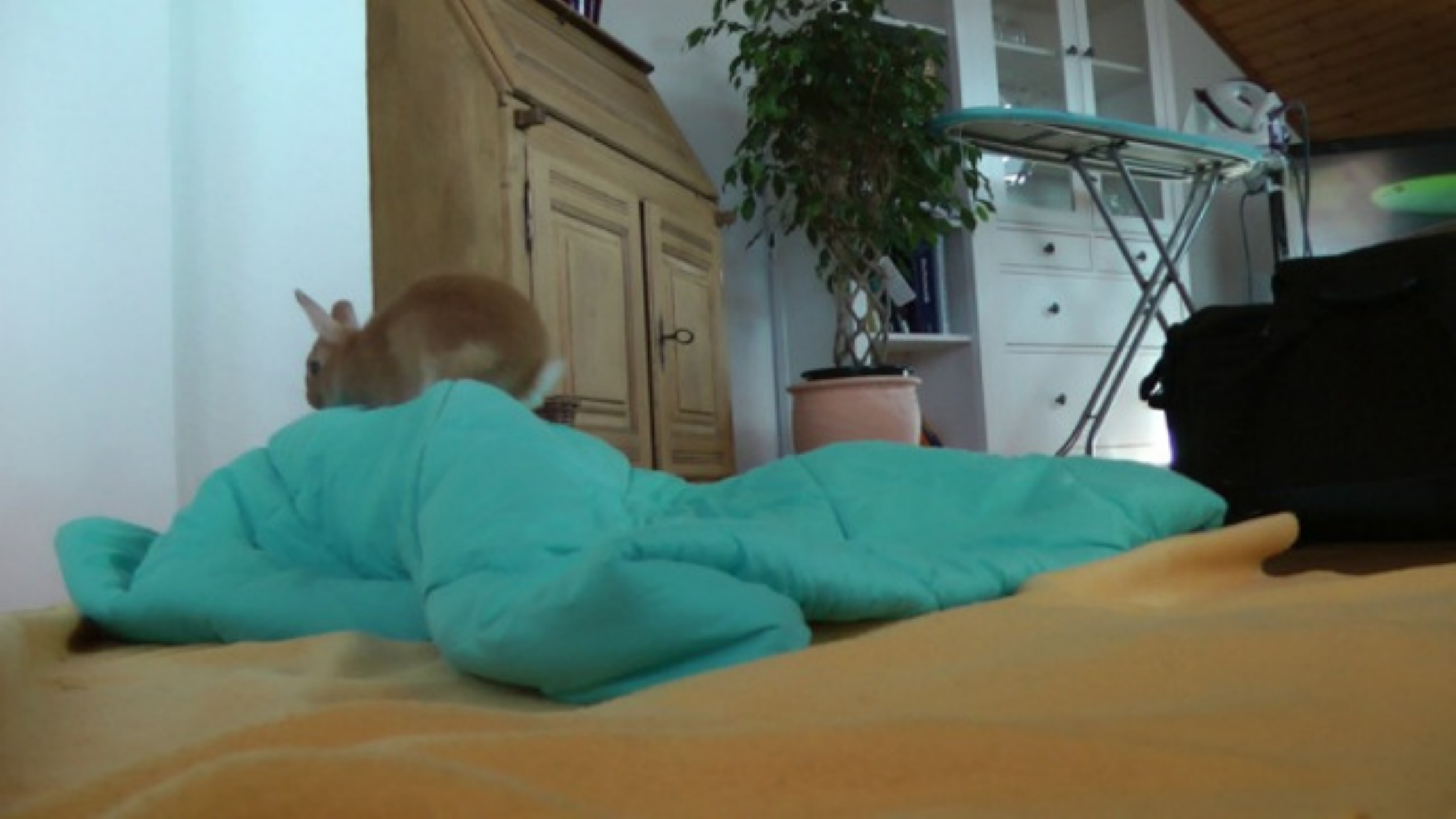}\,
            \includegraphics[width=0.19\textwidth]{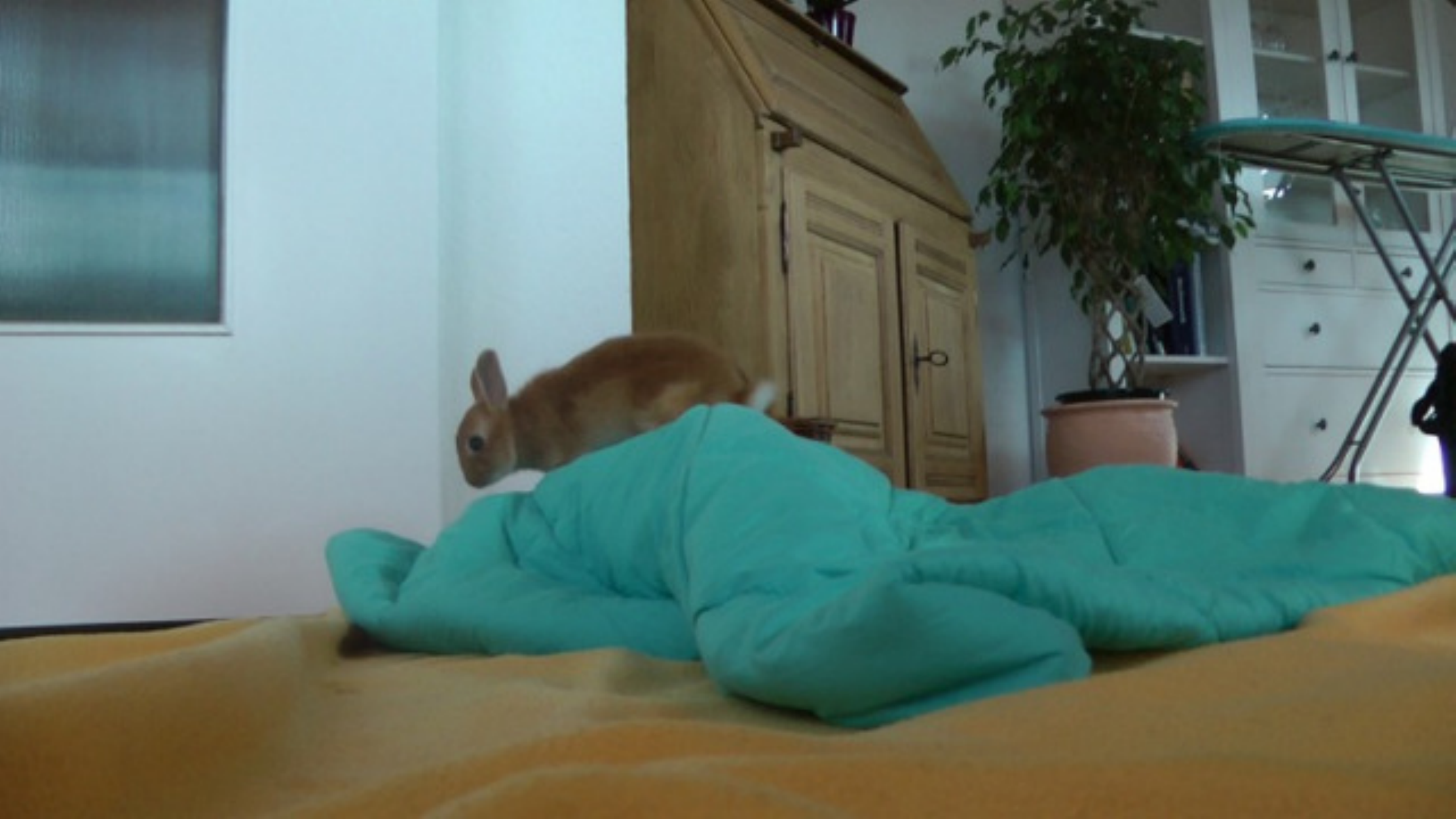}\,
            \includegraphics[width=0.19\textwidth]{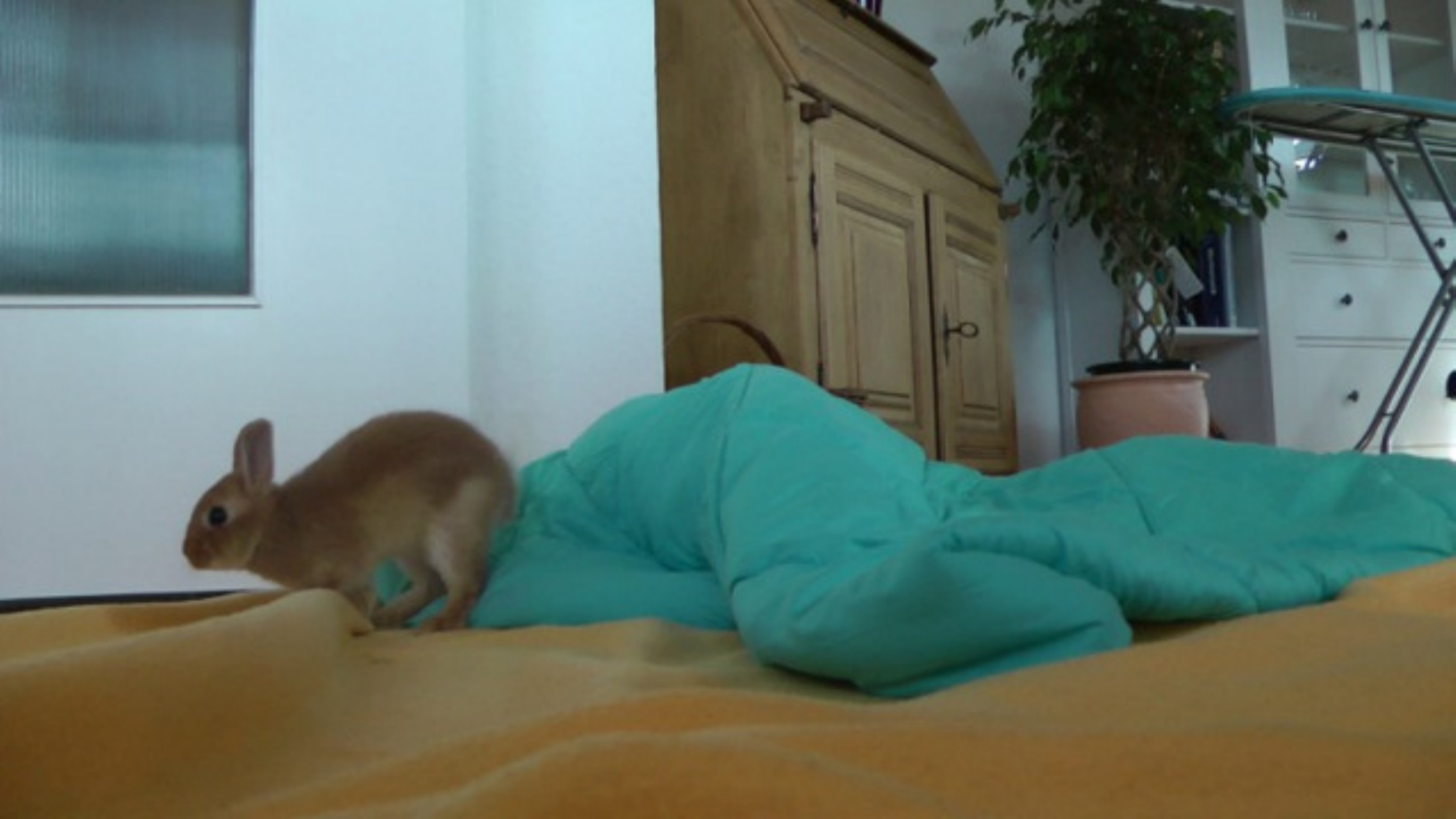}\,
            \includegraphics[width=0.19\textwidth]{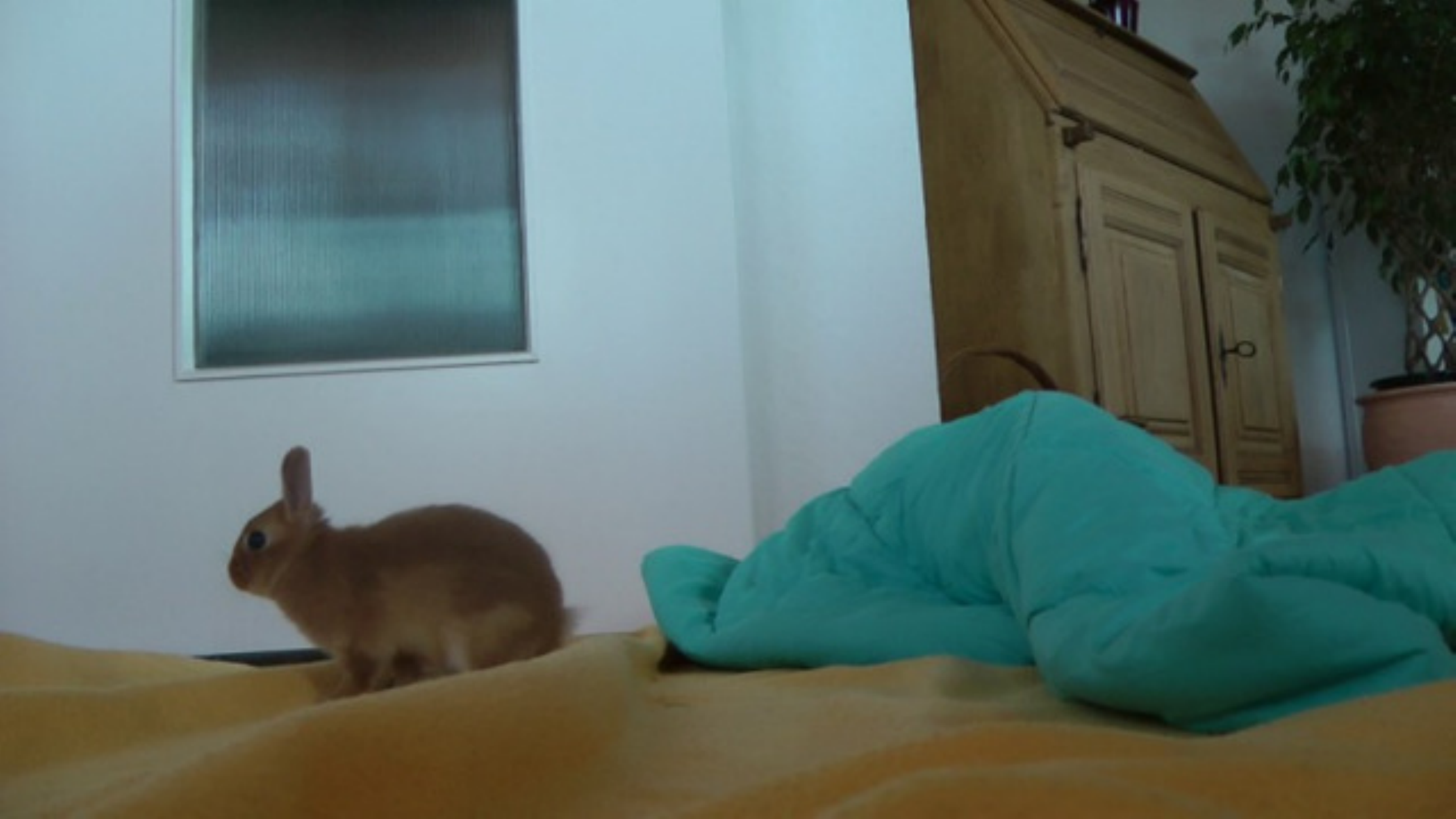}\,
            \includegraphics[width=0.19\textwidth]{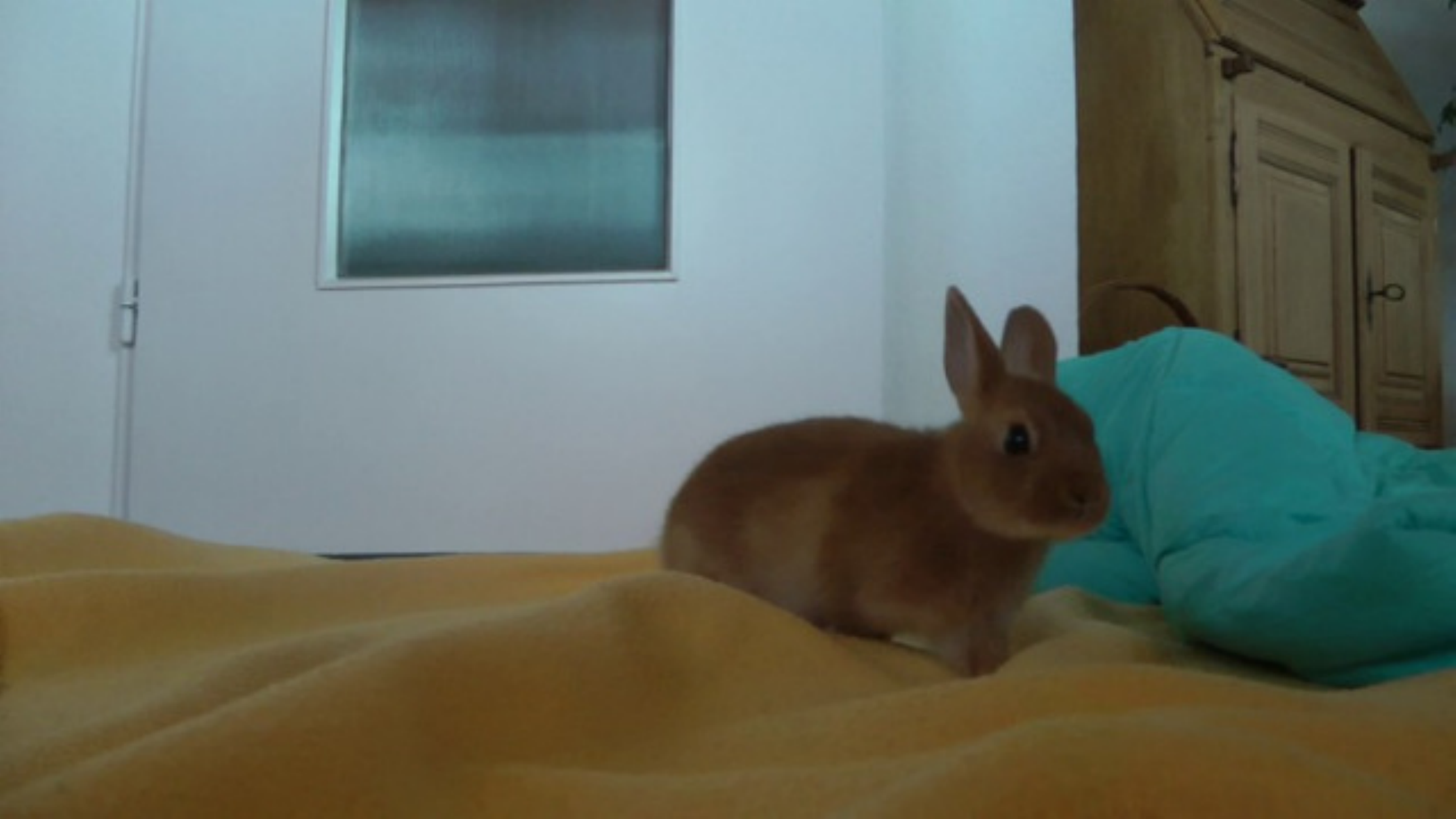}\,\\
            \includegraphics[width=0.19\textwidth]{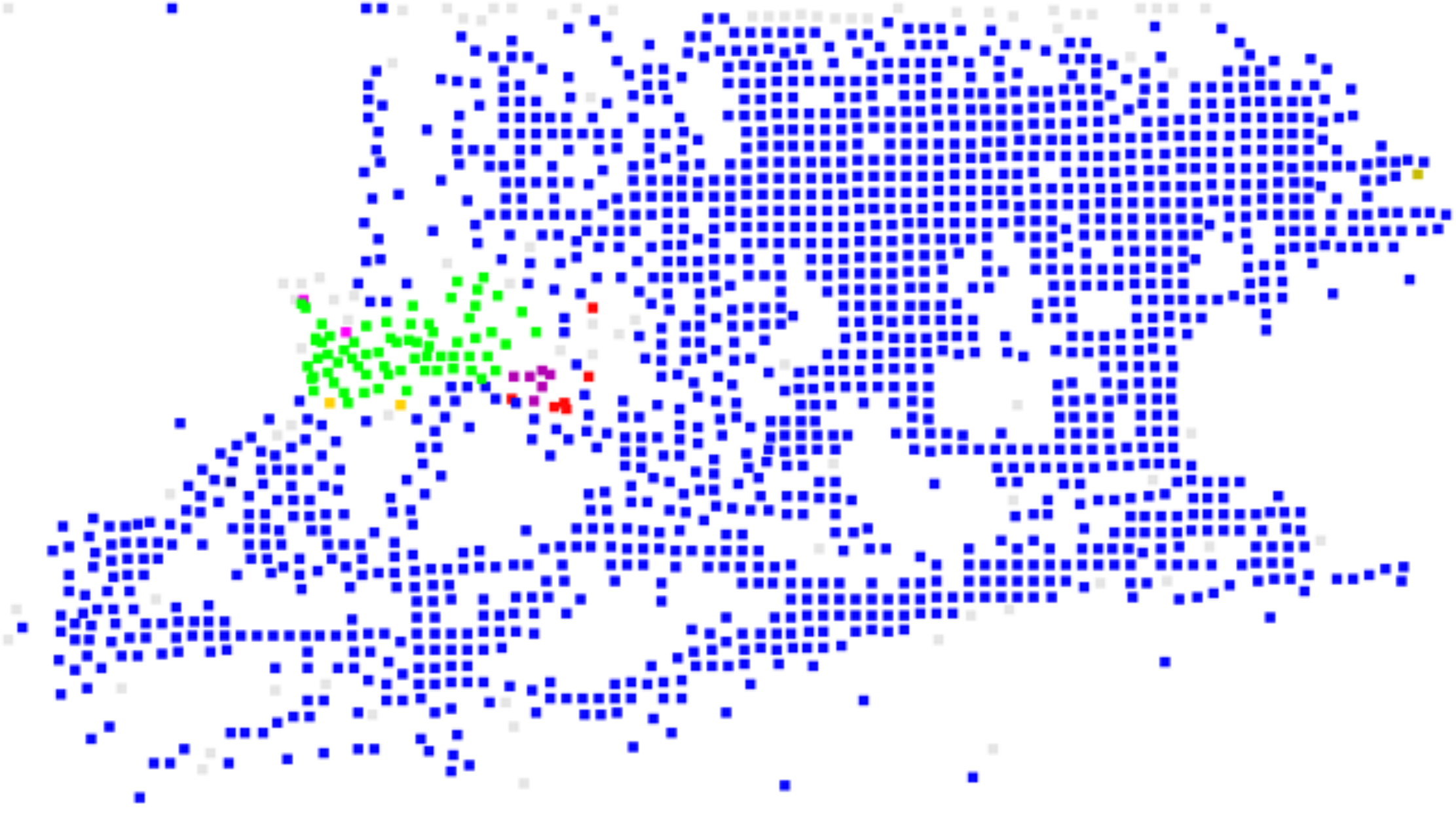}\,
            \includegraphics[width=0.19\textwidth]{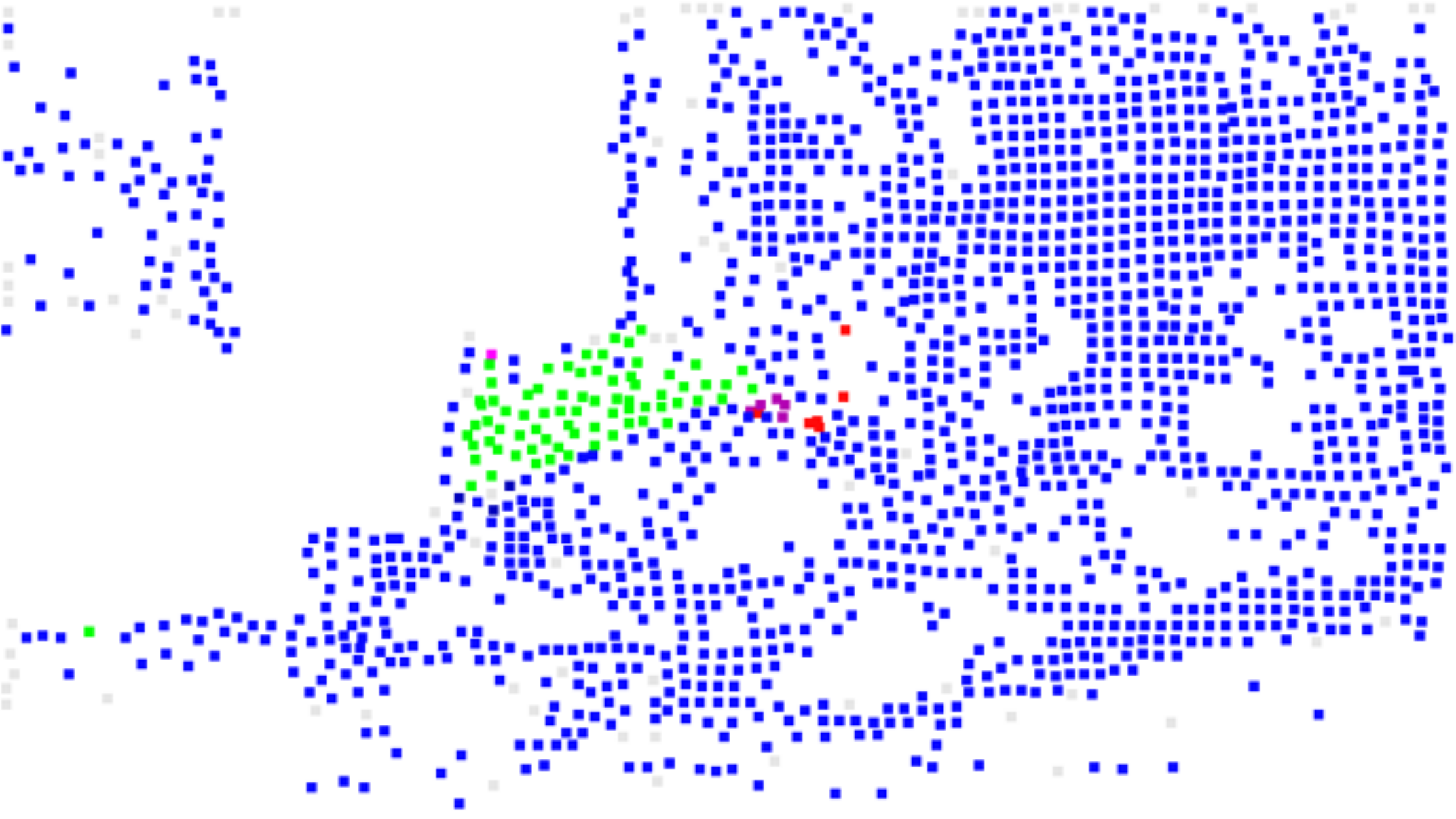}\,
            \includegraphics[width=0.19\textwidth]{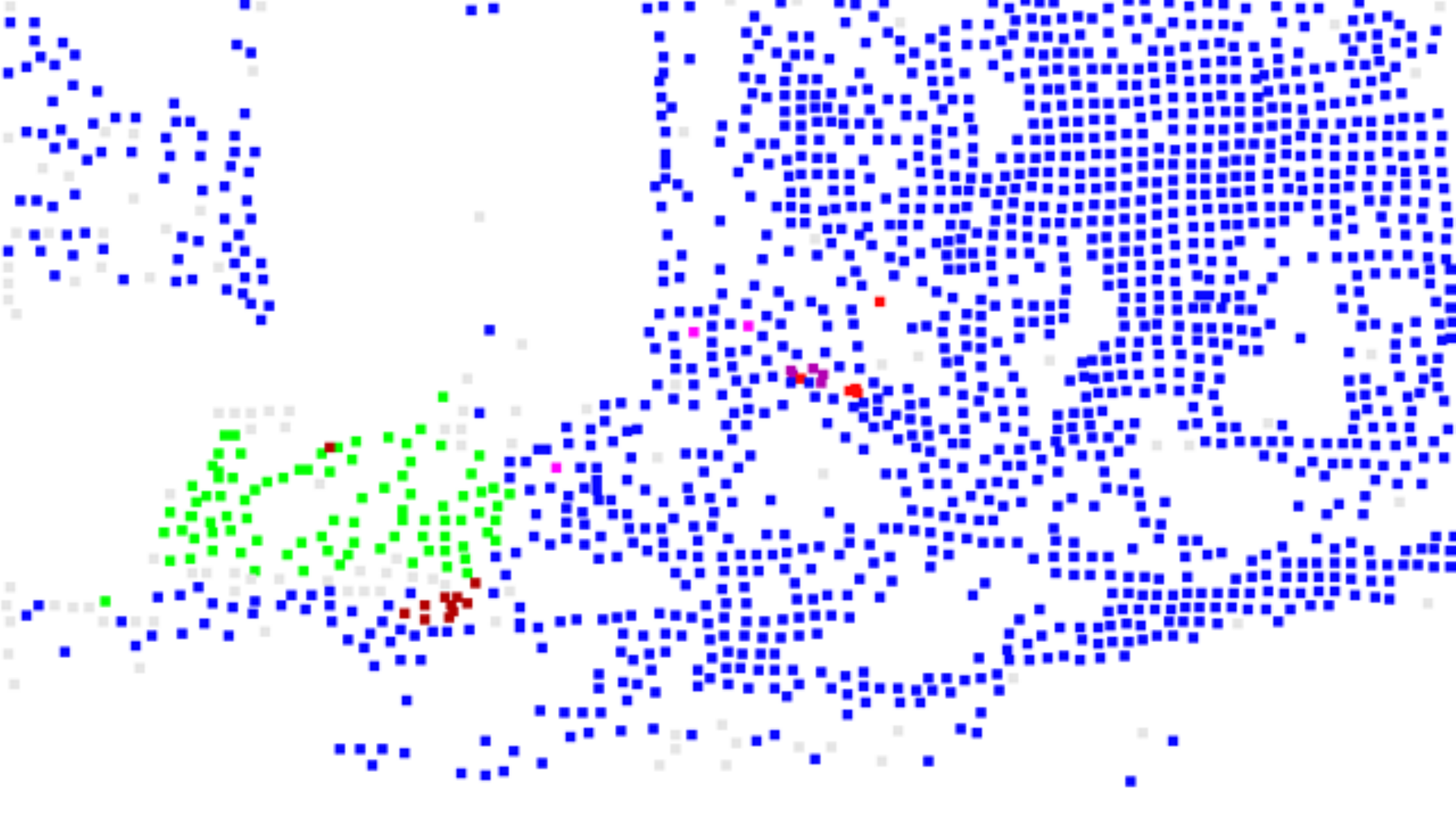}\,
            \includegraphics[width=0.19\textwidth]{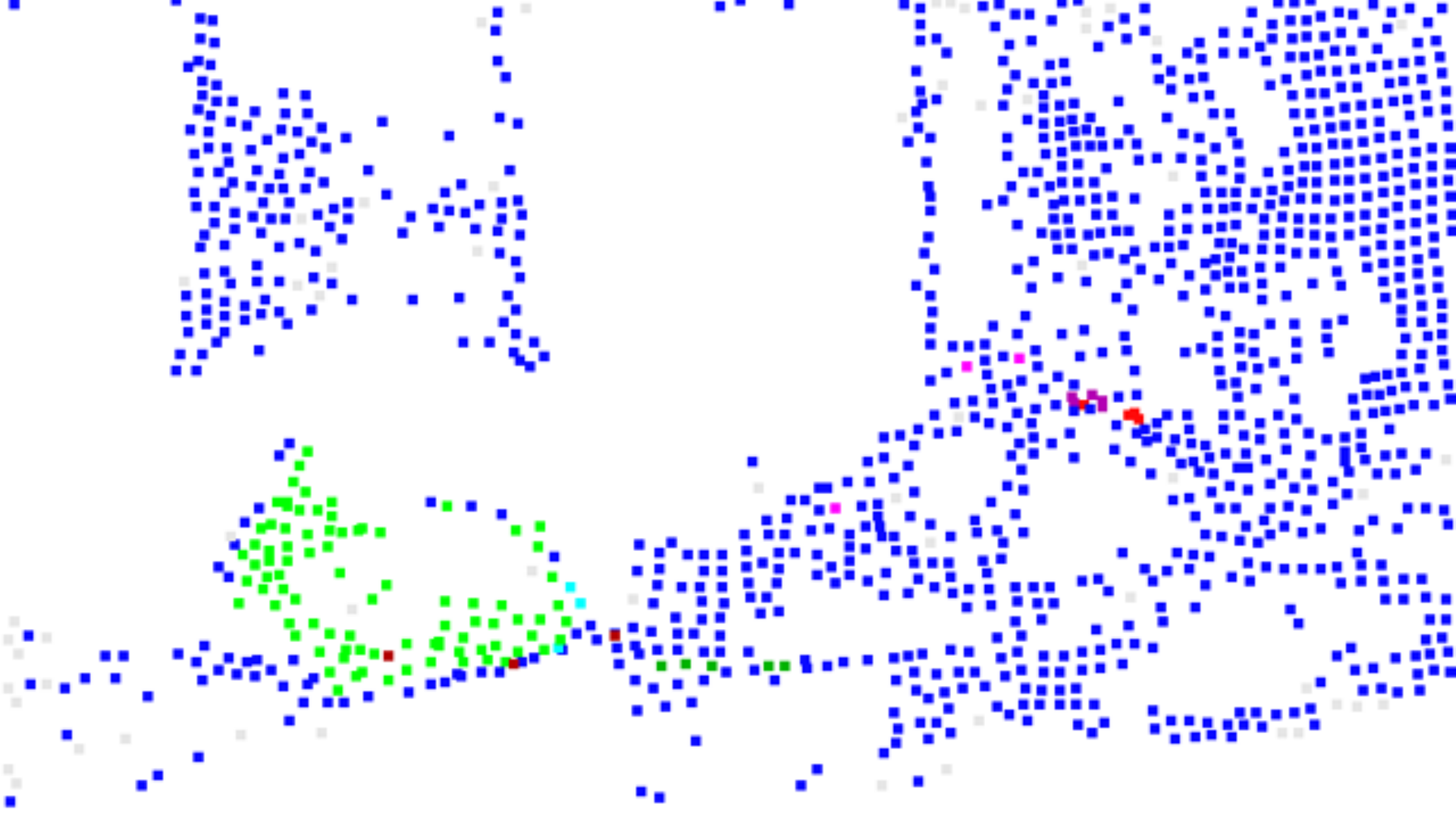}\,
            \includegraphics[width=0.19\textwidth]{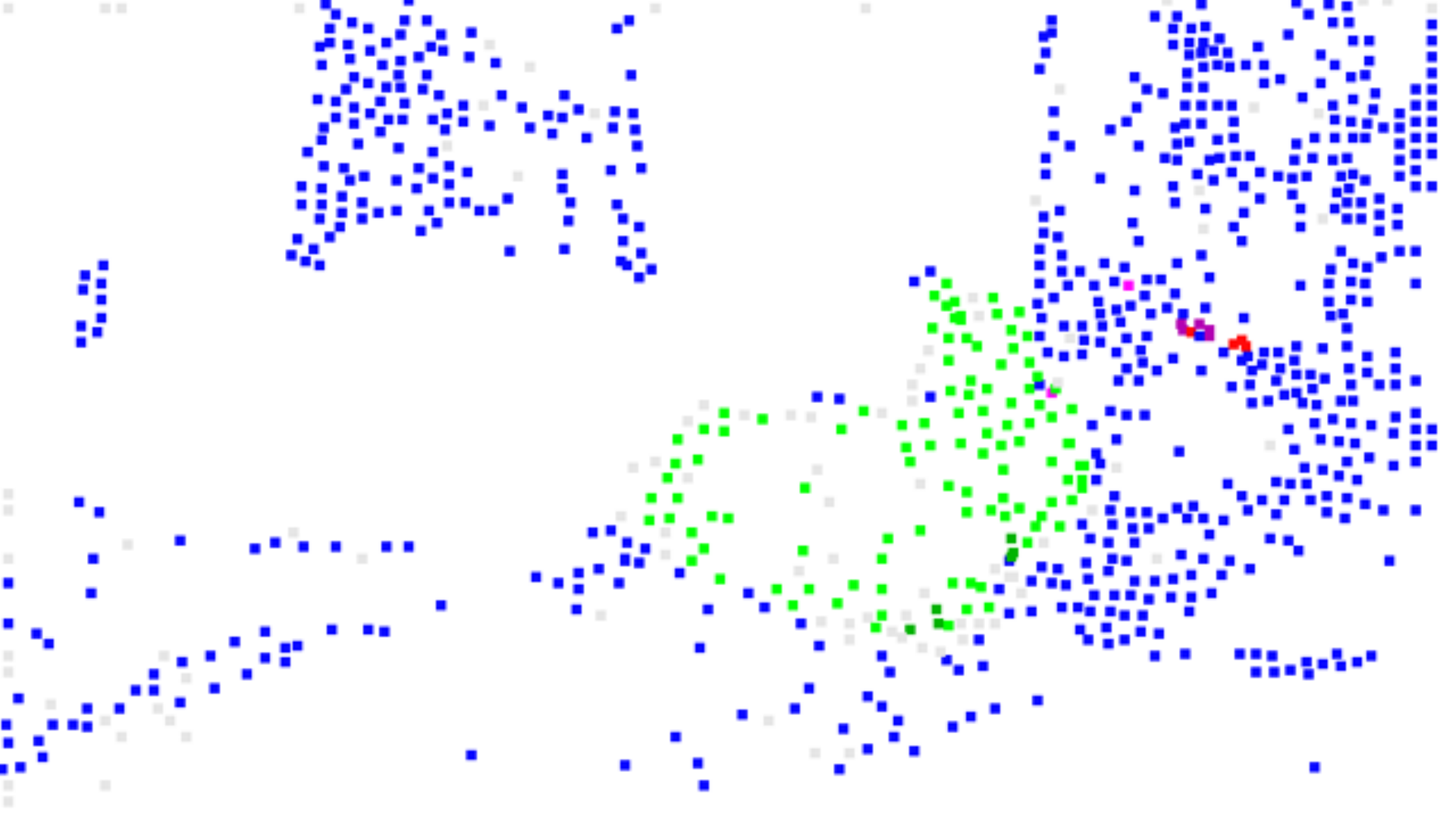}\,\\
            \includegraphics[width=0.19\textwidth]{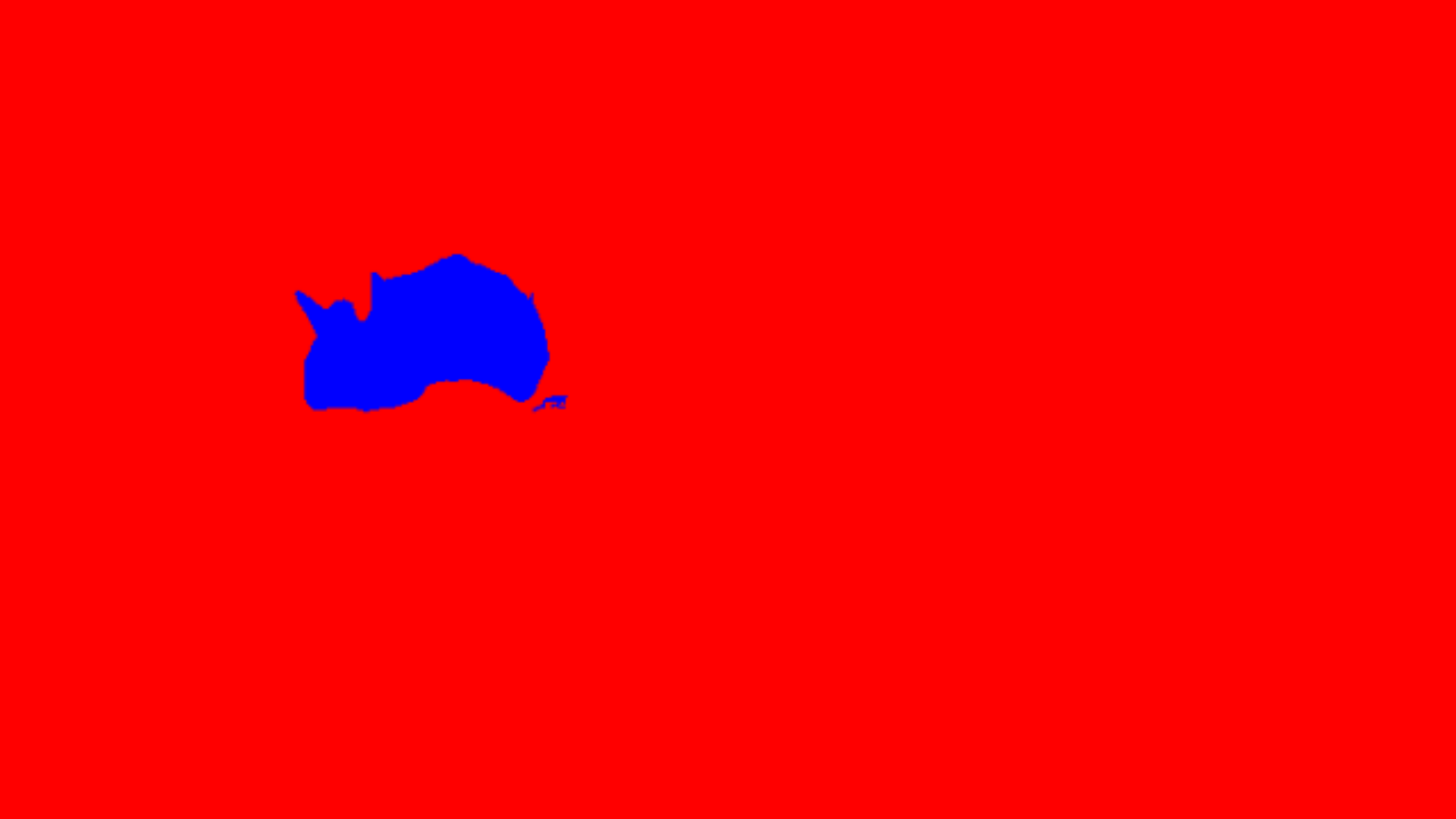}\,
            \includegraphics[width=0.19\textwidth]{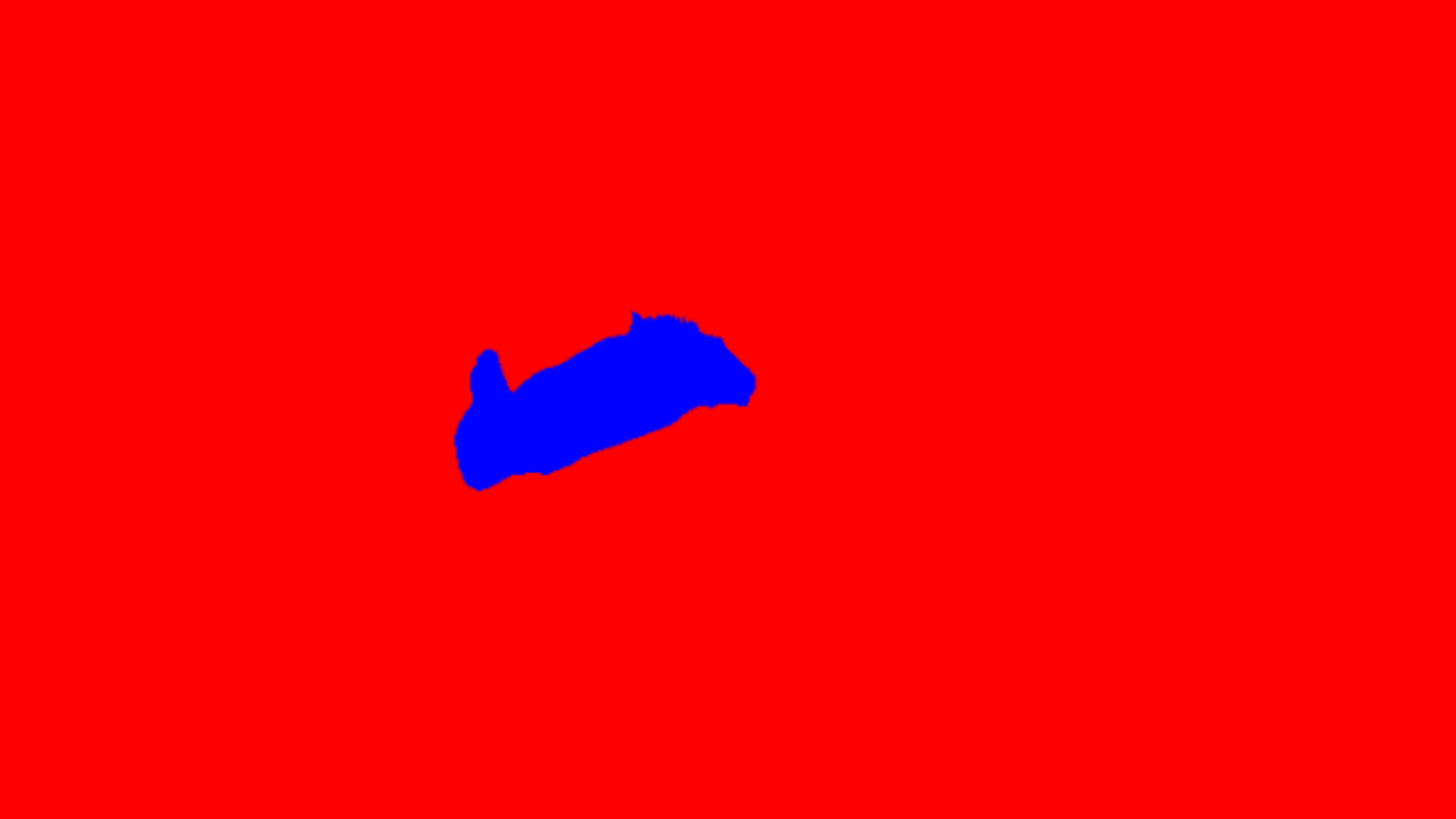}\,
            \includegraphics[width=0.19\textwidth]{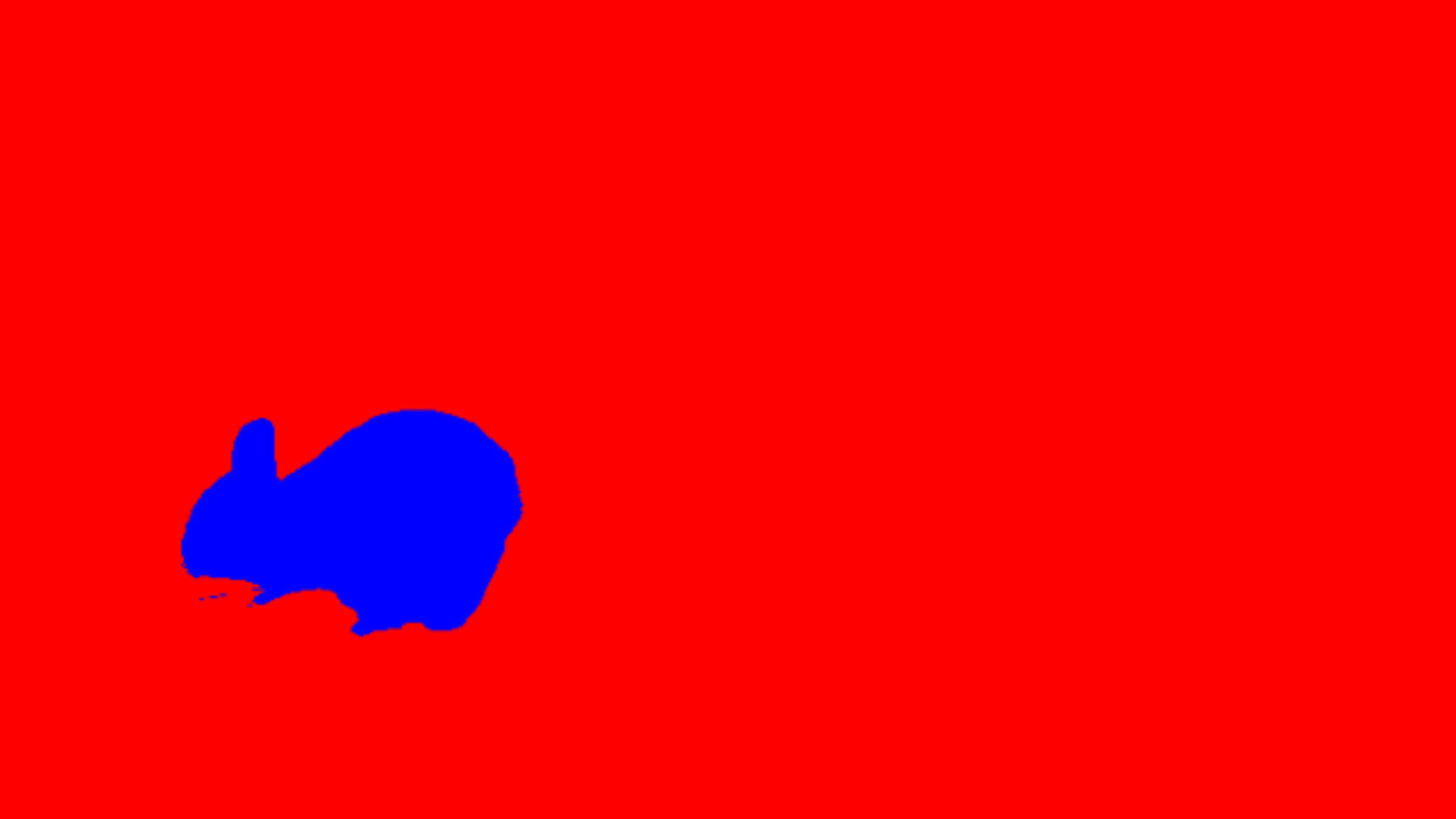}\,
            \includegraphics[width=0.19\textwidth]{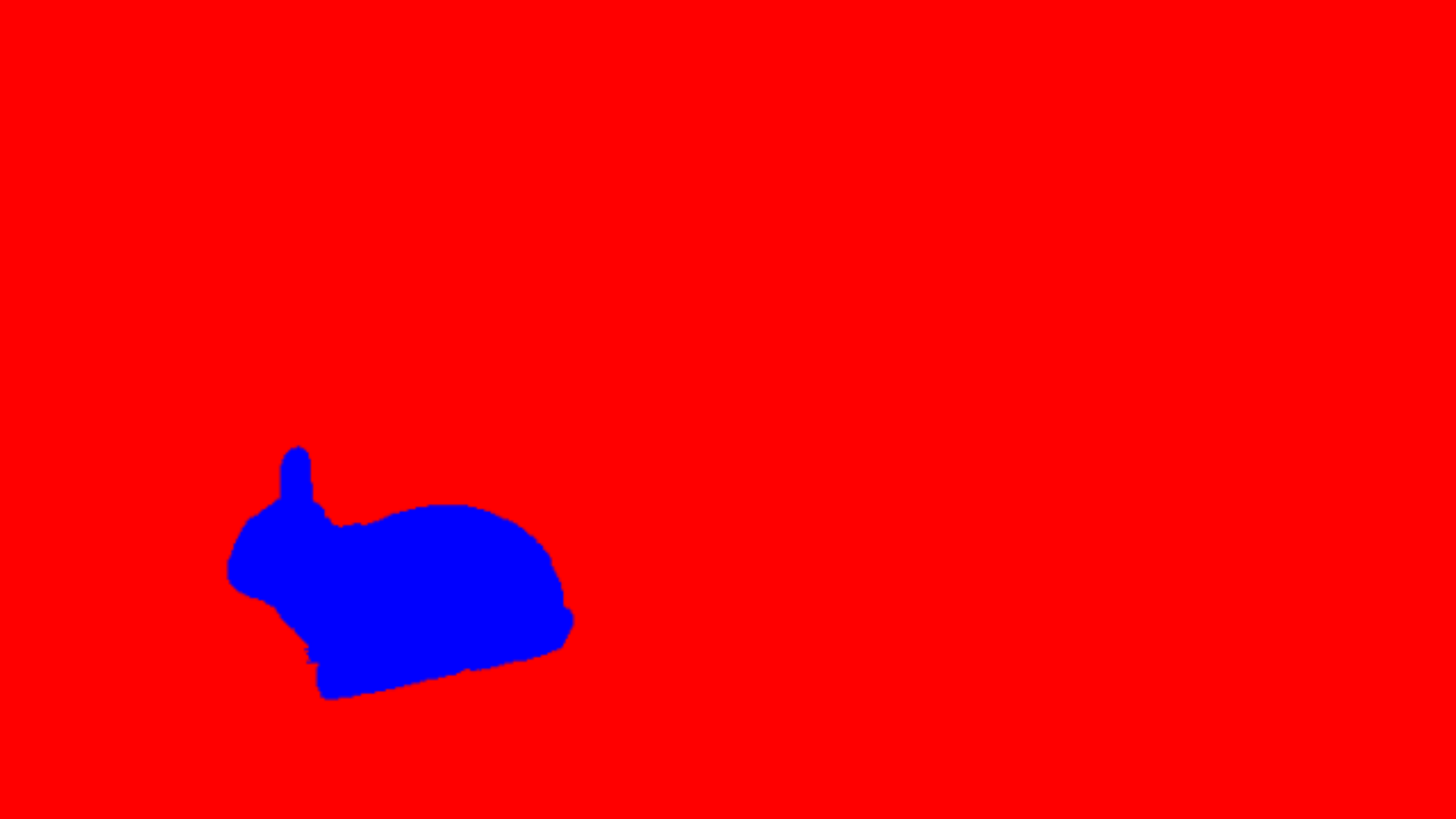}\,
            \includegraphics[width=0.19\textwidth]{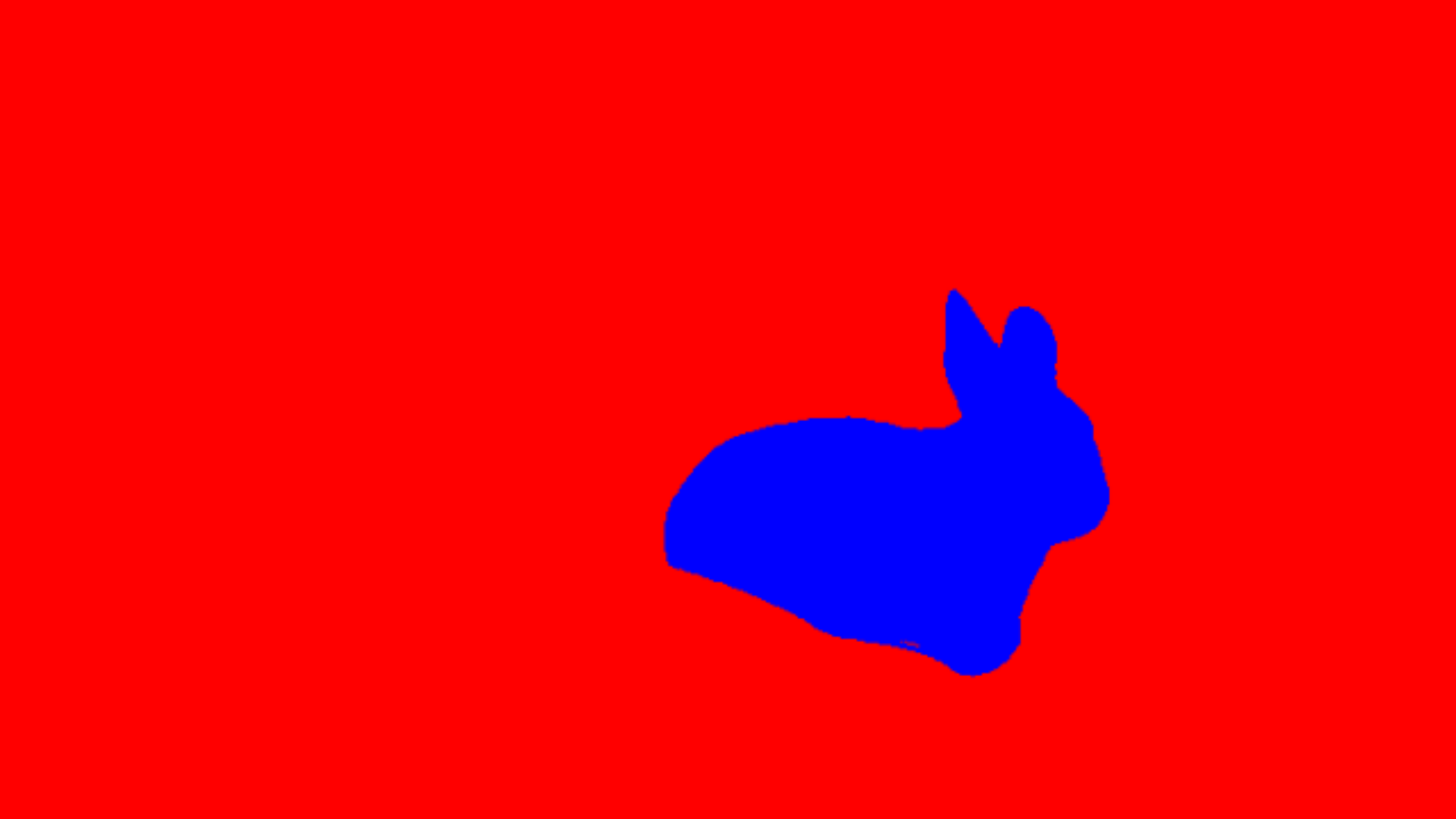}\,\\
            
        \end{tabular}
    \caption{Exemplary single- and multi-label motion segmentation results showing the image and its sparse results as well as dense segmentation for five frames in three different sequences, generated using the proposed U-Net model. 
    The images are from the FBMS59 \cite{Ochs14} dataset. Segmentations with fine details are produced even when training labels were scarce, notice how scarce the labels are for ``rabbit'' images in the 8th row. White areas are parts without any label.}
    \label{examples_fig_1}
    \end{center}
\end{figure}

All CRF versions improve the F-measure. The CRF-general performs on par with dense-CRF by only using our sparse labels for training. See Fig. \ref{examples_fig_2} and Fig. \ref{examples_fig_1} for qualitative results of our model on $\mathrm{DAVIS}_{16}$ \cite{davis_16} and FBMS59 \cite{Ochs14} datasets, respectively.

\begin{table}[t!]
\centering
\caption{Evaluation of self supervised training on sequences from $\mathrm{DAVIS}_{16}$ validation and comparison with other methods is provided. Effect of adding color information (RGB) to the edge maps (sobel) is studied (\emph{ours}) and comparison between (pretrained) dense-CRF (\emph{dense}), CRF-per-seq (\emph{per-seq}) and CRF-general (\emph{general}) is provided (for different versions of CRF refer to \ref{dense_seg_move_obj}). We studied the effect of our best model while training it only on 50\%, 70\% and 90\% of the frames in the last three rows.}
\small
\begin{tabular}{l|c|c|c}
\toprule
                                        & \% of frames & Jaccard Index       & F-measure       \\ 
                                        \midrule
variational \cite{Ochs14}               & 100 & 57.7   & 57.1   \\ 
appearance + GRU \cite{TokmakovICCV}    & 100 & 59.6    & -       \\
sobel + \emph{dense}          & 100 & 62.6   & 54.0   \\
sobel + RGB (\emph{ours})                           & 100 & 61.3   & 49.0   \\ 
\emph{ours} + \emph{dense}  & 100 & \bf{67.1}   & 60.2   \\ 
\emph{ours} + \emph{per-seq}       & 100 & 66.2   & 60.3   \\
\emph{ours} + \emph{general} & 100 & 66.2   & \bf{62.1}   \\
\midrule
\emph{ours} + \emph{general}        & 50  & 59.6   & 50.4  \\
\emph{ours} + \emph{general}        & 70  & 62.3   & 53.5   \\
\emph{ours} + \emph{general}        & 90  & 63.4   & 55.4   \\
\bottomrule
\end{tabular}
\label{densif}
\end{table}

\begin{table}[ht!]
\centering
\caption{We evaluate our densification method on FBMS59 (train) using sparse motion segmentations from Keuper et al. \cite{Keuper_15}. The sparse trajectories are produced with different flow estimation methods (LDOF~\cite{LDOF} and FlowNet2~\cite{FlowNet2}) and densified with our proposed U-Net model (using edge maps (sobel) and color information (RGB) (\emph{ours})). Further, we study on different CRF methods, (pretrained) dense-CRF (\emph{dense}) and CRF-general (\emph{general}). For more details about different versions of CRF refer to section \ref{dense_seg_move_obj}.}
\small
\begin{tabular}{l|c|c|c}
\toprule
                             & Precision     & Recall       & F-measure  \\ \midrule
Ochs et al. \cite{Ochs14}    & 85.31         & 68.70        & 76.11      \\
LDOF + \emph{ours} + \emph{dense}     & 89.35         & 67.67        & 77.01      \\
FlowNet2 + \emph{ours} + \emph{dense} & 89.59         & 68.29        & \bf{77.56}      \\
FlowNet2 + \emph{ours} + \emph{general} & 89.27         & 68.20        & 77.33      \\
\bottomrule
\end{tabular}
\label{fbms_rgb_sobel}
\end{table}

\subsubsection{Partly trained Model}
We further evaluate how well our model  works for video frames for which no sparse labels are given during training. Please note that, throughout the sequences, the appearance of an object can change drastically. In Tab.~\ref{densif} (bottom), we report results for the sequence specific U-Net model + \emph{CRF-general} trained on the first  50\%, 70\% and 90\% of the frames and evaluated on the remaining frames. While there is some loss in Jaccard's index compared to the model evaluated on seen frames (above), the performance only drops slightly as smaller portions of the data are used for training. 

\subsubsection{Densification on FBMS59}
Next, we evaluate our sequence specific model for label densification on FBMS59 \cite{Ochs14}. We study on two different variants of optical flow (FlowNet2 \cite{FlowNet2} and Large Displacement Optical Flow (LDOF)~\cite{LDOF}) for trajectory generation and sparse motion segmentation \cite{Keuper_15}. The results in Tab.~\ref{fbms_rgb_sobel} show that the proposed approach outperforms the approach of Ochs~et~al.~\cite{Ochs14}. Improved optical flow leads to improved results overall. The different CRF versions do not provide significantly different results.

\section{Conclusion}
In this paper, we have addressed the segmentation of moving objects from single frames. To that end, we proposed a GRU-based trajectory embedding to produce high quality sparse segmentations automatically. Furthermore, we closed the gap between sparse and dense results by providing a self-supervised U-Net model trained on sparse labels and relying only on edge maps and color information. The trained model on sparse points provides single and multi-label dense segmentations. 
The proposed approach generalizes to unseen sequences from FBMS59 and $\mathrm{DAVIS}_{16}$ and provides competitive and appealing results.

\section*{Acknowledgment}
We acknowledge funding by the DFG project KE 2264/1-1. We also acknowledge the NVIDIA Corporation for the donation of a Titan Xp GPU.

\FloatBarrier
\bibliographystyle{splncs}
\bibliography{egbib}

\end{document}